\documentclass[runningheads]{llncs}

\usepackage{eccv}

\usepackage{eccvabbrv}

\usepackage{graphicx}
\usepackage{booktabs}

\usepackage[accsupp]{axessibility}  

\usepackage{hyperref}

\usepackage{orcidlink}

\usepackage{bm}

\usepackage{graphicx}
\usepackage{colortbl,xcolor,array}
\usepackage{multirow}
\usepackage{subcaption}

\usepackage{wrapfig}

\usepackage[normalem]{ulem}
\useunder{\uline}{\ul}{}

\newcommand{\layoutcorrector}{Layout-Corrector}
\newcommand{\masktoken}{\texttt{[MASK]}}
\newcommand{\padtoken}{\texttt{[PAD]}}

\begin{document}

\title{Layout-Corrector: Alleviating Layout Sticking Phenomenon in Discrete Diffusion Model} 

\titlerunning{Layout-Corrector}

\author{
    Shoma Iwai\inst{1}\thanks{\scriptsize{This work was done during the first author's internship at LY Corporation.}}\orcidlink{0000-0002-6340-3902} \and
    Atsuki Osanai\inst{2}\orcidlink{0009-0004-8063-5117} \and \\
    Shunsuke Kitada\inst{2}\orcidlink{0000-0002-3330-8779} \and
    Shinichiro Omachi\inst{1}\orcidlink{0000-0001-7706-9995}
}

\authorrunning{Iwai~et al.}

\institute{
    Graduate School of Engineering, Tohoku University, Japan 
    \email{shoma.iwai.s4@dc.tohoku.ac.jp},~
    \email{shinichiro.omachi.b5@tohoku.ac.jp}\\
    \and
    LY Corporation, Japan
    \email{\{atsuki.osanai, s.kitada\}@lycorp.co.jp}
}

\maketitle

\begin{abstract}
Layout generation is a task to synthesize a harmonious layout with elements characterized by attributes such as category, position, and size.
Human designers experiment with the placement and modification of elements to create aesthetic layouts, however, we observed that current discrete diffusion models (DDMs) struggle to correct inharmonious layouts after they have been generated.
In this paper, we first provide novel insights into \textit{layout sticking} phenomenon in DDMs and then propose a simple yet effective layout-assessment module Layout-Corrector, which works in conjunction with existing DDMs to address the \textit{layout sticking} problem.
We present a learning-based module capable of identifying inharmonious elements within layouts, considering overall layout harmony characterized by complex composition.
During the generation process, Layout-Corrector evaluates the correctness of each token in the generated layout, reinitializing those with low scores to the ungenerated state.
The DDM then uses the high-scored tokens as clues to regenerate the harmonized tokens.
Layout-Corrector, tested on common benchmarks, consistently boosts layout-generation performance when in conjunction with various state-of-the-art DDMs.
Furthermore, our extensive analysis demonstrates that the Layout-Corrector (1) successfully identifies erroneous tokens, (2) facilitates control over the fidelity-diversity trade-off, and (3) significantly mitigates the performance drop associated with fast sampling.
\end{abstract}

\section{Introduction}\label{sec:intro}

Creating a layout is one of the most crucial tasks involving human labor when designing~\cite{agrawala2011design}, and there are a wide variety of applications, including academic papers~\cite{zhong2019publaynet}, application user interfaces~\cite{deka2017rico}, and advertisements~\cite{yamaguchi2021canvasvae}.
Layout generation has been formulated as a task that determines a set of elements that consist of categories, positions, and sizes~\cite{lok2001survey, shi2023intelligent}.
In response, layout-generation methods on deep learning have shown remarkable performance, and in particular, discrete generative models such as masked language modeling~\cite{devlin2019bert}-based methods~\cite{turgutlu2022layoutbert,chang2022maskgit,kong2022blt} and discrete diffusion models (DDMs)~\cite{gu2022vector,inoue2023layoutdm,hui2023unifying,zhang2023layoutdiffusion} are the current state-of-the-art (SoTA).

\begin{wrapfigure}[19]{r}[0pt]{0.5\linewidth}
    \centering
    \raisebox{0pt}[\dimexpr\height-2\baselineskip\relax]{\includegraphics[width=\linewidth]{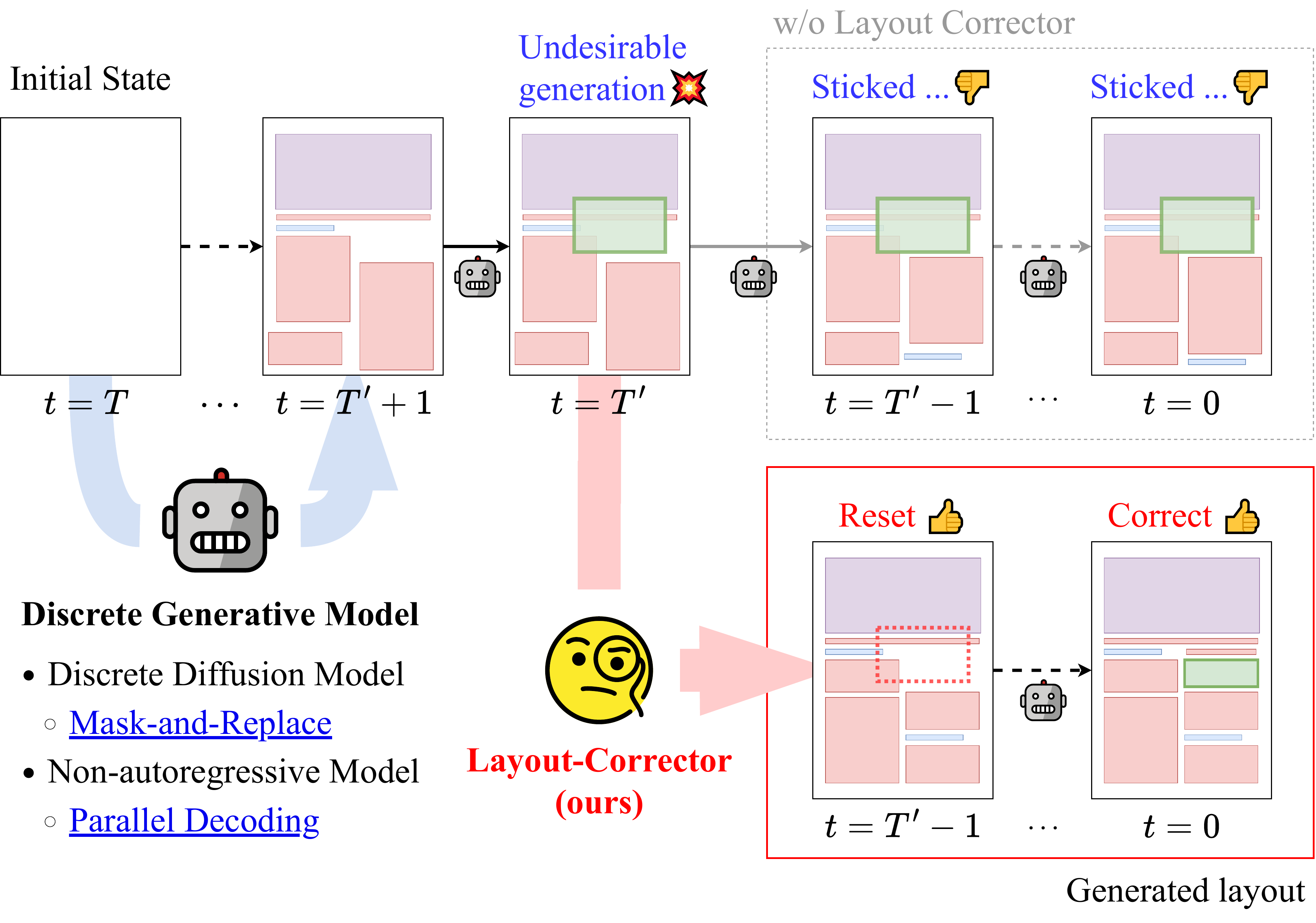}}
    \caption{
        Intuitive overview of \layoutcorrector{}.
        Conventional generative models cannot modify the elements once they have been generated.
        \layoutcorrector{} works in conjunction with DDMs to identify inharmonious elements in the generative process and initialize them to enhance regeneration towards a harmonized layout.        
        }
    \label{fig:intro}
\end{wrapfigure}

To create an aesthetically pleasing layout, human designers typically modify layouts through trial and error.
However, we found that even SoTA DDMs can not update elements in layouts once they have been generated, \ie, \textit{layout sticking}.
An intuitive example of the sticking behavior is depicted in \cref{fig:intro}, where inharmonious elements that arose during generation persist until the final generated result.
While former studies~\cite{merrell2011interactive,o2014learning,kikuchi2021constrained, chen2024towards} tried to refine these elements in the post-processing phase that minimizes the rule-based costs such as alignment, they could not capture higher-order senses that determine layout aesthetics.

In the image generation domain, non-autoregressive (Non-AR) decoding methods with an external critic have demonstrated remarkable performance~\cite{lezama2022improved, lezama2023discrete}.
The module identifies deviated visual tokens from the real distribution and reset them to resample.
Reviewing the success of the masked image modeling~\cite{he2022mae}, a few visual clues can provide plenty of information to identify erroneous tokens.
On the other hand, the layout domain has different characteristics; (i) unlike images with a fixed and enough number of tokens~(\eg, $16\times16$ patches), the number of layout elements is small and varies across samples~(\eg, $1$ to $25$ elements), and (ii) as the element composed of multiple attributes, then partially observed tokens do not provide enough clues.
Thus, it is non-trivial whether the technique in the vision can apply to the layout generation.

In this paper, we propose a simple yet effective approach, named \textit{\layoutcorrector{}}, to address the layout sticking problem.
It works as an external module that evaluates each token's correctness score in a learning-based manner, aiming to identify erroneous tokens in a layout.
As shown in \cref{fig:intro}, during the generation process, tokens with low correctness scores are reset to the ungenerated state (\ie, \masktoken). Then, a DDM regenerates them using the remaining high-scored tokens as clues.
Additionally, to deal with the characteristics of the layout tokens mentioned above, we propose a new objective and application schedule that accommodates variable numbers of elements while providing reliable layout cues.

We conducted extensive experiments on \layoutcorrector{} using three benchmarks~\cite{deka2017rico, zhong2019publaynet, yamaguchi2021canvasvae}.
When in conjunction with strong baselines~\cite{chang2022maskgit,gu2022vector,inoue2023layoutdm}, \layoutcorrector{} significantly enhanced their performance in both unconditional and conditional generation tasks.
Both quantitative and qualitative evaluations confirmed that our approach effectively corrects inharmonious layout elements, addressing the challenge present in SoTA DDMs.
By adjusting the application schedule of the corrector, we also achieved enhanced control over the fidelity-diversity and speed-quality trade-offs, demonstrating \layoutcorrector{}'s versatility across different application scenarios.

Our contributions are summarized as follows: 
\textbf{(1)} We empirically demonstrate that current SoTA DDMs struggle to correct inharmonious elements in layouts; however, they can effectively correct them when erroneous elements are initialized to the ungenerated state, \masktoken. 
\textbf{(2)} We propose \layoutcorrector{} for evaluating the correctness score of each element and resetting the element with a lower score to \masktoken, enabling DDMs to regenerate improved layouts.
\textbf{(3)} We confirm consistent improvements by applying \layoutcorrector{} to various DDMs.
We also analyze the behavior of \layoutcorrector{} and demonstrate that it enhances fidelity-diversity and speed-quality trade-offs.

\section{Related Works}\label{sec:related}

\textbf{Layout Generation.}
Automatic layout generation~\cite{lok2001survey, shi2023intelligent} is a task involving the assignment of positions and categories to multiple elements, which has diverse applications in design areas like application user interfaces and academic papers~\cite{deka2017rico,zhong2019publaynet,yamaguchi2021canvasvae,zhao2021guigan,yang2016automatic,zheng2019content,guo2021vinci,hsu2023posterlayout,zhou2022composition}.
This task includes unconditional and conditional generation, considering user constraints, \eg, partially specified elements.

Early layout generation research explored classical optimization~\cite{o2014learning,o2015designscape} and generative models such as GAN~\cite{goodfellow2014generative}-based models~\cite{li2019layoutgan,zheng2019content,zhou2022composition} and VAE~\cite{kingma2013auto}-based models~\cite{jyothi2019layoutvae,yamaguchi2021canvasvae}. 
Following the success in NLP, Transformer-based approaches~\cite{vaswani2017attention} were proposed. 
Auto-regressive (AR) models~\cite{gupta2021layouttransformer,arroyo2021variational} iteratively generate layouts, however, struggle with conditional generation~\cite{kong2022blt}. 
Non-AR models~\cite{turgutlu2022layoutbert,kong2022blt,chang2022maskgit} overcome this difficulty by using a bidirectional architecture, where user-defined conditions serve as clues to complete blank tokens.
Recently, diffusion model-based~\cite{sohl2015deep,ho2020denoising} layout generation methods in both continuous~\cite{chai2023layoutdm,levi2023dlt} and discrete spaces~\cite{inoue2023layoutdm,hui2023unifying,zhang2023layoutdiffusion} have been developed. 
To enable unconditional and conditional generation within a single framework, it is essential for models to process both discrete and continuous data present in elements.
DDMs can accommodate both data types by quantizing geometric attributes into a binned space.

\noindent
\textbf{Discrete Diffusion Models.}
D3PM~\cite{austin2021structured} introduces the special token \masktoken{}, where regular tokens are absorbed into \masktoken{} through a forward process. 
Based on this, Non-AR models, such as MaskGIT~\cite{chang2022maskgit}, can be understood as a subclass of DDMs.
MaskGIT introduces a scheduled masking rate, akin to the diffusion process in D3PM.
It also adopts the parallel decoding~\cite{ghazvininejad2019mask} based on the token confidence, serving as a deterministic denoising process.
To address the issue of non-regrettable decoding strategy, DPC and Token-Critic~\cite{lezama2022improved,lezama2023discrete} introduce an external module to mitigate discrepancies between training and inference distributions.
VQDiffusion~\cite{gu2022vector} facilitates transitions between regular tokens in addition to \masktoken{}, while LayoutDM~\cite{inoue2023layoutdm} advances the diffusion process to allow transitions within the same modality.
LayoutDiffusion~\cite{zhang2023layoutdiffusion} introduces a mild corruption process that considers the continuity of geometric attributes.
For the layout generation, we explore the potential of the correction during the generation process to alleviate layout token sticking problem.

\noindent
\textbf{Layout Correction.}
There are several studies aimed at layout modification.
In optimization-based methods~\cite{merrell2011interactive,o2014learning,kikuchi2021constrained}, layouts are refined to minimize hand-crafted costs, such as alignment score.
RUITE~\cite{rahman2021ruite} and LayoutFormer++~\cite{jiang2023layoutformer++} learn to restore the original layout from noisy input.
LayoutDM~\cite{inoue2023layoutdm} proposes a logit adjustment under the constraints of noisy layouts.
While previous research has focused on layout refinement, our method aims to correct layouts during the generation process.
Compared to rule-based optimization, our approach achieves superior performance while preserving the distribution of the generated results.
Please refer to the supplementary material for details.
\section{Method}\label{sec:method}
In Sec.~\ref{sec:discrete_models}, we first provide a brief overview of DDMs for layout-generation~\cite{inoue2023layoutdm} and examine the potential for layout correction in Sec.~\ref{sec:preliminary_experiment}.
We then present \layoutcorrector{} in Sec.~\ref{sec:corrector} and explain its application across diverse layout-generation tasks in Sec.~\ref{subsec:generate_w_cor}.

\subsection{Layout Generation Models}\label{sec:discrete_models}
A layout is represented as a set of elements, where an element consists of category, position, and size.
Following previous studies~\cite{gupta2021layouttransformer,arroyo2021variational,kong2022blt}, we use the quantized expression for geometric attributes.
Defining $\bm{l}_i = (c_i, x_i, y_i, w_i, h_i)$, a layout $l$ with $N \in \mathbb{N}$ elements is expressed as $l_N = (\bm{l}_1, \cdots, \bm{l}_N)$, where $c_i \in \{1, \cdots, C\}$ denotes the category~(\eg, text, button), $(x_i, y_i, w_i, h_i) \in \{1, \cdots, B\}^4$ represents the center position and size of $i$-th element, and $B \in \mathbb{N}$ denotes the number of bins. 
Under this representation, we review DDMs to gain insight into the behavior of the generation process, as discussed in \cref{sec:preliminary_experiment}.

Let $T$ represent the total number of timesteps in the corruption process.
We consider a scalar variable $z_t$ with $K \in \mathbb{N}$ classes at $t$, where $z_t \in \{1, \ldots, K\}$. 
Here, $z_t$ substitutes an attribute of an element.
Following LayoutDM~\cite{inoue2023layoutdm}, we include the special tokens \padtoken{} and \masktoken{}, resulting in ($K+2$) classes.
Here, \padtoken{} token is employed to fill the empty element, achieving variable length generation.
\masktoken{} token denotes the absorbing state, to which tokens converge through the diffusion process.
Using a transition matrix $\mathbf{Q}_t \in [0, 1]^{(K+2) \times (K+2)}$, we can define a transition probability from $z_{t-1}$ to $z_t$ as follows:

\begin{equation}
    q(z_{t}|z_{t-1}) = \bm{v}(z_{t})^{\!\top} \mathbf{Q}_{t} \bm{v}(z_{t-1}),
\end{equation}
where $\bm{v}(z_t) \in \{0, 1\}^{K+2}$ is a one-hot vector of $z_t$. 
Due to the Markov property, a transition from $z_0$ to $z_t$ is similarly written as:
$q(z_{t}|z_{0}) = \bm{v}(z_{t})^{\!\top}\bar{\mathbf{Q}}_{t}\bm{v}(z_{0})$, where $\bar{\mathbf{Q}}_{t}\!=\!\mathbf{Q}_{t}\mathbf{Q}_{t-1}\cdots\mathbf{Q}_{1}$.
Applying the Markov property $q(z_{t-1}|z_t, z_0)=q(z_{t-1}|z_t)$, we can obtain the posterior distribution $q(z_{t-1}|z_{t}, z_{0})$~(Eq.~(5) in \cite{gu2022vector}).

For the reverse process, we compute a conditional distribution $p_{\theta}(\mathbf{z}_{t-1}|\mathbf{z}_t) \in [0,1]^{N \times (K+2)}$.
Categorical variable $\mathbf{z}_{t-1}$ is sampled from this distribution.
As proposed in a previous study~\cite{austin2021structured}, we use the re-parametrization trick and obtain a posterior distribution as $p_{\theta}(\bm{z}_{t-1}|\bm{z}_{t}) \propto \sum_{\tilde{\bm{z}}_{0}}q(\bm{z}_{t-1}|\bm{z}_{t},\tilde{\bm{z}}_{0})~\tilde{p}_{\theta}(\tilde{\bm{z}}_{0}|\bm{z}_{t})$, where $\tilde{p}_{\theta}(\tilde{\bm{z}}_{0}|\bm{z}_{t})$ is a neural network that predicts the noiseless token distribution at $t=0$.
Following previous studies~\cite{austin2021structured, gu2022vector, inoue2023layoutdm}, we employ the hybrid loss of variational lower bound and auxiliary denoising loss.

The design of transition matrix $\mathbf{Q}_t$ is pivotal in defining the corruption process.
The token transition is categorized into three types: (1) keeping the current token, (2) replacing the token with other tokens, and (3) replacing the token with \masktoken{}.
For each, we employ the probabilities ($\alpha_t$, $\beta_t$, $\gamma_t$).
Hence, using $\mathbf{Q'}_t = \alpha_{t} \mathbb{I} + \beta_{t} \mathbf{1}\mathbf{1}^\top \in [0, 1]^{(K+1) \times (K+1)}$, $\mathbf{Q}_t \in [0, 1]^{(K+2) \times (K+2)}$ is defined as:
\begin{equation}
    \mathbf{Q}_{t} = \begin{bmatrix}
    \mathbf{Q'}_{t} & \boldsymbol{0}  \\
    \gamma_{t} \cdots \gamma_{t} & 1 \\
    \end{bmatrix}.
    \label{eq:Q_mask_and_replace}
\end{equation}
The cumulative transition matrix $\bar{\mathbf{Q}}_{t}$ can be computed in the closed form as $\bar{\mathbf{Q}}_{t} \bm{v}(x_0) = \bar{\alpha}_t \bm{v}(x_0) + (\bar{\gamma}_t - \bar{\beta}_t) \bm{v}(K + 2) + \bar{\beta}_t$, where $\bar{\alpha}_t = \prod_{i=1}^{t}\alpha_i$, $\bar{\gamma}_t = 1 - \prod_{i=1}^{t}(1 - \gamma_i)$, $\bar{\beta}_{t, K} = (K+1)\bar{\beta}_t = 1 - \bar{\alpha}_t- \bar{\gamma}_t$, and $\bm{v}(K+2)$ denotes the one-hot representation of \masktoken.
A transition with $\bar{\beta}_t > 0$ introduces a layout inconsistency.
Since the corresponding DDM is trained to correct such mismatches, we expect that it can update erroneous tokens caused in the generation process.
However, in the following section, we will demonstrate that the scheduling of $\bar{\beta}_t$ is suboptimal for layout correction.

\begin{figure*}[t]
    \centering
    \begin{subfigure}{0.49\linewidth}
        \includegraphics[width=\linewidth]{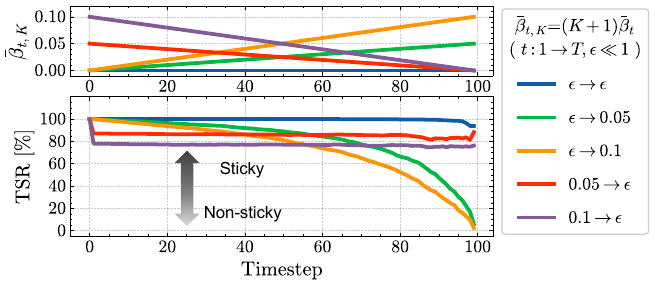}
        \caption{Top: $\bar{\beta}_t$ scheduling for timestep $t$. Bottom: plot of token-sticking-rate~(TSR), representing degree of token matching between $\bm{z}_0$ and $\bm{z}_t$.}
        \label{fig:beta_schedule_and_token_sticking_rate}
    \end{subfigure}
    \hfill
    \begin{subfigure}{0.49\linewidth}
        \includegraphics[width=\linewidth]{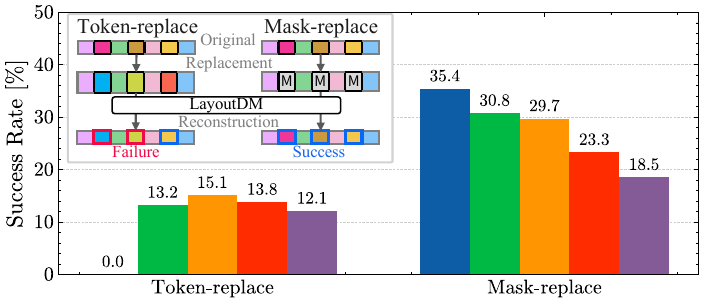}
        \caption{Success rate on token-correction task for Token-replace and Mask-replace strategies. Color of bar corresponds to legend in Fig.~\ref{fig:beta_schedule_and_token_sticking_rate}.}
        \label{fig:token_correction_success_rate}
    \end{subfigure}
    \caption{Results of preliminary experiments on Rico test set~\cite{deka2017rico}. (a)~While $\bar{\beta}_{t,K}=(K+1)\bar{\beta}_t=\epsilon~(\ll 1)$ is affected by token-sticking, $\bar{\beta}_{t,K} > \epsilon$ alleviates it. 
    (b) The results indicate that LayoutDM can restore the original tokens from \masktoken; however, recovery from regular tokens proves challenging. Please refer to Supp. for further results.}
    \label{fig:pre_exp}
\end{figure*}

\subsection{Preliminary: Potential of Token Correction in DDM}\label{sec:preliminary_experiment}

We explore token correction with DDMs, specifically focusing on LayoutDM~\cite{inoue2023layoutdm}, by assessing the impact of the $\bar{\beta}_t$ schedule.
For $\epsilon \ll 1$ and $\bar{\beta}_{t, K} = \epsilon$ for any $t$, $p_{\theta}(\bm{z}_{t-1}|\bm{z}_{t})$ struggles to correct tokens, due to the diffusion process not facilitating token replacement, except for \masktoken.
This limitation is analogous to those seen with parallel decoding methods used in MaskGIT~\cite{chang2022maskgit}.
A possible solution is to increase $\bar{\beta}_{t, K} > \epsilon$ to promote transition between regular tokens.
To verify the effect of $\bar{\beta}_{t, K}$, we compare the token-sticking-rate~(TSR) in the reverse process, which measures the proportion of tokens at $\bm{z}_{0}$ that remain unchanged from $\bm{z}_{t}$.
As depicted in \cref{fig:beta_schedule_and_token_sticking_rate}, $\bar{\beta}_{t, K}=\epsilon$ leads to $\text{TSR}\simeq100\%$ at most $t$, indicating token sticking.
In contrast, when $\bar{\beta}_{t, K} > \epsilon$, the TSR is reduced below $100\%$, indicating that $p_{\theta}(\bm{z}_{t-1}|\bm{z}_{t})$ can update tokens during generation process.

We next evaluate the DDM's error-correction capability by simulating replacements of three randomly selected tokens in a sequence with either \masktoken or other tokens, and then observing the model's ability to restore the original tokens.
These methods are referred to as \textit{Mask-replace} and \textit{Token-replace}, respectively.
In this setup, LayoutDM executes the reverse step from timestep $t=10$ to $1$. 
Our metric is the success rate of token recovery, which is deemed successful if recovery is complete, and we assess this across different $\bar{\beta}_t$ schedules.
\cref{fig:token_correction_success_rate} demonstrates that Token-replace with $\bar{\beta}_{t, K} > \epsilon$ is moderately more effective than when $\bar{\beta}_{t, K} = \epsilon$.
However, Mask-replace exhibits significant improvements over Token-replace.
This finding motivated us to develop \layoutcorrector{}, which resets inharmonious tokens to \masktoken.

\begin{figure*}[t]
    \centering
    \includegraphics[width=\linewidth]{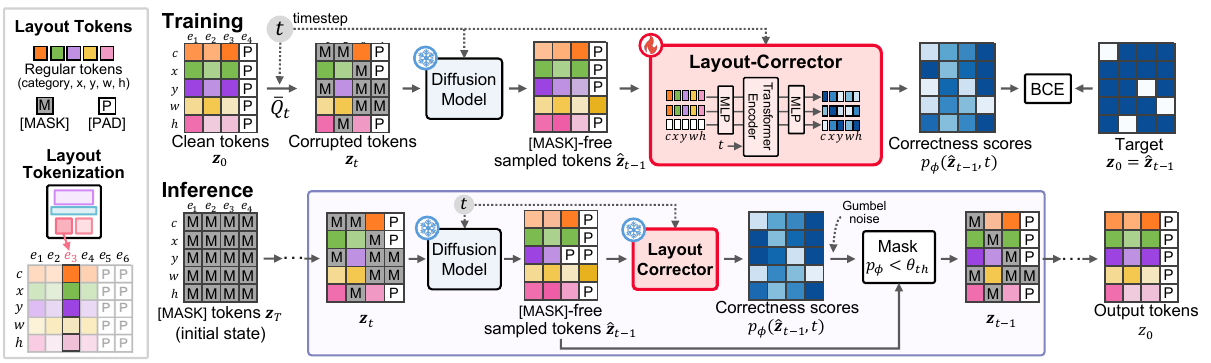}
    \caption{
    The details of \layoutcorrector{}.
    Top: training procedure of \layoutcorrector{}, where the pre-trained DDM is fixed. Bottom: sampling process with \layoutcorrector{}. We execute the generation and correction process in the purple box iteratively.}
    \label{fig:overview}
\end{figure*}

\subsection{Layout-Corrector}\label{sec:corrector}
To enhance the replacement of the erroneous tokens with \masktoken, we introduce \layoutcorrector. 
Functioning as a quality assessor, \layoutcorrector{} evaluates the correctness of each token in a layout during the generation process. 
The tokens with lower correctness scores are replaced with \masktoken, and the updated tokens are fed back into the DDM. 
Therefore, \layoutcorrector{} can explicitly prompt the DDM to modify the erroneous tokens.

\noindent
\textbf{Architecture.}
Evaluating the correctness of each token requires \layoutcorrector{} to consider the relationships between elements in a layout. 
To this end, we use a Transformer~\cite{vaswani2017attention} encoder to capture global contexts, as shown in \cref{fig:overview}.
We first apply a multi-layer perceptron (MLP) to fuse five tokens ($c, x, y, w, h$) of each element, obtaining $N$ element embeddings. These embeddings are processed by a transformer encoder, producing five-channel outputs. Each channel corresponds to the correctness score of each token in the element.
Since the layout elements are order-agnostic, we eliminate positional encoding to avoid unintended biases.

\noindent
\textbf{Training.}
The objective of \layoutcorrector{} is to detect erroneous tokens during the generation process.
To achieve this, we train \layoutcorrector{} as a binary classifier with a pre-trained DDM, which is frozen during training, as shown in \cref{fig:overview}.
Given an original layout $\bm{z}_0$ and $t$, a forward process is applied to obtain a distribution $q(\bm{z}_t|\bm{z}_0)$, and DDM estimates the distribution $p_\theta (\bm{z}_{t-1}|\bm{z}_t)$.
Then, for a \masktoken-free token sequence $\hat{\bm{z}}_{t-1}$ sampled from $p_\theta (\bm{z}_{t-1}|\bm{z}_t)$, \layoutcorrector{} evaluates the correctness score $p_{\phi}(\hat{\bm{z}}_{t-1},t)\in [0, 1]^{5N}$ for each token in $\hat{\bm{z}}_{t-1}$.
Unlike existing assessors~\cite{lezama2022improved, lezama2023discrete}, which are trained to detect tokens in $\hat{\bm{z}}_{t-1}$ that are originally masked in $\bm{z}_t$, we train \layoutcorrector{} to predict whether each token in $\hat{\bm{z}}_{t-1}$ aligns with the corresponding original token in $\bm{z}_0$, as in \cite{clark2020electra}.
It encourages the \layoutcorrector{} to evaluate the correctness of each token directly.
Specifically, we use binary cross-entropy (BCE) loss:
\begin{eqnarray}
    \mathcal{L}_{\mathrm{Corrector}} = \mathrm{BCE}(\bm{m}, p_{\phi}(\hat{\bm{z}}_{t-1}, t)),
\end{eqnarray}
where $m^{(i)}=1$ if $\hat{z}_{t-1}^{(i)} = z_{0}^{(i)}$, otherwise $m^{(i)}=0$.
Through the training, \layoutcorrector{} learns to identify erroneous tokens that disturb the layout harmony.

\subsection{Generating Layout with the \layoutcorrector{}}
\label{subsec:generate_w_cor}

\noindent
\textbf{Unconditional Generation.}
As shown in \cref{fig:overview}, all tokens $\bm{z}_T$ are initialized with \masktoken, and the final output is obtained at $t=0$.
At timestep $t$, a DDM predicts the distribution $p_{\theta}(\bm{z}_{t-1}|\bm{z}_{t})$, from which we sample a \masktoken-free token sequence $\hat{\bm{z}}_{t-1}$.
\layoutcorrector{} then assesses the correctness scores $p_{\phi}(\hat{\bm{z}}_{t-1}, t)$. 
We add Gumbel noise to $p_{\phi}(\hat{\bm{z}}_{t-1}, t)$ to introduce randomness into the token selection. Then, we mask tokens whose scores are lower than a threshold $\theta_{th}$.
Another possible way is to choose tokens with the lowest $5N \cdot \bar{\gamma}_t$ scores, similar to \cite{lezama2022improved,lezama2023discrete}, where $\bar{\gamma}_t$ is the mask ratio at $t$ (see Sec.~\ref{sec:discrete_models}). However, it may mask high-quality tokens when the majority have higher scores, leading to diminished cues for the DDM.
The threshold mitigates this issue by selectively masking only those tokens with lower scores, thus preserving reliable cues for regeneration.

\noindent
\textbf{Conditional Generation.}
\layoutcorrector{} is versatile and can be seamlessly used for various conditional generation tasks without specialized training or fine-tuning.
Given a set of partially known tokens, \eg, element categories or sizes, the goal of conditional generation is to estimate the remaining unknown tokens.
Following LayoutDM~\cite{inoue2023layoutdm}, we utilize the known condition tokens as an initial state for the generation process and maintain these tokens at each $t$. 
When \layoutcorrector{} is applied, a correctness score of 1 is assigned to the conditional tokens. 
In this way, \layoutcorrector{} encourages the DDM to modify erroneous tokens while ensuring that the known tokens are preserved.

\noindent
\textbf{Corrector Scheduling.}
\layoutcorrector{} can be applied at any $t$ during the generation process. 
Unlike existing methods~\cite{lezama2022improved, lezama2023discrete}, which apply the external assessor at every $t$, we selectively apply \layoutcorrector{} at specific timesteps.
Remarkably, alongside LayoutDM~\cite{inoue2023layoutdm}, \layoutcorrector{} enhances generation quality with just three applications, effectively reducing additional forward operations during inference.
Moreover, by adjusting the schedule of \layoutcorrector{}, we can modulate the fidelity-diversity trade-off of the generated layouts.
Specifically, more frequent corrector applications enhance fidelity by removing a larger number of inharmonious tokens, while a more sparse schedule improves diversity.
The experimental section provides a more detailed analysis on the schedules.
\section{Experiments}\label{sec:experiments}

\newcommand{\red}[1]{\textbf{#1}}
\newcommand{\blue}[1]{{#1}}

\begin{table*}[t]
    \centering
    \caption{Performance comparison of baseline models with/without external assessor on unconditional generation.
    \textit{Arch.} represents the architecture of the discrete generative model. 
    Metrics improved by the external module are highlighted in \textbf{bold}. 
    }
    \label{tab:base_plus_cor}
    \resizebox{\textwidth}{!}{
    \begin{tabular}{llccccccccc}
    \toprule
                                       &        & \multicolumn{3}{c}{Rico~\cite{deka2017rico}}             & \multicolumn{3}{c}{Crello~\cite{yamaguchi2021canvasvae}} & \multicolumn{3}{c}{PubLayNet~\cite{zhong2019publaynet}}  \\ \cmidrule(r){3-5} \cmidrule(lr){6-8} \cmidrule(l){9-11}
    Model                              & Arch.  & FID$\downarrow$ & Precision$\uparrow$ & Recall$\uparrow$ & FID$\downarrow$ & Precision$\uparrow$ & Recall$\uparrow$ & FID$\downarrow$ & Precision$\uparrow$ & Recall$\uparrow$ \\ \cmidrule(r){1-1} \cmidrule(lr){2-2} \cmidrule(lr){3-3} \cmidrule(lr){4-4} \cmidrule(lr){5-5} \cmidrule(lr){6-6} \cmidrule(lr){7-7} \cmidrule(lr){8-8} \cmidrule(lr){9-9} \cmidrule(lr){10-10} \cmidrule(l){11-11}
    MaskGIT~\cite{chang2022maskgit}    & Non-AR & 70.37 & 0.793 & 0.437 & 35.32 & 0.802 & 0.376 & 34.23 & 0.587 & 0.460          \\
    ~~+ Token-Critic~\cite{lezama2022improved}             &        & \red{15.65} & \blue{0.682} & \red{0.843} & \red{7.59} & \blue{0.735} & \red{0.815} & \red{17.55} & \blue{0.579} & \red{0.825}     \\
    \rowcolor[HTML]{EFEFEF} 
    ~~\textbf{+ Corrector (ours)} &        & \red{14.40} & \red{0.814} & \red{0.744} & \red{11.17} & \red{0.839} & \red{0.696} & \red{13.74} & \blue{0.501} & \red{0.883}     \\ \cmidrule(r){1-1} \cmidrule(lr){2-2} \cmidrule(lr){3-3} \cmidrule(lr){4-4} \cmidrule(lr){5-5} \cmidrule(lr){6-6} \cmidrule(lr){7-7} \cmidrule(lr){8-8} \cmidrule(lr){9-9} \cmidrule(lr){10-10} \cmidrule(l){11-11}
    VQDiffusion~\cite{gu2022vector}    & DDMs   & 7.83 & 0.716 & 0.907 & 5.57 & 0.740 & 0.884 & 12.38 & 0.567 & 0.925            \\
    ~~+ Token-Critic~\cite{lezama2022improved}             &        & \blue{15.22} & \red{0.842} & \blue{0.731} & \blue{10.05} & \red{0.834} & \blue{0.657} & \blue{17.53} & \red{0.812} & \blue{0.628}     \\
    \rowcolor[HTML]{EFEFEF} 
    ~~\textbf{+ Corrector (ours)} &        & \red{5.29} & \red{0.809} & \blue{0.898} & \red{4.70} & \red{0.793} & \blue{0.842} & \red{9.89} & \red{0.699} & \blue{0.903}    \\ \cmidrule(r){1-1} \cmidrule(lr){2-2} \cmidrule(lr){3-3} \cmidrule(lr){4-4} \cmidrule(lr){5-5} \cmidrule(lr){6-6} \cmidrule(lr){7-7} \cmidrule(lr){8-8} \cmidrule(lr){9-9} \cmidrule(lr){10-10} \cmidrule(l){11-11}
    LayoutDM~\cite{inoue2023layoutdm}  & DDMs   & 6.37 & 0.759 & 0.906 & 5.28 & 0.768 & 0.875 & 13.72 & 0.557 & 0.919             \\
    ~~+ Token-Critic~\cite{lezama2022improved}             &        & \blue{17.97} & \red{0.884} & \blue{0.670} & \blue{9.01} & \red{0.844} & \blue{0.678} & \blue{22.27} & \red{0.836} & \blue{0.582}     \\
    \rowcolor[HTML]{EFEFEF} 
    ~~\textbf{+ Corrector (ours)} & & \red{4.79} & \red{0.811} & \blue{0.891} & \red{4.36} & \red{0.822} & \blue{0.851} & \red{11.85} & \red{0.711} & \blue{0.890} \\
    \bottomrule
    \end{tabular}
    }
\end{table*}

\subsection{Experimental Settings}\label{sec:experiment_settings}

\textbf{Datasets.} 
We evaluated \layoutcorrector{} on the following three challenging layout datasets over different domains:
\textit{Rico}~\cite{deka2017rico} contains user interface designs for mobile applications. It contains 25 element categories such as text, button, and icon.
\textit{Crello}~\cite{yamaguchi2021canvasvae} consists of design templates for various formats, such as social media posts and banner ads.
\textit{PubLayNet}~\cite{zhong2019publaynet} comprises academic papers with 5 categories, such as table, image, and text. 
We follow the dataset splits presented in a previous study~\cite{inoue2023layoutdm} for Rico and PubLayNet and use the official splits for Crello.
We excluded layouts with more than 25 elements as in \cite{inoue2023layoutdm}.

\noindent
\textbf{Evaluation Metrics.}
We used the following evaluation metrics: 
\textit{Fréchet Inception Distance (FID)}~\cite{heusel2017gans} evaluates the similarity between distributions of generated and real data in the feature space using the feature extractor~\cite{kikuchi2021constrained}.
\textit{Alignment (Align.)}~\cite{li2020attribute} measures the alignment of elements in generated layouts.
This metric is normalized by the number of elements, as in \cite{kikuchi2021constrained}.
\textit{Maximum IoU (Max-IoU)}~\cite{kikuchi2021constrained} evaluates the similarity of the elements in bounding boxes of the same category, comparing the generated layouts to the ground truth.
For fidelity and diversity, we used \textit{Precision} and \textit{Recall}~\cite{kynkaanniemi2019improved}.

\noindent
\textbf{Layout-generation Tasks.}
We evaluated our \layoutcorrector{} across three tasks: 
\textit{Unconditional} generates a layout without any constraints. 
\textit{Category $\rightarrow$ size $+$ position (C $\rightarrow$ S $+$ P)} generates a layout given only the category of each element. 
\textit{Category $+$ size $\rightarrow$ position (C $+$ S $\rightarrow$ P)} generates a layout given the category and size of each element.

\noindent
\textbf{Implementation Details.}
We used DDMs, \ie, LayoutDM~\cite{inoue2023layoutdm} and VQDiffusion~\cite{gu2022vector}, as well as a non-AR model \ie, MaskGIT~\cite{chang2022maskgit}, as baseline models, and applied \layoutcorrector{} to them.
Since non-ARs can be understood as a subclass of DDMs as discussed in \cref{sec:related}, we can seamlessly apply \layoutcorrector{} to them.
We used the publicly available pre-trained LayoutDM on Rico and PubLayNet, while we trained other models using LayoutDM implementation.
Unless otherwise specified, the total timesteps $T$ for LayoutDM and VQDiffusion were set to 100, and \layoutcorrector{} was applied at $t=\{10,20,30\}$, leading to a total of 103 forward operations. 
In MaskGIT~\cite{chang2022maskgit}, $T=10$ and \layoutcorrector{} was applied at every $t$.
For the threshold $\theta_{th}$, we set it to 0.7 for LayoutDM and VQDiffusion, and 0.3 for MaskGIT.
To train \layoutcorrector{}, we used AdamW~\cite{kingma2014adam, loshchilov2018decoupled} with an initial learning rate of $5.0\times 10^{-4}$ and $(\beta_1, \beta_2)=(0.9, 0.98)$. 
Refer to supplementary material for more details.

\begin{figure*}[t]
    \centering
    \begin{minipage}{0.43\linewidth}
        \includegraphics[width=\linewidth]{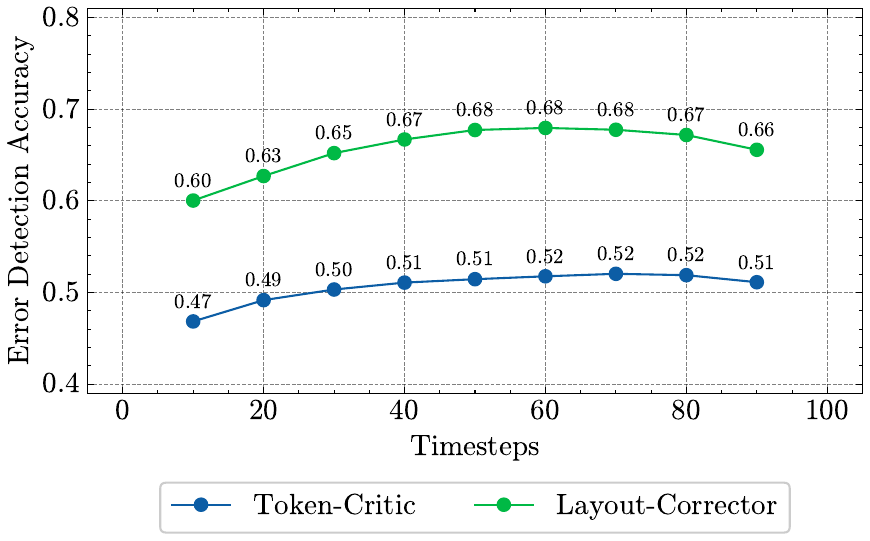}
        \caption{Accuracy of detecting erroneous tokens when three tokens are replaced randomly.}
        \label{fig:error_token_detection}
    \end{minipage}
    \hfill
    \begin{minipage}{0.52\linewidth}
        \includegraphics[width=\linewidth]{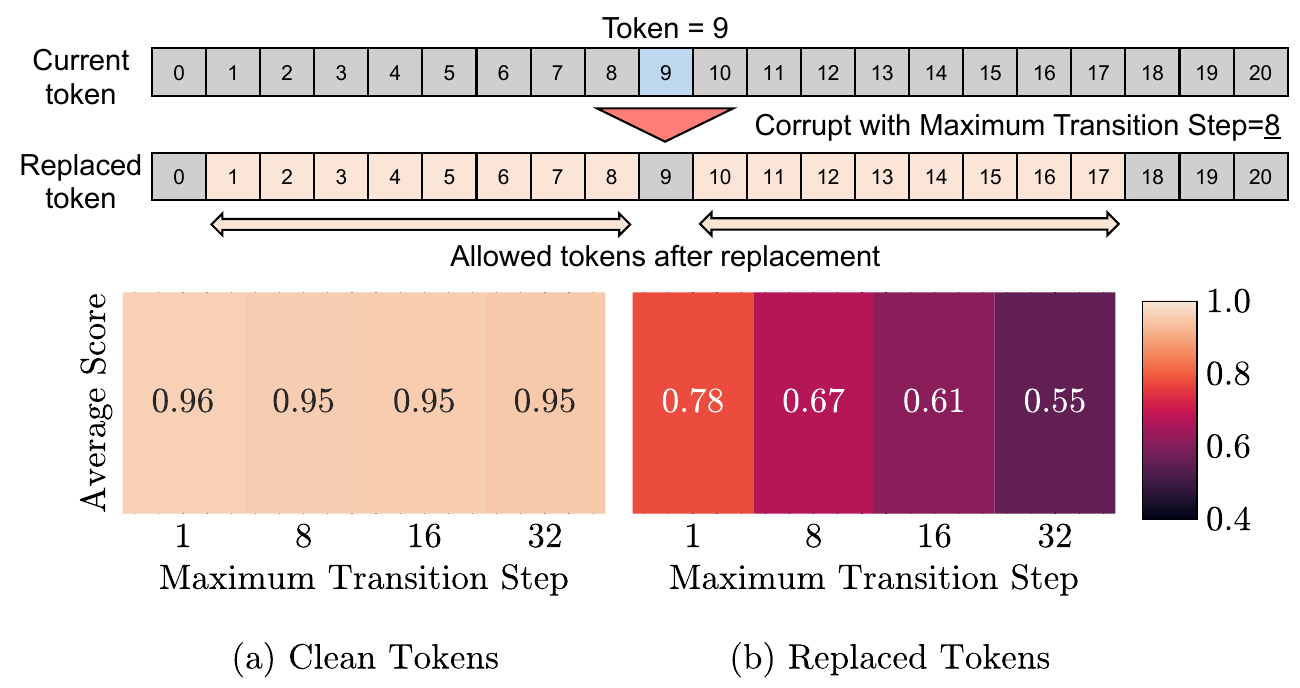}
        \caption{Correctness scores against an extent of layout disruption controlled by various maximum transition steps.}
        \label{fig:corruption_degree_vs_score}
    \end{minipage}
\end{figure*}

\subsection{Effectiveness of Layout-Corrector}\label{sec:corrector_w_baseline}

To evaluate the applicability and effectiveness of \layoutcorrector{} with various discrete generative models, we applied our corrector and Token-Critic~\cite{lezama2022improved} to MaskGIT~\cite{chang2022maskgit}, VQDiffusion~\cite{gu2022vector}, and LayoutDM~\cite{inoue2023layoutdm}. 
The results in \cref{tab:base_plus_cor} show that \layoutcorrector{} consistently improved FID across all tested models, confirming its effectiveness.
In contrast, while Token-Critic~\cite{lezama2022improved} enhanced FID when applied to MaskGIT, its application to VQDiffusion and LayoutDM resulted in diminished performance. These results demonstrate that the direct application of Token-Critic can lead to suboptimal performance in layout generation, highlighting the importance of tailored approaches.
Regarding fidelity and diversity, evaluated using Precision and Recall~\cite{kynkaanniemi2019improved}, we observed different trends between a non-AR model and DDMs.
For MaskGIT, \layoutcorrector{} boosted both diversity and fidelity. The parallel decoding in MaskGIT keeps high-confidence tokens and rejects low-confidence ones. It often leads to stereotypical token patterns, resulting in high fidelity but low diversity.
\layoutcorrector{} resets such patterns while considering the overall harmony, thereby improving diversity without sacrificing fidelity.
In the case of VQDiffusion and LayoutDM, \layoutcorrector{} increased fidelity while maintaining diversity.
While the stochastic nature of DDMs promotes diversity, it can produce low-quality tokens.
\layoutcorrector{} mitigates this issue by resetting these tokens, thereby enhancing fidelity.

\begin{figure*}[t]
    \begin{minipage}[t]{0.46\linewidth}
        \centering
        \includegraphics[width=\linewidth]{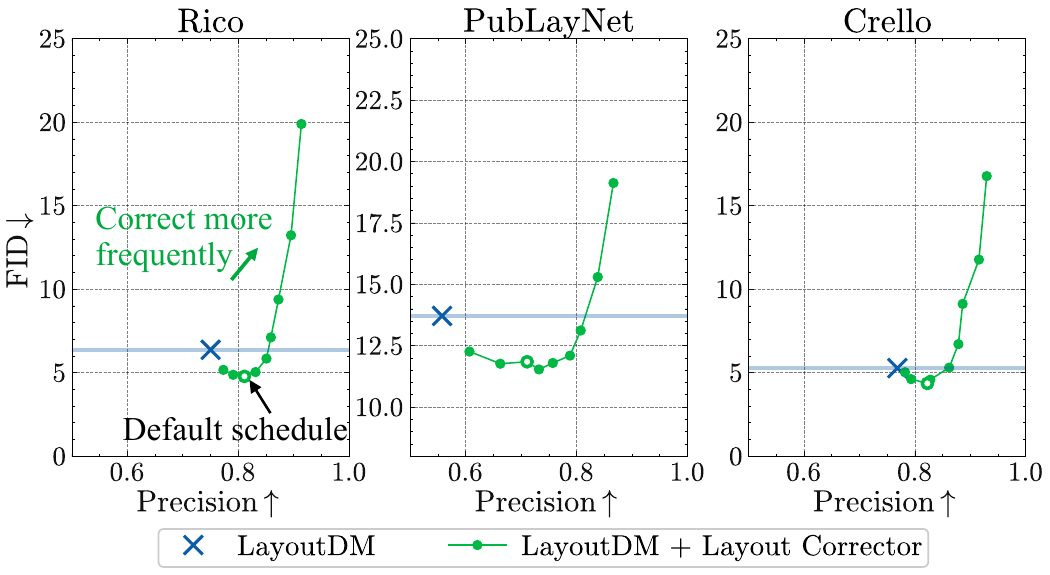}
        \caption{FID-Precision trade-off on unconditional generation on different corrector schedules.}
        \label{fig:fd_tradeoff}
    \end{minipage}
    \hfill
    \begin{minipage}[t]{0.52\linewidth}
        \centering
        \includegraphics[width=\linewidth]{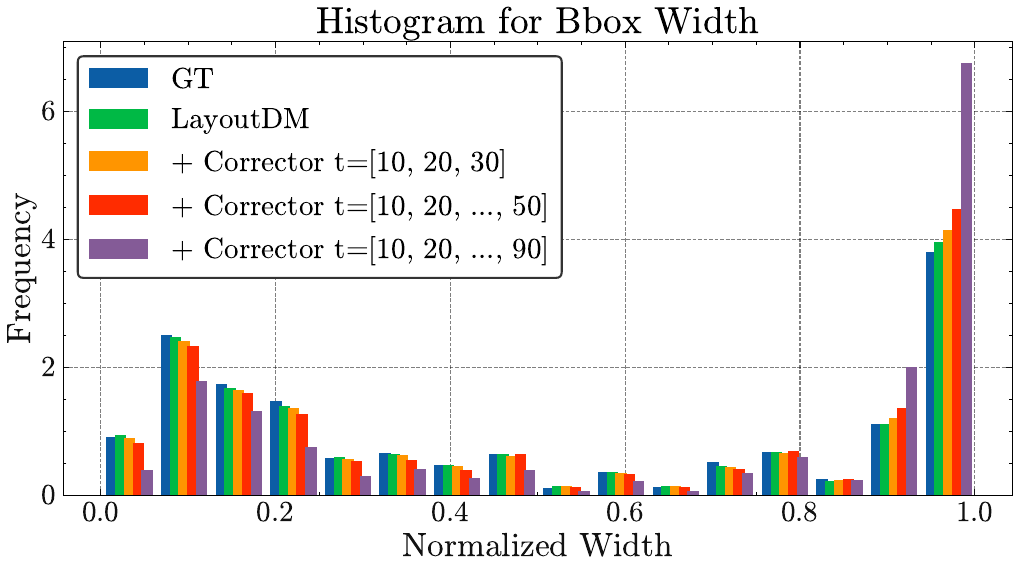}
        \caption{Histogram of the width of bounding boxes on the Rico dataset on different corrector schedules.}
        \label{fig:attr_hist}
    \end{minipage}
\end{figure*}

\subsection{Analysis}\label{sec:corrector_analysis}

\textbf{Intrinsic Evaluation of Corrector.}
We assess \layoutcorrector{}'s ability to detect erroneous tokens. 
To this end, we randomly replace three tokens in the ground truth with alternate ones using the test set of Rico dataset.
The goal is for the corrector to identify these altered tokens.
For \layoutcorrector{} and Token-Critic trained with the same LayoutDM, we evaluate the detection accuracy of these altered tokens, selecting the three tokens with the lowest corrector scores for comparison. 
Fig.~\ref{fig:error_token_detection} shows that \layoutcorrector{} outperforms Token-Critic due to the objective that directly estimates the correctness of tokens, underscoring its effectiveness in layout assessment.

Furthermore, we analyze the correlation between the degree of layout corruption and the correctness scores. 
To modulate the extent of disruption, we limit the maximum transition step for the geometric attributes when replacing. 
Fig.~\ref{fig:corruption_degree_vs_score} depicts the average correctness scores for the three replaced and the other clean tokens within the corrupted layouts.
We observe that a greater deviation from the original token leads to a lower correctness score against clean tokens.
These results suggest that the corrector can measure the degree of discrepancy between the ideal and the actual layouts.

\begin{figure*}[t]
    \begin{minipage}[t]{0.48\linewidth}
        \centering
        \includegraphics[width=\linewidth]{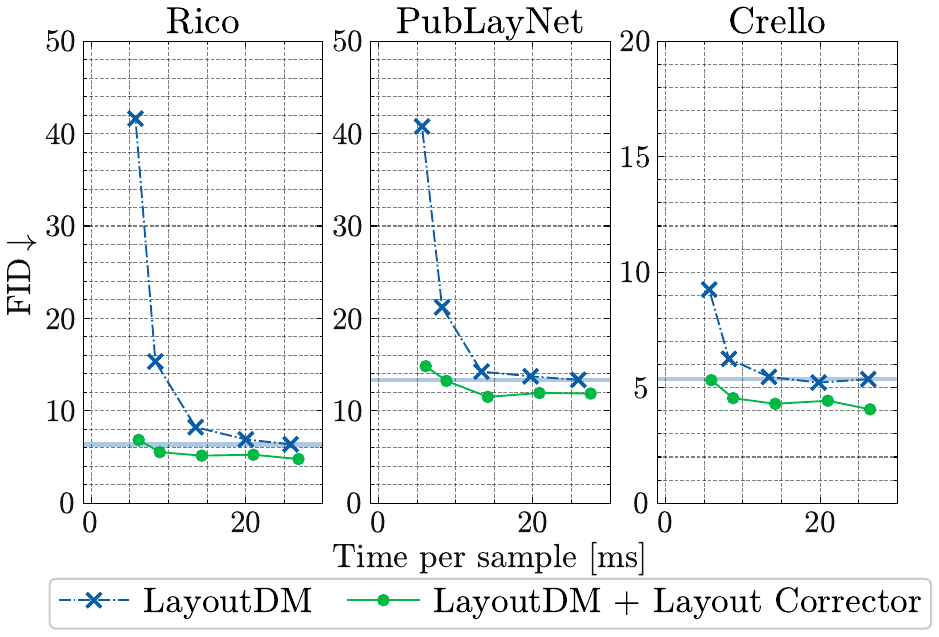}
        \caption{Speed-quality (FID) trade-off on unconditional generation. The \layoutcorrector{} mitigates the quality decline at smaller total timesteps.}
        \label{fig:sq_tradeoff}
    \end{minipage}
    \hfill
    \begin{minipage}[t]{0.48\linewidth}
        \centering
        \includegraphics[width=\linewidth]{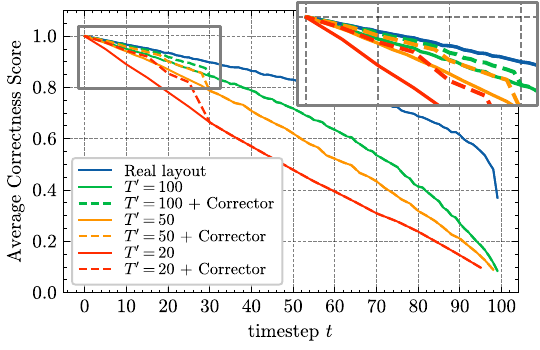}
        \caption{Average correctness scores on intermediate generated layouts of LayoutDM at each $t$ across different total timesteps $T'$ on Rico dataset.}
        \label{fig:sq_corrector_scores}
    \end{minipage}
\end{figure*}

\noindent
\textbf{Corrector Scheduling: Impact on Fidelity and Diversity.}
We applied \layoutcorrector{} to LayoutDM with various schedules.
Specifically, it is applied at $t=[ \{10\} , \{10, 20\}, \ldots, \{10, 20, \ldots, 90\}]$, yielding nine distinct schedules.
\cref{fig:fd_tradeoff} shows the FID-Precision trade-offs with and without \layoutcorrector{}. It can be observed that varying the schedule effectively adjusts the FID-Precision trade-off. More frequent corrector applications enhance fidelity (\ie, Precision), while decreasing diversity (\ie, FID). 
These results align with \layoutcorrector's role of resetting poor-quality tokens. Its frequent use leads to high-fidelity outputs; however, this also reduces the stochastic nature of DDMs, thus diminishing diversity.
Moreover, it shows that our default schedule $t=\{10, 20, 30\}$ yields preferable FID, confirming the effectiveness of our schedule.

To further explore the impact of the corrector schedule, we analyzed the distribution of tokens' width attribute on Rico dataset.
\cref{fig:attr_hist} compares the histograms of real and generated tokens under various corrector schedules.
It reveals that more frequent correction amplifies the frequency trends of the original distribution. Concretely, values that are already common in the real data (\eg, $w=1.0$ in \cref{fig:attr_hist}) become more prevalent, while the occurrence of rarer values further diminishes.
This observation aligns with the trend in \cref{fig:fd_tradeoff}, where frequent correction leads to higher fidelity but at the cost of reduced diversity.

\noindent
\textbf{Speed-Quality Trade-off.}
Since generation speed is crucial in practical applications, we examined the speed-quality trade-off.
To adjust the runtime of LayoutDM, we used the fast-sampling technique~\cite{austin2021structured}, which uses modulated distribution $p_{\theta}(\bm{z}_{t-\Delta}|\bm{z}_{t})$ instead of $p_{\theta}(\bm{z}_{t-1}|\bm{z}_{t})$.
$\Delta$ is a step size, and the total steps are reduced to $T'=T/\Delta$.
Regardless of $\Delta$, \layoutcorrector{} is applied at $t=\{10, 20, 30\}$.
\cref{fig:sq_tradeoff} presents the results on $T'=\{20, 30, 50, 75, 100\}$.
Compared with LayoutDM alone, \layoutcorrector{} improves the FID score with only a minimal increase in runtime, offering a superior trade-off.
While the original LayoutDM's performance significantly degrades with a smaller $T'$,  \layoutcorrector{} effectively mitigates this issue by rectifying the misgenerated tokens, demonstrating the robustness in smaller $T'$.
Notably, \layoutcorrector{} attains a competitive FID to that of the original LayoutDM ($T'=100$) with just $T'=20$.

To further analyze the benefits of  \layoutcorrector{} for smaller $T'$, we examined the correctness scores $p_{\phi}(\hat{\bm{z}}_{t-1},t)$ across different $T'$.
\cref{fig:sq_corrector_scores} presents the average scores for intermediate tokens $\hat{\bm{z}}_{t}$ at each $t$ on $T'=\{20, 50, 100\}$. We include the scores of real layouts for reference.
Note that $p_{\phi}(\hat{\bm{z}}_{t-1},t)$ is affected by the input $t$ because of the training procedure. 
For example, at smaller $t$, the corrupted tokens $\bm{z}_{t}$ become closer to $\bm{z}_{0}$; therefore, most tokens in $\hat{\bm{z}}_{t-1}$ align with $\bm{z}_{0}$. Thus, the corrector learns to predict higher $p_{\phi}(\hat{\bm{z}}_{t-1},t)$ at smaller $t$.
For the original LayoutDM (solid lines), there is a gap between scores at $T'=100$ and smaller $T'$.
As indicated in \cref{fig:corruption_degree_vs_score}, a lower $p_{\phi}(\hat{\bm{z}}_{t-1},t)$ suggests a greater discrepancy, resulting in inferior FID in \cref{fig:sq_tradeoff}.
Conversely, by applying \layoutcorrector{} (dashed lines), low-scored tokens are reset, mitigating the gap in correctness scores at the smaller $T'$. 
It allows LayoutDM to leverage high-quality tokens as clues, effectively avoiding the deterioration in
FID.

\begin{table}[t]

\centering
\caption{Comparison with SoTA models on unconditional task. 
For \textit{Align.}$\rightarrow$, values that more closely match those of real data are preferred, and we scale the values 100× for visibility.
Best and second-best results are in \textbf{bold} and with {\ul underline}, respectively.}
\label{tab:vs_sota_uncond}
\scriptsize{
\begin{tabular}{@{}clcccccc}
\toprule
                                                & \multicolumn{1}{c}{}                                   & \multicolumn{2}{c}{Rico~\cite{deka2017rico}}                                & \multicolumn{2}{c}{Crello~\cite{yamaguchi2021canvasvae}}                       & \multicolumn{2}{c}{PubLayNet~\cite{zhong2019publaynet}}                   \\ \cmidrule(l){3-8} 
\multirow{-2}{*}{Task}                          & \multicolumn{1}{c}{\multirow{-2}{*}{Model}}            & FID$\downarrow$                       & Align.$\rightarrow$                 & FID$\downarrow$                       & Align.$\rightarrow$                    & FID.$\downarrow$                    & Align.$\rightarrow$                 \\ \midrule
                                                & DLT~\cite{levi2023dlt}                                 & {\ul 6.20}                            & 0.386                               & {\ul 4.71}                            & 0.484                                  & \textbf{7.87}                       & \textbf{0.121}                      \\
                                                & LayoutTransformer~\cite{gupta2021layouttransformer}    & 7.63                                  & \textbf{0.068}                      & 5.93                                  & 0.305                                  & 13.90                               & {\ul 0.127}                         \\
                                                & LayoutDM~\cite{inoue2023layoutdm}                      & 6.39                                  & 0.223                               & 5.28                                  & {\ul 0.279}                            & 13.69                               & 0.185                               \\
                                                & \cellcolor[HTML]{EFEFEF}\textbf{LayoutDM + Corrector}  & \cellcolor[HTML]{EFEFEF}\textbf{4.79} & \cellcolor[HTML]{EFEFEF}{\ul 0.167}       & \cellcolor[HTML]{EFEFEF}\textbf{4.36} & \cellcolor[HTML]{EFEFEF}\textbf{0.232} & \cellcolor[HTML]{EFEFEF}{\ul 11.85} & \cellcolor[HTML]{EFEFEF}0.172       \\ \cmidrule{2-8} 
                                                & \multicolumn{1}{c}{\tiny{Large models}}                &                                       &                                     &                                       &                                        &                                     &                                     \\
                                                & LayoutDiffusion~\cite{zhang2023layoutdiffusion}        & \textbf{3.84}                         & \textbf{0.092}                      & 6.61                                  & 0.228                                  & \textbf{7.57}                       & \textbf{0.077}                      \\
                                                & LayoutDM*~\cite{inoue2023layoutdm}                     & 4.93                                  & 0.146                               & {\ul 4.40}                            & \textbf{0.315}                         & 10.92                               & 0.158                               \\
                                                & \cellcolor[HTML]{EFEFEF}\textbf{LayoutDM* + Corrector} & \cellcolor[HTML]{EFEFEF}{\ul 4.23}    & \cellcolor[HTML]{EFEFEF}{\ul 0.127} & \cellcolor[HTML]{EFEFEF}\textbf{4.11} & \cellcolor[HTML]{EFEFEF}{\ul 0.278}    & \cellcolor[HTML]{EFEFEF}{\ul 9.85}  & \cellcolor[HTML]{EFEFEF}{\ul 0.122} \\ \cmidrule{2-8} 
\multirow{-9}{*}{\rotatebox{90}{Unconditional}} & \textit{Real data}                                     & 1.85                                  & 0.109                               & 2.32                                  & 0.338                                  & 6.25                                & 0.0214                              \\ \bottomrule
\end{tabular}
}

\end{table}

\begin{table}[t]

\centering
\caption{Comparison with SoTA models on conditional tasks. 
Best and second-best results are in \textbf{bold} and with {\ul underline}, respectively.}
\scriptsize{
\label{tab:vs_sota_cond}
\begin{tabular}{clcccccc}
    \toprule
                                                         & \multicolumn{1}{c}{}                                   & \multicolumn{2}{c}{Rico~\cite{deka2017rico}}                                   & \multicolumn{2}{c}{Crello~\cite{yamaguchi2021canvasvae}}                       & \multicolumn{2}{c}{PubLayNet~\cite{zhong2019publaynet}}                     \\ \cmidrule{3-8} 
    \multirow{-2}{*}{Task}                               & \multicolumn{1}{c}{\multirow{-2}{*}{Model}}            & FID$\downarrow$                       & Max-IoU$\uparrow$                      & FID$\downarrow$                       & Max-IoU$\uparrow$                      & FID$\downarrow$                       & Max-IoU$\uparrow$                   \\ \midrule
                                                         & LayoutGAN++~\cite{kikuchi2021constrained}              & 6.84                                  & 0.267                                  & -                                     & -                                      & 24.00                                 & 0.263                               \\
                                                         & DLT~\cite{levi2023dlt}                                 & 3.97                                  & \textbf{0.288}                         & 4.29                                  & \textbf{0.212}                         & \textbf{4.30}                         & \textbf{0.345}                      \\
                                                         & LayoutTransformer~\cite{gupta2021layouttransformer}    & 5.57                                  & 0.223                                  & 6.42                                  & {\ul 0.203}                            & 14.10                                 & 0.272                               \\
                                                         & LayoutDM~\cite{inoue2023layoutdm}                      & {\ul 3.51}                            & 0.276                                  & {\ul 4.04}                            & 0.197                                  & 7.94                                  & 0.309                               \\
                                                         & \cellcolor[HTML]{EFEFEF}\textbf{LayoutDM + Corrector}  & \cellcolor[HTML]{EFEFEF}\textbf{2.39} & \cellcolor[HTML]{EFEFEF}{\ul 0.283}    & \cellcolor[HTML]{EFEFEF}\textbf{3.39} & \cellcolor[HTML]{EFEFEF}0.202          & \cellcolor[HTML]{EFEFEF}{\ul 5.84}    & \cellcolor[HTML]{EFEFEF}{\ul 0.319} \\ \cmidrule{2-8} 
                                                         & \multicolumn{1}{c}{\tiny{Large models}}                &                                       &                                        &                                       &                                        &                                       &                                     \\
                                                         & LayoutNUWA~\cite{tang2023layoutnuwa}                   & 2.52                                  & \textbf{0.445}                         & -                                     & -                                      & 6.58                                  & \textbf{0.385}                      \\
                                                         & LayoutDiffusion~\cite{zhang2023layoutdiffusion}        & \textbf{1.13}                         & {\ul 0.357}                            & 4.68                                  & \textbf{0.253}                         & \textbf{3.09}                         & {\ul 0.351}                         \\
                                                         & LayoutDM*~\cite{inoue2023layoutdm}                     & 2.12                                  & 0.302                                  & {\ul 3.04}                            & 0.206                                  & 6.25                                  & 0.322                               \\
    \multirow{-10}{*}{\rotatebox{90}{C$\rightarrow$S+P}} & \cellcolor[HTML]{EFEFEF}\textbf{LayoutDM* + Corrector} & \cellcolor[HTML]{EFEFEF}{\ul 1.71}    & \cellcolor[HTML]{EFEFEF}0.305          & \cellcolor[HTML]{EFEFEF}\textbf{2.84} & \cellcolor[HTML]{EFEFEF}{\ul 0.210}    & \cellcolor[HTML]{EFEFEF}{\ul 5.01}    & \cellcolor[HTML]{EFEFEF}0.329       \\ \midrule
                                                         & LayoutGAN++~\cite{kikuchi2021constrained}              & 6.22                                  & 0.348                                  & -                                     & -                                      & 9.94                                  & 0.342                               \\
                                                         & DLT~\cite{levi2023dlt}                                 & 3.28                                  & 0.385                                  & 3.68                                  & \textbf{0.278}                         & \textbf{1.53}                         & \textbf{0.425}                      \\
                                                         & LayoutTransformer~\cite{gupta2021layouttransformer}    & 3.73                                  & 0.323                                  & 3.87                                  & {\ul 0.258}                            & 16.90                                 & 0.320                               \\
                                                         & LayoutDM~\cite{inoue2023layoutdm}                      & {\ul 2.17}                            & {\ul 0.390}                            & {\ul 3.55}                            & 0.248                                  & 4.22                                  & 0.380                               \\
                                                         & \cellcolor[HTML]{EFEFEF}\textbf{LayoutDM + Corrector}  & \cellcolor[HTML]{EFEFEF}\textbf{1.91} & \cellcolor[HTML]{EFEFEF}\textbf{0.398} & \cellcolor[HTML]{EFEFEF}\textbf{3.32} & \cellcolor[HTML]{EFEFEF}0.253          & \cellcolor[HTML]{EFEFEF}{\ul 2.93}    & \cellcolor[HTML]{EFEFEF}{\ul 0.390} \\ \cmidrule{2-8} 
                                                         & \multicolumn{1}{c}{\tiny{Large models}}                &                                       &                                        &                                       &                                        &                                       &                                     \\
                                                         & LayoutNUWA~\cite{tang2023layoutnuwa}                   & 2.87                                  & \textbf{0.564}                         & -                                     & -                                      & 3.70                                  & \textbf{0.483}                      \\
                                                         & LayoutDM*~\cite{inoue2023layoutdm}                     & {\ul 1.29}                            & 0.460                                  & {\ul 2.71}                            & {\ul 0.269}                            & {\ul 2.69}                            & 0.408                               \\
    \multirow{-9}{*}{\rotatebox{90}{C+S$\rightarrow$P}}  & \cellcolor[HTML]{EFEFEF}\textbf{LayoutDM* + Corrector} & \cellcolor[HTML]{EFEFEF}\textbf{1.22} & \cellcolor[HTML]{EFEFEF}{\ul 0.463}    & \cellcolor[HTML]{EFEFEF}\textbf{2.69} & \cellcolor[HTML]{EFEFEF}\textbf{0.271} & \cellcolor[HTML]{EFEFEF}\textbf{2.05} & \cellcolor[HTML]{EFEFEF}{\ul 0.415} \\ \bottomrule
\end{tabular}
}

\end{table}

{
    \setlength\tabcolsep{1pt}
    \begin{figure*}[t]
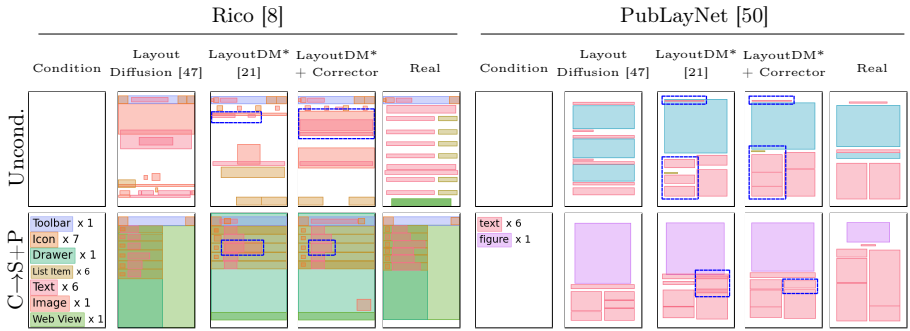

    \centering
    \newcommand{\length}{0.14}
    \newcommand{\figcmd}[3]{
        \frame{\includegraphics[height=\length\hsize]{figures/png/comparison/#1_#2/id_#3_input.png}} &
        \frame{\includegraphics[height=\length\hsize]{figures/png/comparison/#1_#2/id_#3_layout_diffusion.png}} &
        \frame{\includegraphics[height=\length\hsize]{figures/png/comparison/#1_#2/id_#3_layoutdm_12layer.png}} &
        \frame{\includegraphics[height=\length\hsize]{figures/png/comparison/#1_#2/id_#3_layoutdm_12layer_corrector.png}} &
        \frame{\includegraphics[height=\length\hsize]{figures/png/comparison/#1_#2/id_#3_Real.png}}
    }
    \newcommand{\figcmdcwh}[3]{
        \frame{\includegraphics[height=\length\hsize]{figures/png/comparison/#1_#2/id_#3_input.png}} &
        \frame{\includegraphics[height=\length\hsize]{figures/png/comparison/#1_#2/id_#3_layout_former_pp.png}} &
        \frame{\includegraphics[height=\length\hsize]{figures/png/comparison/#1_#2/id_#3_layoutdm_12layer.png}} &
        \frame{\includegraphics[height=\length\hsize]{figures/png/comparison/#1_#2/id_#3_layoutdm_12layer_corrector.png}} &
        \frame{\includegraphics[height=\length\hsize]{figures/png/comparison/#1_#2/id_#3_Real.png}}
    }
    \resizebox{\textwidth}{!}{%
        \begin{tabular}{cccccccccccc}
            & \multicolumn{5}{c}{\small{Rico~\cite{deka2017rico}}} & & \multicolumn{5}{c}{\small{PubLayNet~\cite{zhong2019publaynet}}} \\
            \cmidrule(lr){2-6} \cmidrule(lr){7-12} & 
            \multirow{2}{*}{\tiny{Condition}} &
            \multirow{2}{*}{\tiny{\shortstack{Layout\\Diffusion~\cite{zhang2023layoutdiffusion}}}} & 
            \multirow{2}{*}{\tiny{\shortstack{LayoutDM*\\ \cite{inoue2023layoutdm}}}} &
            \multirow{2}{*}{\tiny{{\shortstack{LayoutDM*\\+ Corrector}}}} &
            \multirow{2}{*}{\tiny{Real}} & 
            &
            \multirow{2}{*}{\tiny{Condition}} &
            \multirow{2}{*}{\tiny{\shortstack{Layout\\Diffusion~\cite{zhang2023layoutdiffusion}}}} & 
            \multirow{2}{*}{\tiny{\shortstack{LayoutDM*\\ \cite{inoue2023layoutdm}}}} &
            \multirow{2}{*}{\tiny{{\shortstack{LayoutDM*\\+ Corrector}}}} &
            \multirow{2}{*}{\tiny{Real}} \\
            & & & & & & & & & & & \\
            \raisebox{2.0\normalbaselineskip}[0pt][0pt]{\rotatebox[origin=c]{90}{Uncond.}} & \figcmd{rico}{uncond}{00005} & \hfill & \figcmd{publaynet}{uncond}{00000} \\
            \raisebox{2.0\normalbaselineskip}[0pt][0pt]{\rotatebox[origin=c]{90}{C$\rightarrow$S+P}} & \figcmd{rico}{c}{00087} & \hfill & \figcmd{publaynet}{c}{00060} \\
        \end{tabular}
    }
    \caption{Comparison of unconditional and conditional generation.
    In \textit{LayoutDM* + Corrector}, \layoutcorrector{} is applied to the same intermediate states of LayoutDM*~\cite{inoue2023layoutdm} at $t=\{10, 20, 30\}$ during generation process. Consequently, generation processes of both methods are identical from $t=100$ to $30$. While LayoutDM* generates unnatural elements, as highlighted in {\color{blue}\dotuline{the blue dashed-line boxes}}, they are rectified by \layoutcorrector{} in \textit{LayoutDM* + Corrector}. Refer to Supp. for more results.}
    \label{fig:comparison_inpainting}
    \end{figure*}
}

\subsection{Comparison with State-of-the-Arts}\label{sec:comp_with_sotas}

\cref{tab:vs_sota_uncond,tab:vs_sota_cond} present comparisons of \layoutcorrector{} with SoTA approaches for unconditional and conditional generation tasks, respectively.
Following \cite{inoue2023layoutdm}, we used FID and Alignment for the unconditional task, and Max IoU alongside FID for the conditional tasks.
For the comparison, we chose the combination of LayoutDM and \layoutcorrector{} since it achieves the best performance in \cref{tab:base_plus_cor}. For comparison with larger models~\cite{tang2023layoutnuwa,zhang2023layoutdiffusion}, we trained enlarged LayoutDM with 12 transformer layers, which is denoted as \textit{LayoutDM*}. Note that the architecture of the \layoutcorrector{} remains the same for \textit{LayoutDM*}.

As shown in \cref{tab:vs_sota_uncond,tab:vs_sota_cond}, our approach achieves superior or competitive FID scores compared with SoTA methods. 
While LayoutDiffusion~\cite{zhang2023layoutdiffusion} outperformed our approach on PubLayNet and Rico, \layoutcorrector{} achieves the best FID on Crello.
On conditional tasks, although LayoutNUWA~\cite{tang2023layoutnuwa} achieved a notably higher Max IoU by using the substantially larger Code Llama 7B~\cite{roziere2023code} and additional pre-training data, our method showed superior FID across the board.
Overall, the consistent high performance of \layoutcorrector{} across a variety of tasks and datasets underscores its versatility and practical utility.

\subsection{Qualitative Evaluation}
\cref{fig:comparison_inpainting} illustrates the layouts generated by different models under two tasks on the Rico and PubLayNet datasets. To demonstrate \layoutcorrector's impact, the results of LayoutDM*~\cite{inoue2023layoutdm} and LayoutDM* + Corrector in the figure share the same intermediate states $\bm{z}_t$ in the generation process until the $t$ when the corrector is first applied.
While the overall structures of their outputs are similar, the enhancements from \layoutcorrector{} are clearly recognizable. For example, in the Rico dataset, \layoutcorrector{} successfully fixes the misalignments in LayoutDM's output.
Moreover, the corrector rearranges overlapping elements in the PubLayNet.
Compared with the other layout-generation methods, our approach consistently generates high-quality layouts, indicating its effectiveness.
\begin{wraptable}[11]{r}[0pt]{0.46\linewidth}
\caption{Ablation study on Rico~\cite{deka2017rico} with unconditional generation.}
\label{tab:ablation}
\centering
\resizebox{\linewidth}{!}{
\begin{tabular}{lcccc}
\toprule
                   & \multicolumn{2}{c}{$T'=100$}      &  \multicolumn{2}{c}{$T'=20$}       \\ \cmidrule(r){2-3} \cmidrule(l){4-5}
                   & FID$\downarrow$           & Align.$\rightarrow$   & FID$\downarrow$           & Align.$\rightarrow$         \\ \cmidrule(r){1-1} \cmidrule(lr){2-2} \cmidrule(lr){3-3} \cmidrule(lr){4-4} \cmidrule(l){5-5}
\rowcolor[HTML]{EFEFEF}Layout-Corrector & \textbf{4.79} & 0.167 &\textbf{ 6.84} & \textbf{0.159} \\ \cmidrule(r){1-1} \cmidrule(lr){2-2} \cmidrule(lr){3-3} \cmidrule(lr){4-4} \cmidrule(l){5-5}
w/o self-attention & 5.54 & 0.151 & 9.11 & 0.213 \\
Mask estimation & 5.22 & \textbf{0.148} & 9.66 & 0.232 \\
Lowest-K & 10.6 & 0.263 & 11.5 & 0.359 \\
Correct at every $t$ & 97.4 & 0.002 & 70.1 & 0.006 \\ 
\cmidrule(r){1-1} \cmidrule(lr){2-2} \cmidrule(lr){3-3} \cmidrule(lr){4-4} \cmidrule(l){5-5}
\textit{Real Data}          & 1.85 & 0.109 & 1.85 & 0.109 \\ \bottomrule
\end{tabular}
}
\end{wraptable}

\subsection{Ablation Study}\label{subsec:ablation}

To validate each strategy in \layoutcorrector{}, we trained and tested the corrector with LayoutDM~\cite{inoue2023layoutdm} under the following configurations. 
In \textit{w/o self-attention}, the corrector uses only MLPs, lacking the ability to consider the global harmony of the layout. 
In \textit{Mask estimation}, as in \cite{lezama2022improved}, the corrector predicts whether each token was originally masked rather than estimating if it aligns with the original one. 
In \textit{Lowest-K}, we mask tokens with the lowest $5N \cdot \bar{\gamma}_t$ scores instead of using the threshold.
In \textit{Correct at every $t$}, the corrector is applied at every $t$ during the reverse process.
Except for this setting, the corrector was applied at $t=\{10, 20, 30\}$.

\cref{tab:ablation} shows the results in unconditional generation for $T'=\{100,20\}$.
\layoutcorrector{} achieved the best FID across different $T'$. In contrast, \textit{w/o self-attention} resulted in inferior performance, showing the importance of capturing the relationship between elements.
The results also demonstrate that the differences between \layoutcorrector{} and existing modules, \ie, Token-Critic~\cite{lezama2022improved} and DPC~\cite{lezama2023discrete}, contribute to higher performance. As described in \cref{sec:method}, the primary distinctions are (1) the training objective, (2) introducing threshold, and (3) selective scheduling. The results of \textit{Mask estimation}, \textit{Lowest-K}, and \textit{Correcting at every $t$} indicate the effectiveness of each modification, respectively.

\section{Conclusion}\label{sec:conclusion}
We introduced \layoutcorrector, a novel module working with a DDM-based layout-generation model.
Our preliminary experiments highlighted (1) the token-sticking problem with DDMs and (2) the importance of masking mis-generated tokens to correct them.
Based on these insights, we design \layoutcorrector{} to assess the correctness score of each token and replace the tokens with low scores with \masktoken, guiding the generative model to correct these tokens.
Our experiments showed that \layoutcorrector{} enhances the generation quality of various generative models on various tasks. 
Additionally, we have shown that it effectively controls the fidelity-diversity trade-off through its application schedule and mitigates the performance decline associated with fast sampling.

\noindent
\textbf{Limitations and Future Work.}
While \layoutcorrector{} adds marginal runtime, it increases memory usage and total parameter count.
Our future work will aim to incorporate layout-specific mechanisms, e.g., element-relation embeddings~\cite{hui2023unifying}, to improve \layoutcorrector's layout understanding capabilities.

\section*{Acknowledgment}
We thank Dr. Kent Fujiwara and Dr. Hirokatsu Kataoka for their invaluable feedback and constructive suggestions on this paper.

\bibliographystyle{splncs04}
\bibliography{references}

\clearpage 
\appendix

\section{Details of Token Refinement Task}\label{supp:potential_token_refinement}
In \cref{sec:preliminary_experiment}, we conducted preliminary experiments to evaluate the token refinement capability of DDMs.
We present the experimental setup and results in more detail.

\subsection{Transition Probability Design}
We use LayoutDM~\cite{inoue2023layoutdm} as the representative of DDMs.
The default setting of $\bar{\beta}_{t,K} = (K+1)\bar{\beta}_{t}$ in LayoutDM is not exactly zero but is sufficiently close.
As a baseline, we set $\bar{\beta}_{t,K} = \epsilon$ for any timestep $t$, where $\epsilon$ equals to $10^{-6}$.
In this setting, the diffusion process primarily induces transitions from regular tokens to \masktoken, and transitions between regular tokens rarely occur. 
Therefore, we can not expect corrections of errors in regular tokens during the generation process.
Setting a large value for $\bar{\beta}_{t,K}$ is expected to facilitate transitions between regular tokens during the diffusion process, allowing the corresponding denoising model $p_{\theta}(\bm{z}_{t-1}|\bm{z}_{t})$ to acquire the capability to correct regular tokens. 
To verify the effect of $\bar{\beta}_{t,K}$ schedule, we consider schedules for $\bar{\beta}_{t,K}$ based on two guidelines. 
The first involves assigning high $\bar{\beta}_{t,K}$ values later in the diffusion process, while the second involves high $\bar{\beta}_{t,K}$ values earlier. 
A detailed schedule, including $\bar{\alpha}_t$ and $\bar{\gamma}_t$, is shown in \cref{figure:supp_trans-prob-schedule}. 
Here, we adopt a linear scheduling for timesteps.

\subsection{Impact of Transition Schedules on FID}
\begin{figure*}[t]
    \centering
    \begin{subfigure}{0.49\linewidth}
        \includegraphics[width=\linewidth]{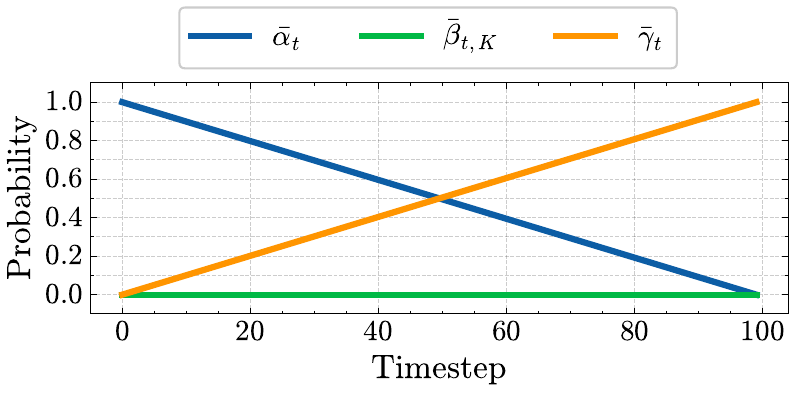}
        \caption{$\bar{\beta}_{t,K}: \epsilon \rightarrow \epsilon$}
        \label{figure:schedule_beta0to0}
    \end{subfigure}

    \vspace{0.5cm}
    \begin{subfigure}{0.49\linewidth}
        \includegraphics[width=\linewidth]{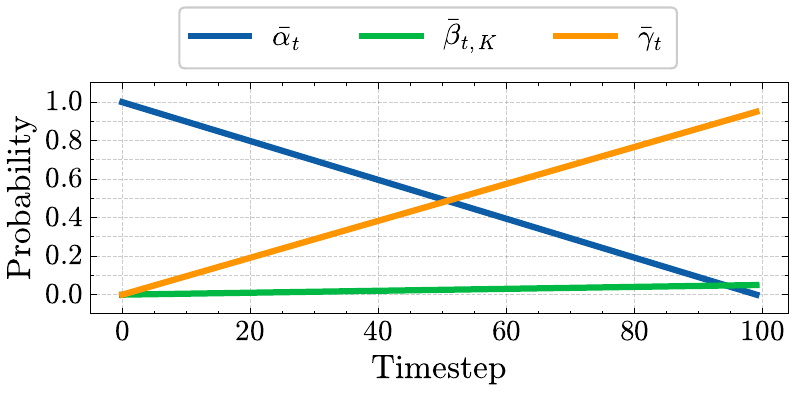}
        \caption{$\bar{\beta}_{t,K}: \epsilon \rightarrow 0.05$}
        \label{figure:schedule_beta0to005}
    \end{subfigure}
    \hfill
    \begin{subfigure}{0.49\linewidth}
        \includegraphics[width=\linewidth]{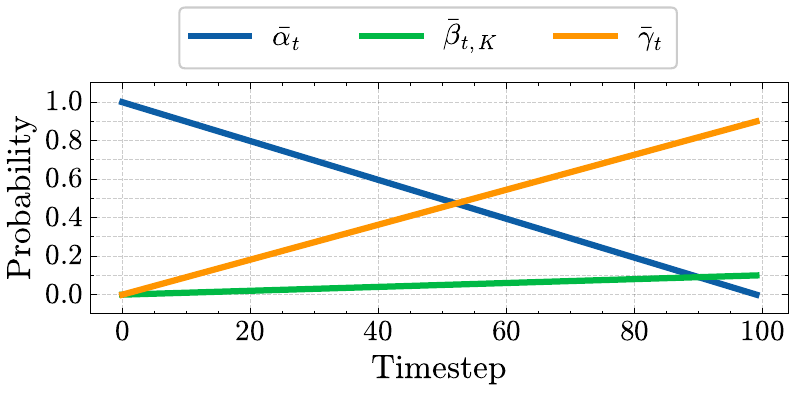}
        \caption{$\bar{\beta}_{t,K}: \epsilon \rightarrow 0.1$}
        \label{figure:schedule_beta0to01}
    \end{subfigure}

    \vspace{0.5cm}
    \begin{subfigure}{0.49\linewidth}
        \includegraphics[width=\linewidth]{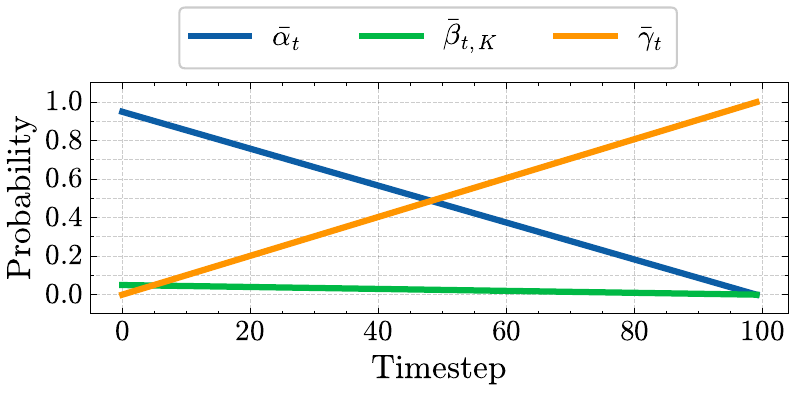}
        \caption{$\bar{\beta}_{t,K}: 0.05 \rightarrow \epsilon$}
        \label{figure:schedule_beta005to0}
    \end{subfigure}
    \hfill
    \begin{subfigure}{0.49\linewidth}
        \includegraphics[width=\linewidth]{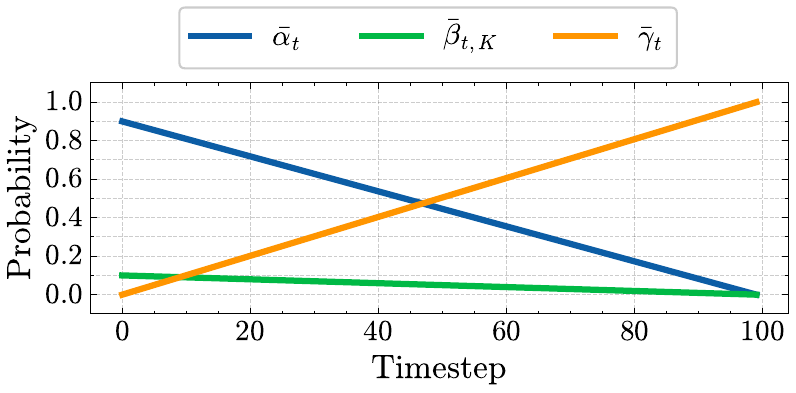}
        \caption{$\bar{\beta}_{t,K}: 0.1 \rightarrow \epsilon$}
        \label{figure:schedule_beta01to0}
    \end{subfigure}

    \caption{Scheduling of transition probabilities for preliminary experiments. \cref{figure:schedule_beta0to0} illustrates the baseline schedule used in LayoutDM, where $\bar{\beta}_{t,K}$ is approximately zero at any timestep. 
    \cref{figure:schedule_beta0to005} and \cref{figure:schedule_beta0to01} demonstrate the schedules that introduce transitions between regular tokens in the later stages of the diffusion process. Conversely, the schedules of \cref{figure:schedule_beta005to0} and \cref{figure:schedule_beta01to0} promote transitions between them in the early stage of the diffusion process.}
    \label{figure:supp_trans-prob-schedule}
\end{figure*}

In \cref{tab:supp_beta-schedule_fid}, we report the results of FID for each schedule depicted in \cref{figure:supp_trans-prob-schedule}. 
When $\bar{\beta}_{t,K}$ increases from $\epsilon$ to 0.05 or 0.1 with the timestep $t$, the performance is comparable to the baseline for the unconditional generation task; however, we observe degradation in the conditional generation tasks. 
Conversely, when $\bar{\beta}_{t,K}$ decreases from 0.05 or 0.1 to $\epsilon$, the performance is inferior to the baseline for both unconditional and conditional tasks.

Regarding the degradation in conditional generation tasks, we hypothesize that it stems from the condition gap between training and inference time.
Training is conducted in an unconditional manner, where, especially for $\bar{\beta}_{t,K} > \epsilon$, the model learns to restore the original layout while correcting substitutions of regular tokens. 
On the other hand, in conditional settings, the model is expected to preserve the conditioned regular tokens, leading to the discrepancy between the training and inference phases.
When $\bar{\beta}_{t,K} = \epsilon$, substitutions between regular tokens rarely occur, which means that conditioning on regular tokens does not negatively impact the generation process.

Additionally, applying high $\bar{\beta}_{t,K}$ values in the earlier timesteps leads to poor FID in the unconditional setting. 
This schedule causes rapid replacements of regular tokens, indicated by $\bar{\alpha}_t < 1$, as observed in \cref{figure:schedule_beta005to0} and \cref{figure:schedule_beta01to0}. 
The results imply that it is necessary to design a schedule for $\bar{\alpha}_t$ that starts at 1.0 when $t=0$ and gradually decreases as $t$ increases, reflecting the fundamental concept of the discrete diffusion process.

\begin{table}[t]
\centering
\caption{FID scores of LayoutDM for various $\bar{\beta}_{t,K}$ schedules. The best and second-best results are highlighted in \textbf{bold} and with {\ul underline}, respectively.}

\small{
\begin{tabular}{@{}cccc@{}}
\toprule
                                & \multicolumn{3}{c}{FID$\downarrow$}                      \\ \cmidrule(l){2-4} 
$\bar{\beta}_{t,K}$ schedule    & Unconditional    & C$\rightarrow$S+P & C+S$\rightarrow$P \\ \cmidrule(r){1-1} \cmidrule(lr){2-2} \cmidrule(lr){3-3} \cmidrule(l){4-4}
$\epsilon \rightarrow \epsilon$ & 6.37             & \textbf{3.51}     & \textbf{2.17}     \\
$\epsilon \rightarrow 0.05$     & \textbf{6.22}    & \underline{4.03}  & \underline{4.38}  \\
$\epsilon \rightarrow 0.1$      & \underline{6.29} & 4.68              & 5.69              \\
$0.05 \rightarrow \epsilon$     & 7.98             & 5.21              & 5.11              \\
$0.1 \rightarrow \epsilon$      & 10.71            & 8.00              & 7.96              \\ \bottomrule
\end{tabular}
}
\label{tab:supp_beta-schedule_fid}
\end{table}

\section{More Detailed Experimental Setup}\label{supp:setting}

In this section, we describe the experimental setup in detail in addition to the description in \cref{sec:experiment_settings}.

\subsection{Datasets}

We provide a more detailed explanation of the benchmark datasets used for evaluation, focusing particularly on how the datasets are divided and their respective sample numbers.

\begin{itemize}
    \item \textbf{Rico}~\cite{deka2017rico}: We follow the dataset split in \cite{inoue2023layoutdm}, resulting in 35,851 / 2,109 / 4,218 samples for train, validation, and test set. 
    \item \textbf{PubLayNet}~\cite{zhong2019publaynet}: We use the dataset split in \cite{inoue2023layoutdm}, resulting in 315,757 / 16,619 / 11,142 samples for train, validation, and test splits. 
    \item \textbf{Crello}~\cite{yamaguchi2021canvasvae}: While the dataset provides various attributes for each element, such as opacity, color, and image data, we only utilize category, position, and size. We use the official splits, which result in 18,714 / 2,316 / 2,331 samples for train, validation, and test set, respectively.
\end{itemize}

\begin{figure*}[t]
    \centering
    \includegraphics[width=\linewidth]{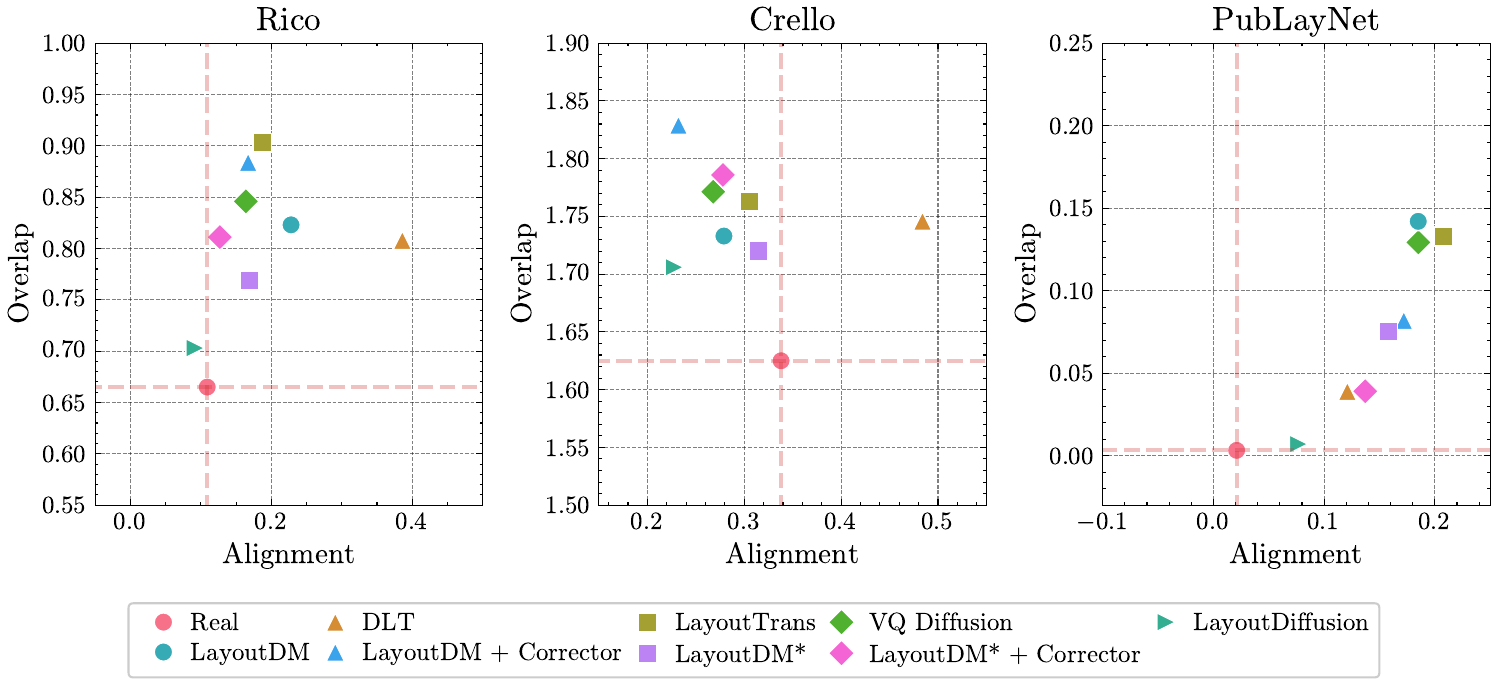}
    \caption{Alignment and overlap~\cite{kikuchi2021constrained} scores across various methods and real data on three datasets. Alignment score is scaled by $100\times$ for visibility.}
    \label{fig:supp_scatter_align_overlap}
\end{figure*}

\subsection{Implementation Details}

\noindent
\textbf{Model Architecture.}
Our \layoutcorrector{} employs a 4-layer Transformer Encoder with 8 multi-heads. For Token-Critic~\cite{lezama2022improved} in Table~\cref{tab:base_plus_cor}, we used the same architecture as \layoutcorrector{}.
For LayoutDM~\cite{inoue2023layoutdm}, VQDiffusion~\cite{gu2022vector}, and MaskGIT~\cite{chang2022maskgit} experiments, we utilized the official implementation of LayoutDM.\footnote{\url{https://github.com/CyberAgentAILab/layout-dm}} 
For LayoutDM* in \cref{sec:comp_with_sotas}, we used a 12-layer Transformer with 12 multi-heads to obtain the same model size as LayoutDiffusion~\cite{zhang2023layoutdiffusion}. Note that we used the same \layoutcorrector{} architecture with a 4-layer Transformer for LayoutDM*.
For LayoutDiffusion~\cite{zhang2023layoutdiffusion}, we used the official implementation\footnote{\url{https://github.com/microsoft/LayoutGeneration/tree/main/LayoutDiffusion}}.

\noindent
\textbf{Training.}
We employed the shared pre-trained LayoutDM models on Rico and PubLayNet datasets. 
For other models, including \layoutcorrector{}, we followed the training configuration of LayoutDM, using AdamW optimizer~\cite{kingma2014adam,loshchilov2018decoupled} with an initial learning rate of $5.0 \times 10^{-4}$, $(\beta_1, \beta_2)=(0.9,0.98)$, and batch size of 64.
The number of training epochs varied according to the dataset: 20 for PubLayNet, 50 for Rico, and 75 for Crello.
For LayoutDiffusion~\cite{zhang2023layoutdiffusion}, since the dataset configuration (\ie, the maximum number of elements in a layout) in the official implementation is different from our setting, we trained the model from scratch instead of using the official checkpoints.

\section{Quantitative Evaluation}\label{supp:quantitative}

In this section, we present additional quantitative evaluation results, including the effectiveness of \layoutcorrector{} on conditional generation and the results of Alignment and Overlap metrics.

\subsection{Effectiveness of \layoutcorrector{} on Conditional Generation}
\begin{table*}[t]
    \caption{
    Performance comparison of baseline models with/without external assessor on conditional generation.
    \textit{Arch.} represents the architecture of the discrete generative model. 
    Metrics improved by the external module are highlighted in \textbf{bold}. 
    }
    \label{tab:baseline_plus_corrector_conditional_generation}
    \begin{subtable}{\linewidth}
            \centering
    \resizebox{\textwidth}{!}{%
    \begin{tabular}{llccccccccc}
    \toprule
                                       &        & \multicolumn{3}{c}{Rico~\cite{deka2017rico}}             & \multicolumn{3}{c}{Crello~\cite{yamaguchi2021canvasvae}} & \multicolumn{3}{c}{PubLayNet~\cite{zhong2019publaynet}}  \\ \cmidrule(r){3-5} \cmidrule(lr){6-8} \cmidrule(l){9-11}
    Model                              & Arch.  & FID$\downarrow$ & Precision$\uparrow$ & Recall$\uparrow$ & FID$\downarrow$ & Precision$\uparrow$ & Recall$\uparrow$ & FID$\downarrow$ & Precision$\uparrow$ & Recall$\uparrow$ \\ \cmidrule(r){1-1} \cmidrule(lr){2-2} \cmidrule(lr){3-3} \cmidrule(lr){4-4} \cmidrule(lr){5-5} \cmidrule(lr){6-6} \cmidrule(lr){7-7} \cmidrule(lr){8-8} \cmidrule(lr){9-9} \cmidrule(lr){10-10} \cmidrule(l){11-11}
    MaskGIT~\cite{chang2022maskgit}    & Non-AR & 30.25 & 0.759 & 0.526 & 31.03 & 0.821 & 0.456 & 16.62 & 0.498 & 0.801         \\
    ~~+ Token-Critic~\cite{lezama2022improved}             &        & \textbf{10.93} & 0.734 & \textbf{0.817} & \textbf{5.85} & 0.759 & \textbf{0.821} & \textbf{8.07} & \textbf{0.679} & \textbf{0.854}     \\
    \rowcolor[HTML]{EFEFEF} 
    ~~\textbf{+ Corrector (ours)} &        & \textbf{7.78} & \textbf{0.814} & \textbf{0.795} & \textbf{6.53} & \textbf{0.843} & \textbf{0.789} & \textbf{7.86} & \textbf{0.503} & \textbf{0.937}     \\ \cmidrule(r){1-1} \cmidrule(lr){2-2} \cmidrule(lr){3-3} \cmidrule(lr){4-4} \cmidrule(lr){5-5} \cmidrule(lr){6-6} \cmidrule(lr){7-7} \cmidrule(lr){8-8} \cmidrule(lr){9-9} \cmidrule(lr){10-10} \cmidrule(l){11-11}
    VQDiffusion~\cite{gu2022vector}    & DDMs   & 4.01 & 0.750 & 0.877 & 3.98 & 0.757 & 0.874 & 7.57 & 0.595 & 0.942           \\
    ~~+ Token-Critic~\cite{lezama2022improved}             &        & \textbf{2.89} & \textbf{0.828} & 0.836 & 4.82 & \textbf{0.802} & 0.829 & \textbf{5.96} & \textbf{0.789} & 0.827     \\
    \rowcolor[HTML]{EFEFEF} 
    ~~\textbf{+ Corrector (ours)} &        & \textbf{2.53} & \textbf{0.790} & \textbf{0.878} & \textbf{3.63} & \textbf{0.791} & 0.834 & \textbf{5.61} & \textbf{0.678} & 0.932    \\ \cmidrule(r){1-1} \cmidrule(lr){2-2} \cmidrule(lr){3-3} \cmidrule(lr){4-4} \cmidrule(lr){5-5} \cmidrule(lr){6-6} \cmidrule(lr){7-7} \cmidrule(lr){8-8} \cmidrule(lr){9-9} \cmidrule(lr){10-10} \cmidrule(l){11-11}
    LayoutDM~\cite{inoue2023layoutdm}  & DDMs   & 3.51 & 0.768 & 0.899 & 4.04 & 0.759 & 0.876 & 7.94 & 0.549 & 0.939            \\
    ~~+ Token-Critic~\cite{lezama2022improved}             &        & \textbf{3.15} & \textbf{0.842} & 0.846 & 4.43 & \textbf{0.822} & 0.816 & \textbf{6.51} & \textbf{0.806} & 0.819     \\
    \rowcolor[HTML]{EFEFEF} 
    ~~\textbf{+ Corrector (ours)} & & \textbf{2.39} & \textbf{0.808} & \textbf{0.905} & \textbf{3.39} & \textbf{0.797} & 0.855 & \textbf{5.84} & \textbf{0.660} & 0.933 \\
    \bottomrule
    \end{tabular}%
    }
    \caption{C $\rightarrow$ S $+$ P task}
    \label{tab:base_plus_cor_category_to_size_plus_position}

    \end{subtable}
    \\ \\
    \begin{subtable}{\linewidth}
        \centering
    \resizebox{\textwidth}{!}{%
    \begin{tabular}{llccccccccc}
    \toprule
                                       &        & \multicolumn{3}{c}{Rico~\cite{deka2017rico}}             & \multicolumn{3}{c}{Crello~\cite{yamaguchi2021canvasvae}} & \multicolumn{3}{c}{PubLayNet~\cite{zhong2019publaynet}}  \\ \cmidrule(r){3-5} \cmidrule(lr){6-8} \cmidrule(l){9-11}
    Model                              & Arch.  & FID$\downarrow$ & Precision$\uparrow$ & Recall$\uparrow$ & FID$\downarrow$ & Precision$\uparrow$ & Recall$\uparrow$ & FID$\downarrow$ & Precision$\uparrow$ & Recall$\uparrow$ \\ \cmidrule(r){1-1} \cmidrule(lr){2-2} \cmidrule(lr){3-3} \cmidrule(lr){4-4} \cmidrule(lr){5-5} \cmidrule(lr){6-6} \cmidrule(lr){7-7} \cmidrule(lr){8-8} \cmidrule(lr){9-9} \cmidrule(lr){10-10} \cmidrule(l){11-11}
    MaskGIT~\cite{chang2022maskgit}    & Non-AR & 8.15 & 0.821 & 0.840 & 9.59 & 0.822 & 0.741 & 5.05 & 0.584 & 0.905          \\
    ~~+ Token-Critic~\cite{lezama2022improved}             &        & \textbf{4.51} & 0.797 & \textbf{0.905} & \textbf{4.68} & 0.771 & \textbf{0.871} & \textbf{3.83} & \textbf{0.630} & \textbf{0.917}     \\
    \rowcolor[HTML]{EFEFEF} 
    ~~\textbf{+ Corrector (ours)} &        & \textbf{3.61} & \textbf{0.825} & \textbf{0.894} & \textbf{4.26} & \textbf{0.826} & \textbf{0.842} & \textbf{3.97} & \textbf{0.607} & \textbf{0.934}     \\ \cmidrule(r){1-1} \cmidrule(lr){2-2} \cmidrule(lr){3-3} \cmidrule(lr){4-4} \cmidrule(lr){5-5} \cmidrule(lr){6-6} \cmidrule(lr){7-7} \cmidrule(lr){8-8} \cmidrule(lr){9-9} \cmidrule(lr){10-10} \cmidrule(l){11-11}
    VQDiffusion~\cite{gu2022vector}    & DDMs   & 2.37 & 0.828 & 0.929 & 3.89 & 0.779 & 0.878 & 4.05 & 0.612 & 0.949            \\
    ~~+ Token-Critic~\cite{lezama2022improved}             &        & \textbf{2.24} & \textbf{0.845} & 0.926 & 3.99 & \textbf{0.787} & \textbf{0.881} & \textbf{2.58} & \textbf{0.724} & 0.927     \\
    \rowcolor[HTML]{EFEFEF} 
    ~~\textbf{+ Corrector (ours)} &        & \textbf{2.02} & \textbf{0.845} & 0.921 & \textbf{3.46} & \textbf{0.813} & 0.876 & \textbf{2.72} & \textbf{0.679} & 0.935    \\ \cmidrule(r){1-1} \cmidrule(lr){2-2} \cmidrule(lr){3-3} \cmidrule(lr){4-4} \cmidrule(lr){5-5} \cmidrule(lr){6-6} \cmidrule(lr){7-7} \cmidrule(lr){8-8} \cmidrule(lr){9-9} \cmidrule(lr){10-10} \cmidrule(l){11-11}
    LayoutDM~\cite{inoue2023layoutdm}  & DDMs   & 2.17 & 0.844 & 0.928 & 3.55 & 0.800 & 0.885 & 4.22 & 0.587 & 0.941             \\
    ~~+ Token-Critic~\cite{lezama2022improved}             &        & \textbf{2.06} & \textbf{0.860} & 0.912 & 3.57 & \textbf{0.803} & \textbf{0.888} & \textbf{2.60} & \textbf{0.712} & 0.925     \\
    \rowcolor[HTML]{EFEFEF} 
    ~~\textbf{+ Corrector (ours)} & & \textbf{1.91} & \textbf{0.856} & 0.922 & \textbf{3.32} & \textbf{0.808} & 0.882 & \textbf{2.93} & \textbf{0.667} & 0.936 \\
    \bottomrule
    \end{tabular}%
    }
    \caption{C $+$ S $\rightarrow$ P task}
    \label{tab:base_plus_cor_category_plus_size_to_position}

    \end{subtable}
\end{table*}

\cref{tab:baseline_plus_corrector_conditional_generation} shows a comparison of the performance of Token-Critic~\cite{lezama2022improved} and \layoutcorrector{} on conditional generation when applied to three baseline models~(\ie., MaskGIT, VQDiffusion, and LayoutDM). \layoutcorrector{} constantly improves the FID scores of the baseline models.

\subsection{Alignment and Overlap}
\cref{fig:supp_scatter_align_overlap} shows the relationship between Alignment and Overlap~\cite{kikuchi2021constrained} on three datasets. 
We also show the scores of real data for reference. 
When compared with the baseline of LayoutDM, \layoutcorrector{} reduces Alignment on three datasets. Regarding Overlap, the score is increased by applying \layoutcorrector{} on Rico and Crello datasets, while it is reduced on PubLayNet.
While those hand-crafted metrics express a quality for intuitive visual appearance, as seen in \cref{subsec:supp_post_process}, they do not necessarily correlate with improvements in the higher-order generative quality represented by the FID score.

\section{Qualitative Evaluation}

\label{supp:qualitative}

In this section, we present additional qualitative results, including additional visualization of the generation results, visualization of the generation process, fidelity-diversity trade-off of the generation results, and failure cases.

\subsection{Additional Results}

We report additional qualitative results for three datasets, including Rico, Crello, and PubLaynet.
\cref{fig:sup_comparison_unconditional_rico},
\cref{fig:sup_comparison_unconditional_crello}, and
\cref{fig:sup_comparison_unconditional_publaynet} show the samples of unconditional generation.
\cref{fig:sup_comparison_c_rico},
\cref{fig:sup_comparison_c_crello}, and
\cref{fig:sup_comparison_c_publaynet} show the samples of C$\rightarrow$P+S task.
\cref{fig:sup_comparison_cwh_rico}, 
\cref{fig:sup_comparison_cwh_crello}, and
\cref{fig:sup_comparison_cwh_publaynet} show the samples of C+S$\rightarrow$P task.
To demonstrate the diversity, we show eight samples for unconditional generation. For conditional generation, we show four samples for each conditional input.

\subsection{Visualization of the Generation Process}
\begin{figure*}[ht]

    \centering
    \begin{subfigure}[b]{0.48\linewidth}
        \includegraphics[width=\linewidth]{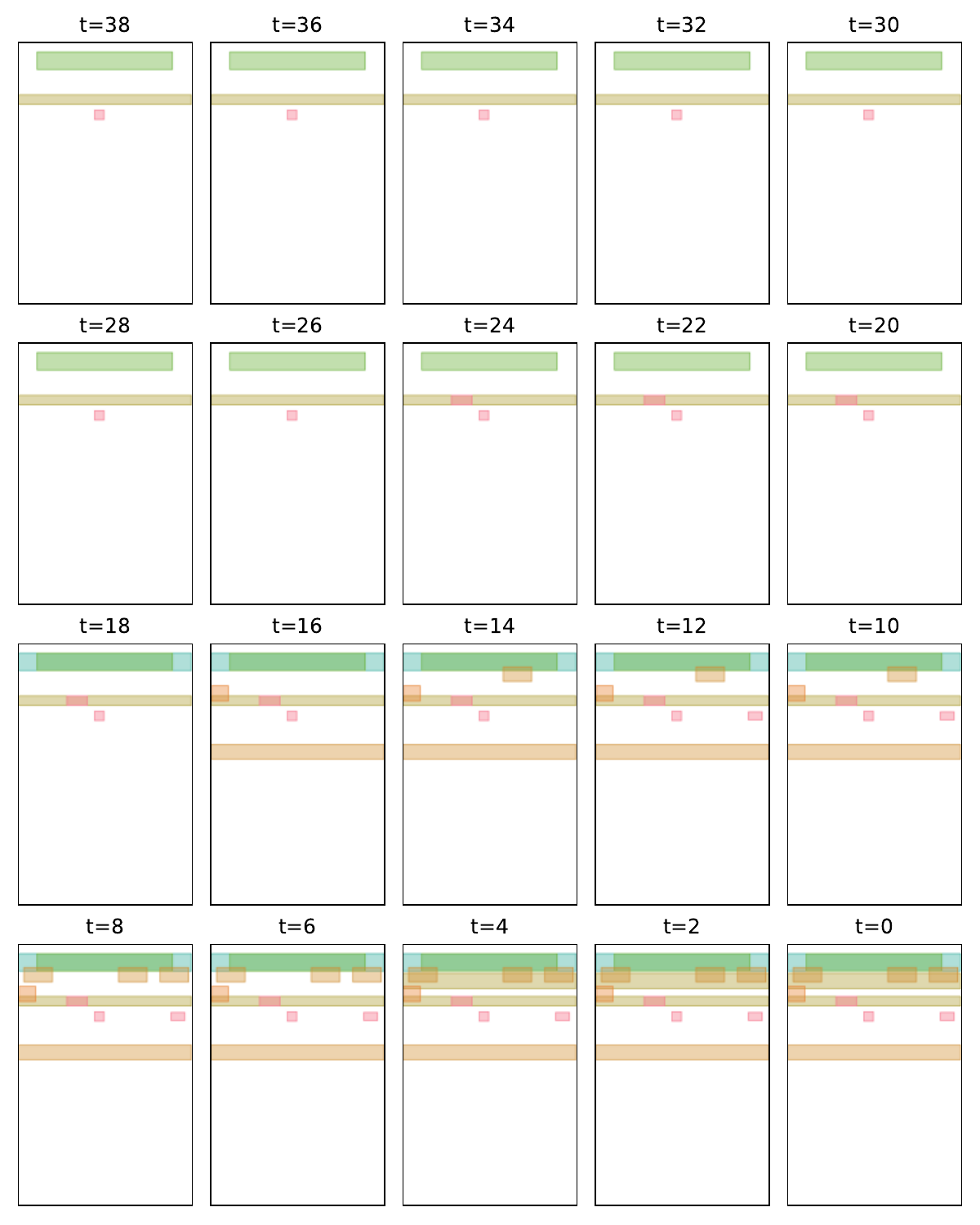}
        \caption{Example1: LayoutDM}
        \label{figure:supp_gen-process-rico_example1_layoutdm}
    \end{subfigure}
    \hfill
    \begin{subfigure}[b]{0.48\linewidth}
        \includegraphics[width=\linewidth]{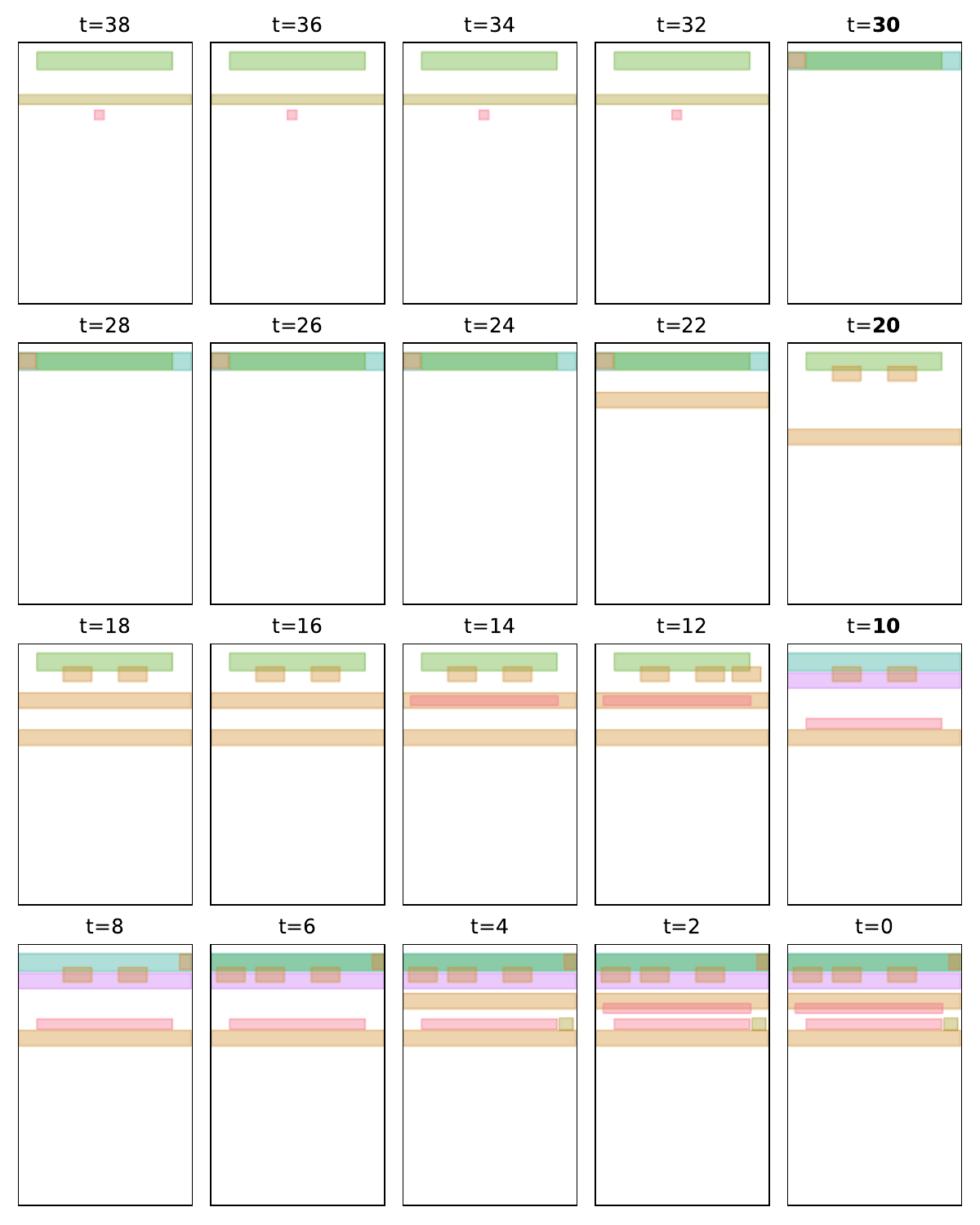}
        \caption{Example1: LayoutDM + Corrector}
        \label{figure:supp_gen-process-rico_example1_corrector}
    \end{subfigure}
    \vspace{10pt}

    \begin{subfigure}[b]{0.48\linewidth}
        \includegraphics[width=\linewidth]{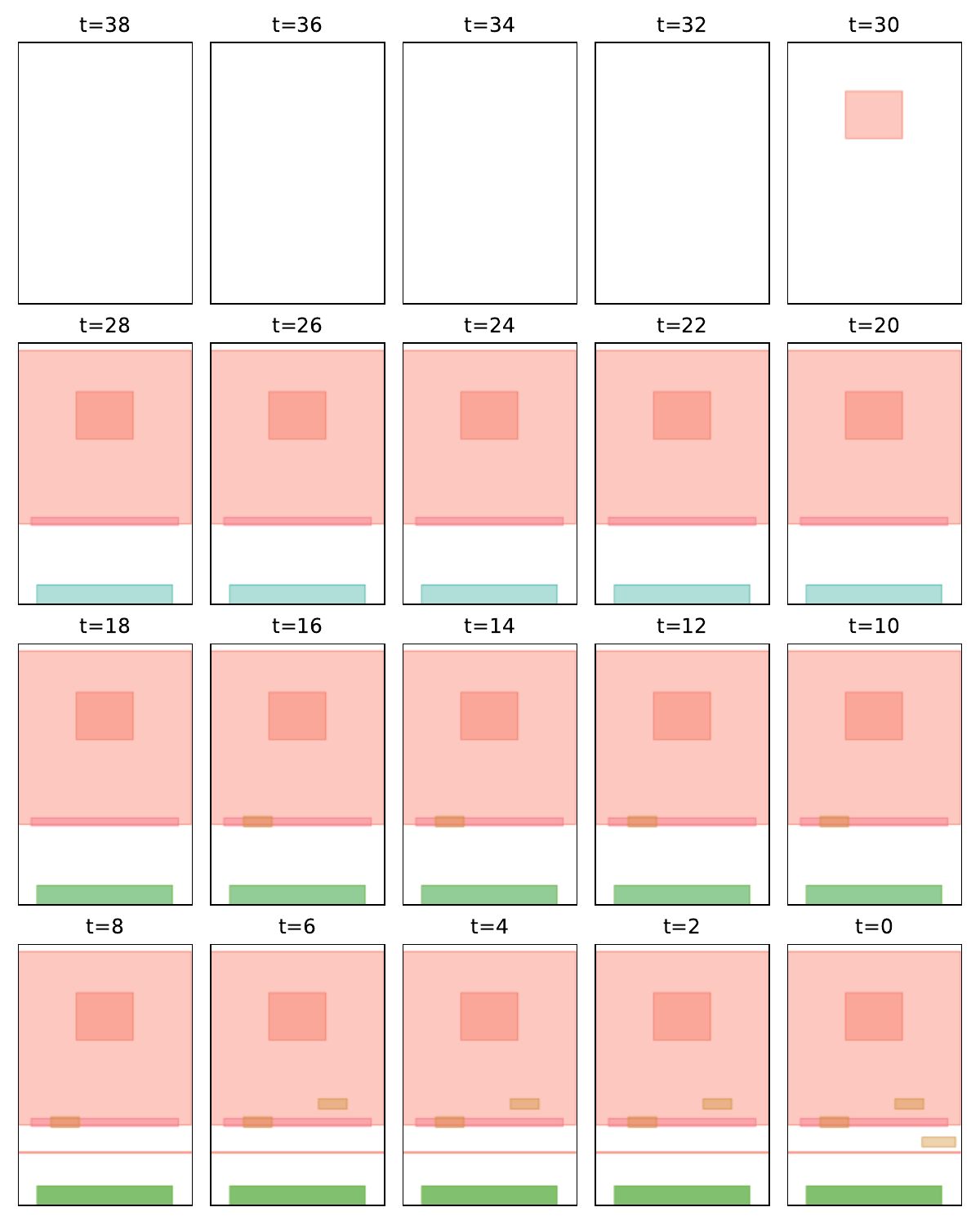}
        \caption{Example2: LayoutDM}
        \label{figure:supp_gen-process-rico_example2_layoutdm}
    \end{subfigure}
    \hfill
    \begin{subfigure}[b]{0.48\linewidth}
        \includegraphics[width=\linewidth]{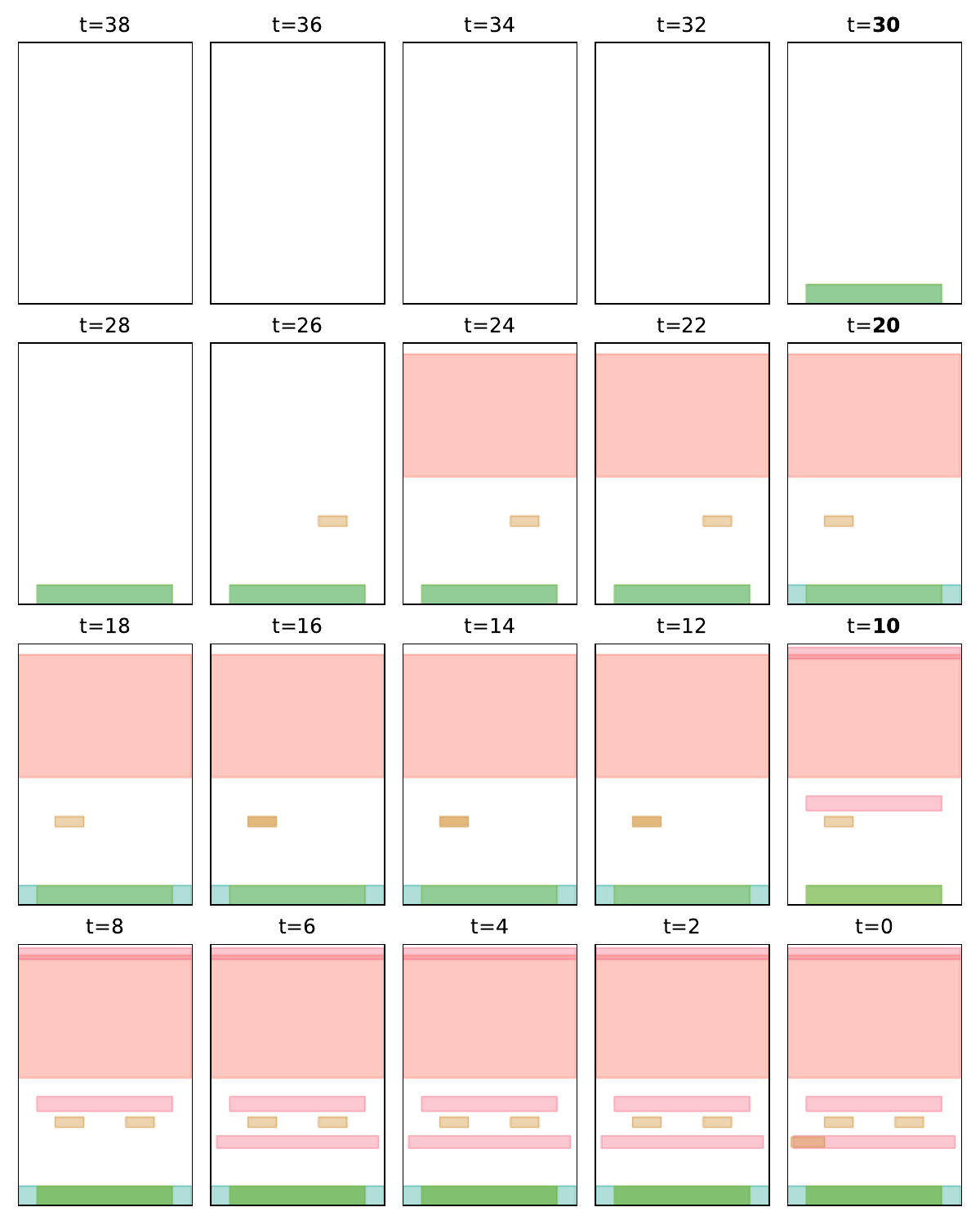}
        \caption{Example2: LayoutDM + Corrector}
        \label{figure:supp_gen-process-rico_example2_corrector}
    \end{subfigure}
    \caption{Comparison of unconditional generation process for Rico. Left:~the results of LayoutDM. Right:~the results of LayoutDM in conjunction with \layoutcorrector{}. 
    The timestep is denoted at the top of each layout visualization, and the timesteps when the corrector is applied are highlighted by \textbf{bold} in \cref{figure:supp_gen-process-rico_example1_corrector} and \cref{figure:supp_gen-process-rico_example2_corrector}.}
    \label{figure:supp_gen-process-rico}
\end{figure*}

\begin{figure*}[ht]

    \centering
    \begin{subfigure}[b]{0.45\linewidth}
        \includegraphics[width=\linewidth]{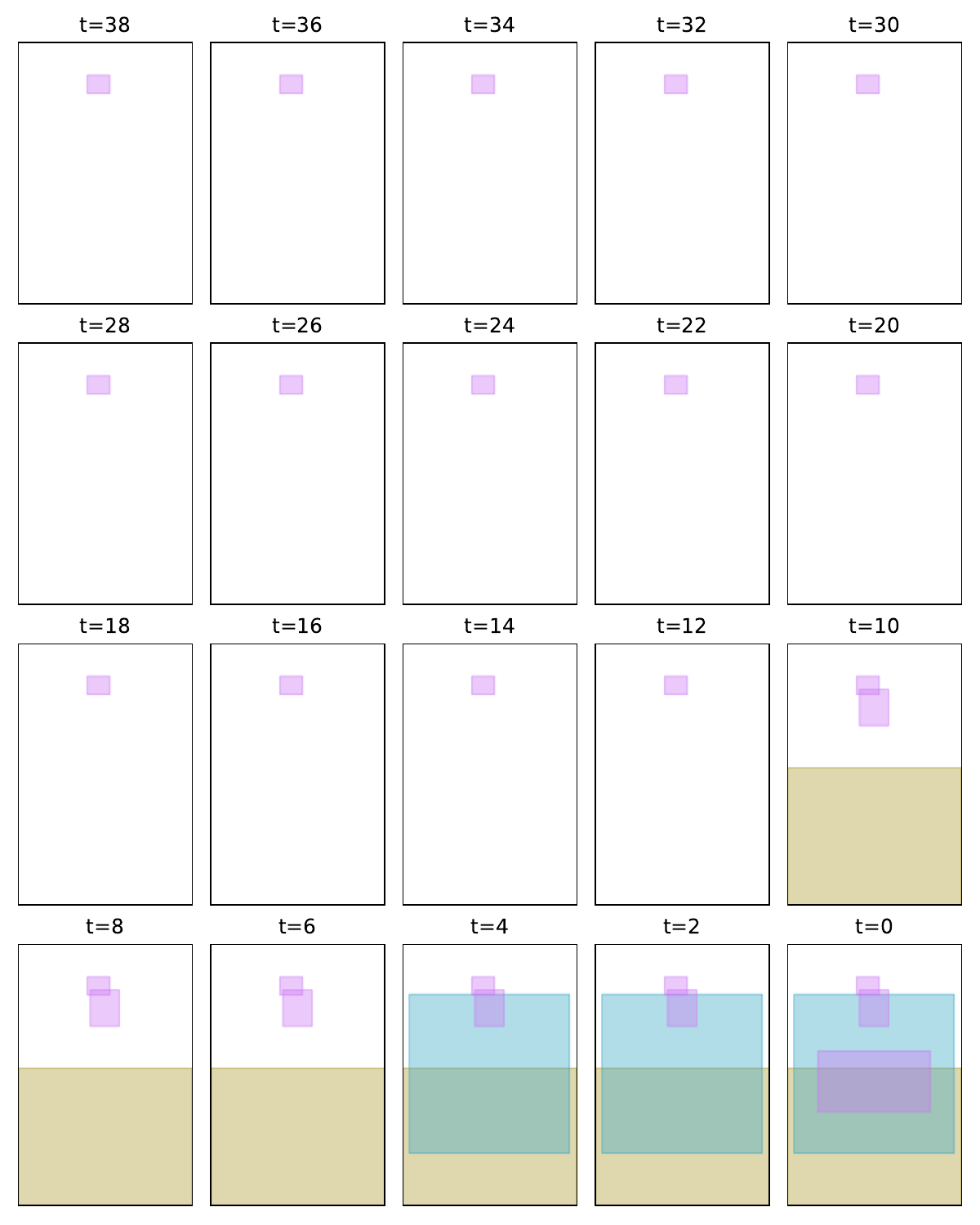}
        \caption{Example1: LayoutDM}
        \label{figure:supp_gen-process-crello_example1_original}
    \end{subfigure}
    \hfill
    \begin{subfigure}[b]{0.45\linewidth}
        \includegraphics[width=\linewidth]{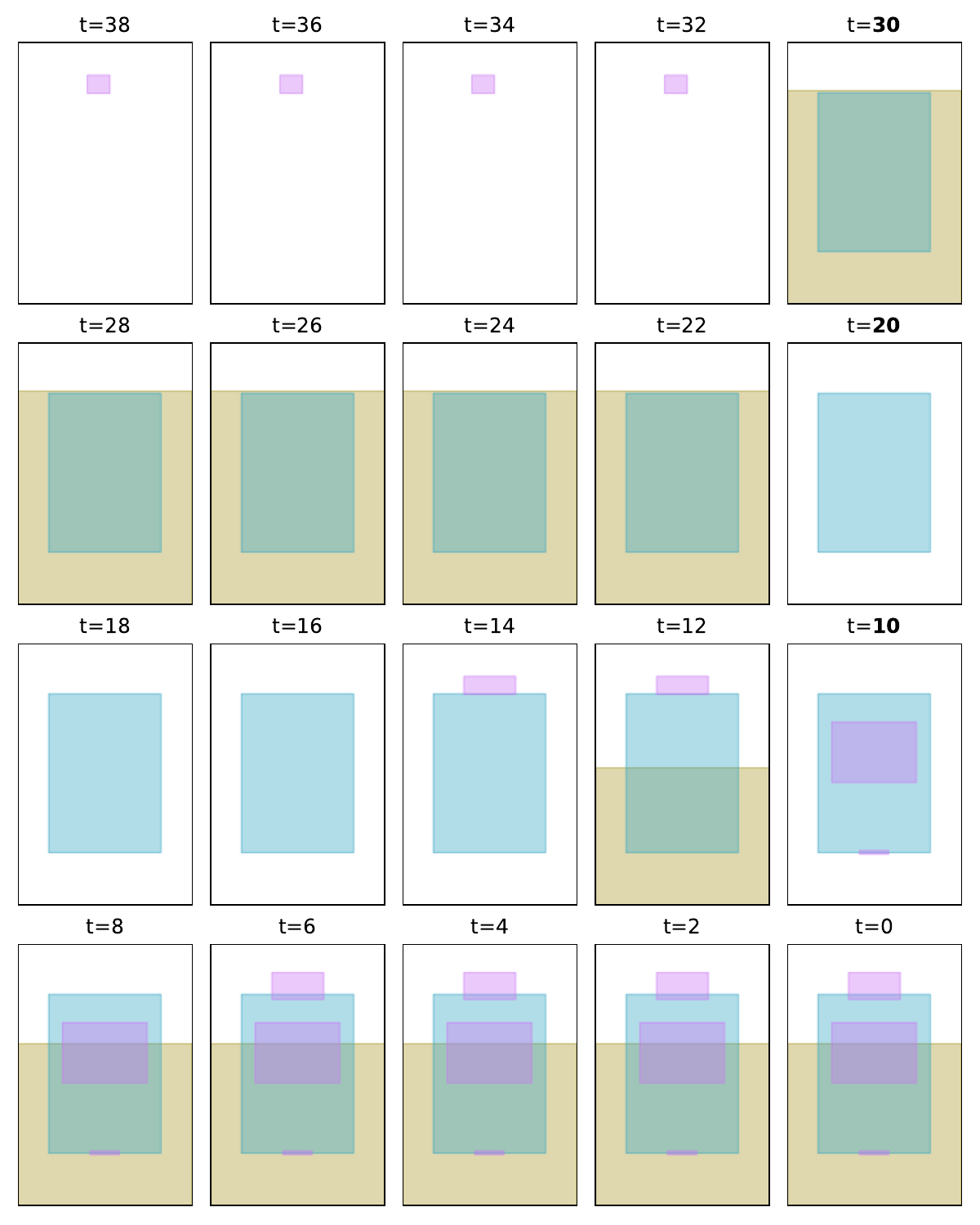}
        \caption{Example1: LayoutDM + Corrector}
        \label{figure:supp_gen-process-crello_example1_corrector}
    \end{subfigure}
    \vspace{10pt}

    \begin{subfigure}[b]{0.45\linewidth}
        \includegraphics[width=\linewidth]{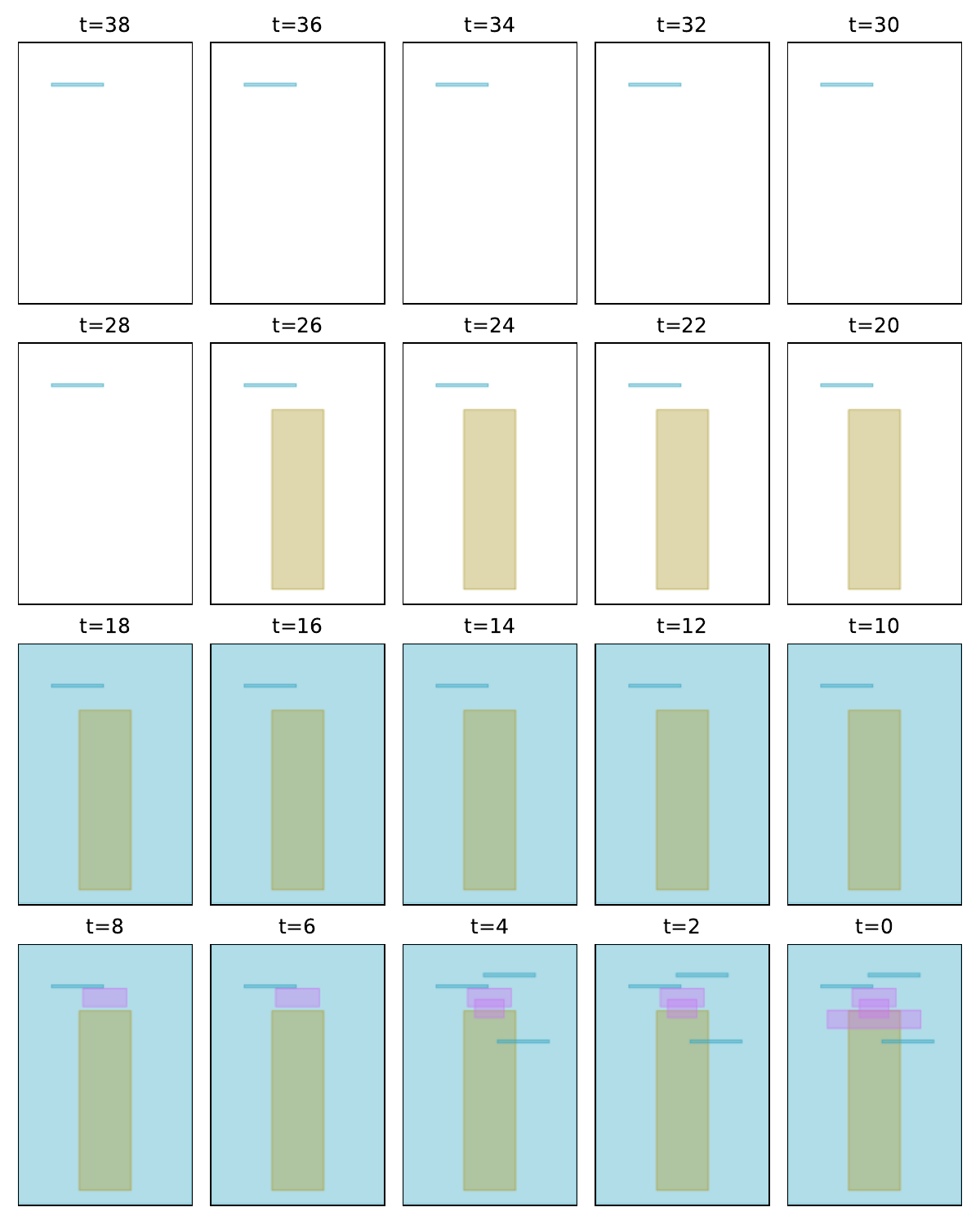}
        \caption{Example2: LayoutDM}
        \label{figure:supp_gen-process-crello_example2_original}
    \end{subfigure}
    \hfill
    \begin{subfigure}[b]{0.45\linewidth}
        \includegraphics[width=\linewidth]{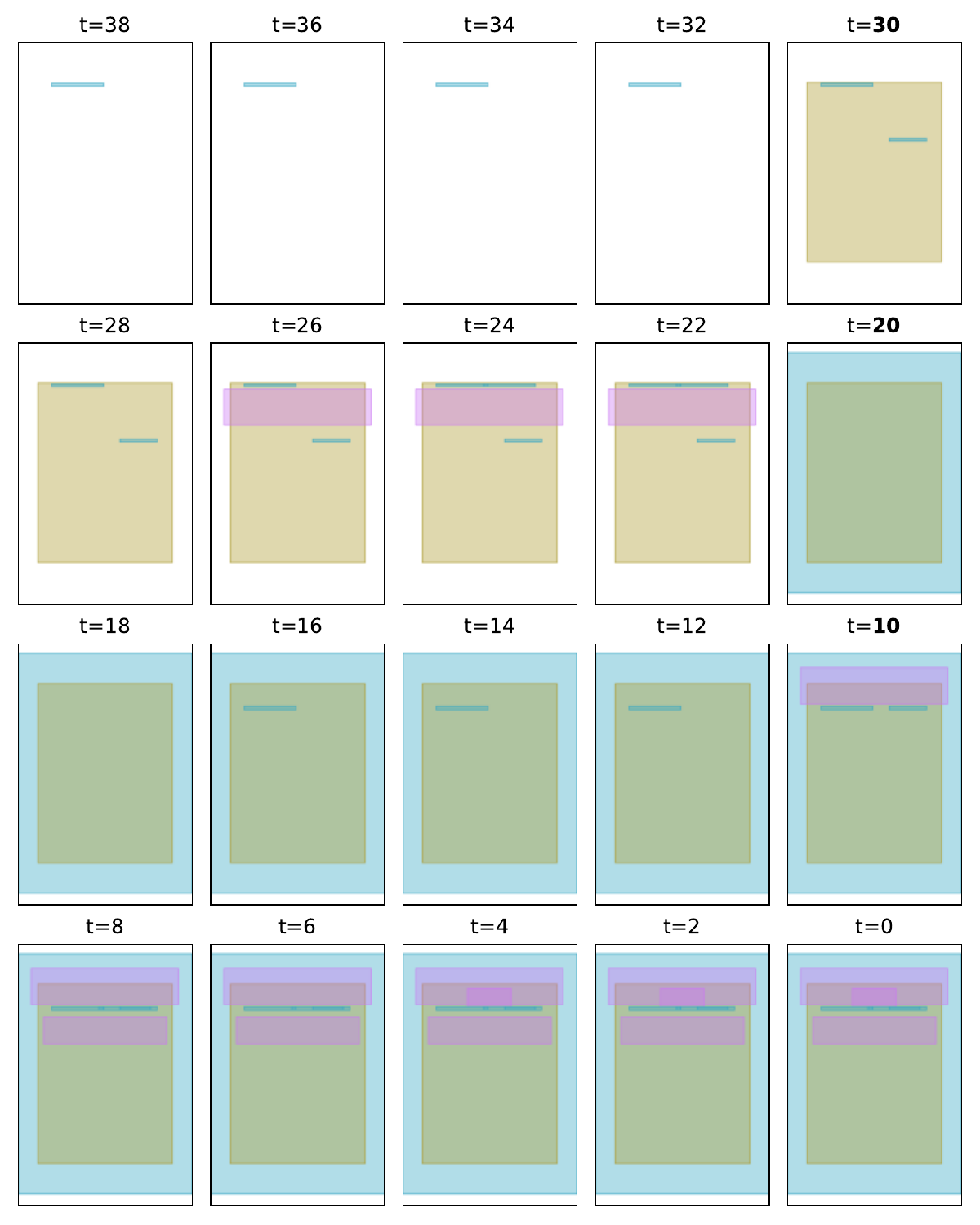}
        \caption{Example2: LayoutDM + Corrector}
        \label{figure:supp_gen-process-crello_example2_corrector}
    \end{subfigure}
    \caption{Comparison of unconditional generation process for Crello. Left:~the results of LayoutDM. Right:~the results of LayoutDM in conjunction with \layoutcorrector{}. 
    The timestep is denoted at the top of each layout visualization, and the timesteps when the corrector is applied are highlighted by \textbf{bold} in \cref{figure:supp_gen-process-crello_example1_corrector} and \cref{figure:supp_gen-process-crello_example2_corrector}.}
    \label{figure:supp_gen-process-crello}
\end{figure*}

\begin{figure*}[ht]

    \centering
    \begin{subfigure}[b]{0.48\linewidth}
        \includegraphics[width=\linewidth]{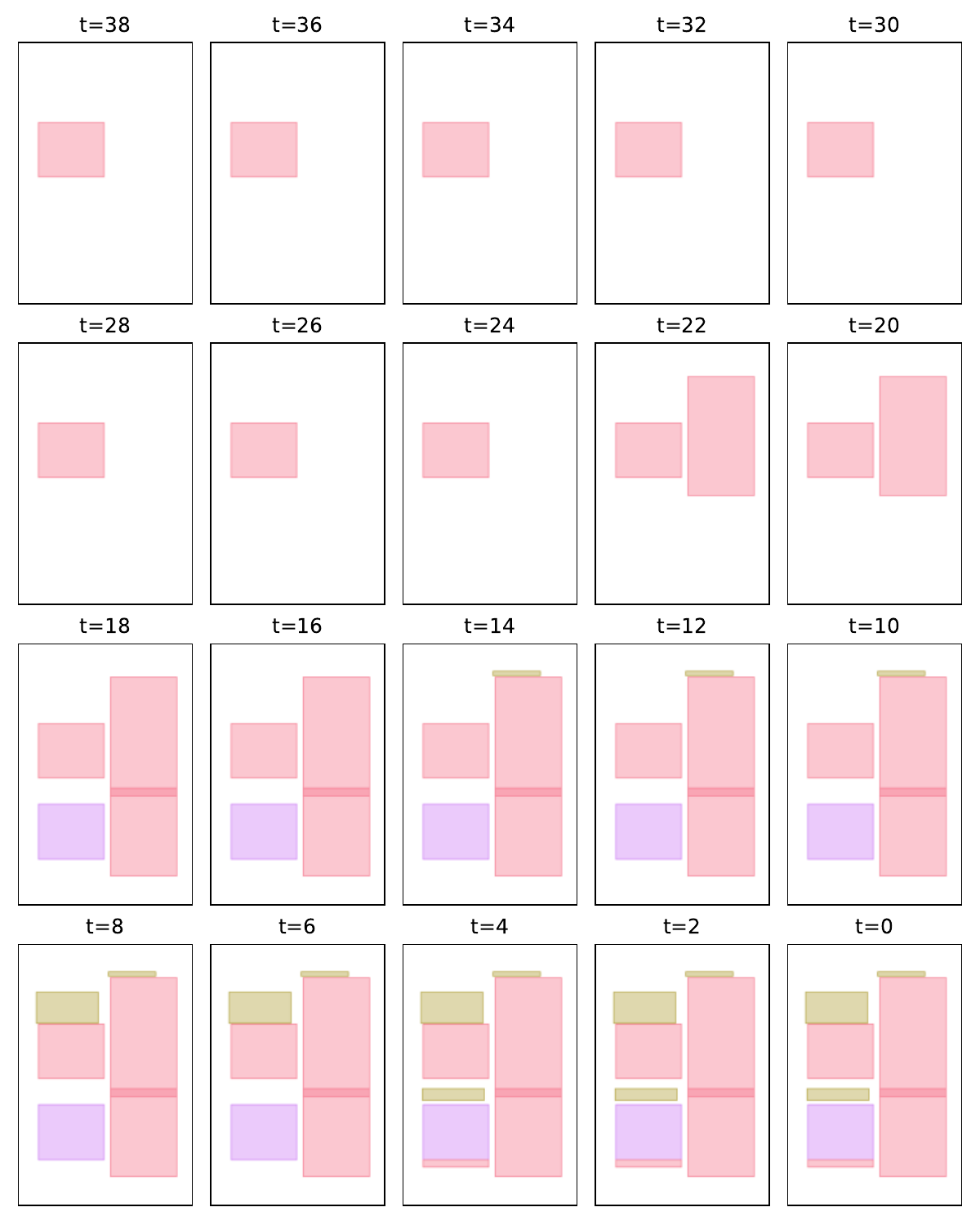}
        \caption{Example1: LayoutDM}
        \label{figure:supp_gen-process-publaynet_example1_layoutdm}
    \end{subfigure}
    \hfill
    \begin{subfigure}[b]{0.48\linewidth}
        \includegraphics[width=\linewidth]{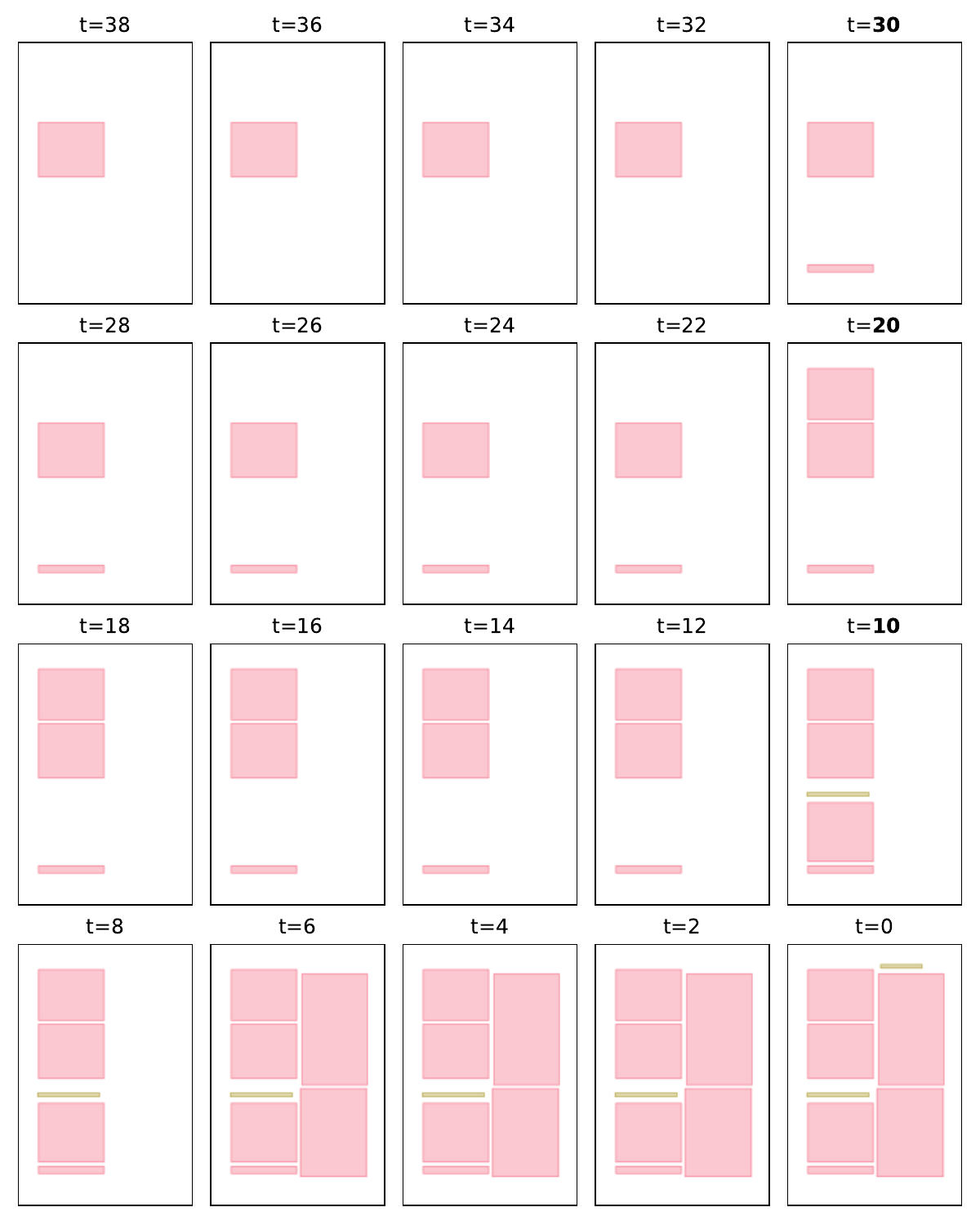}
        \caption{Example1: LayoutDM + Corrector}
        \label{figure:supp_gen-process-publaynet_example1_corrector}
    \end{subfigure}
    \vspace{10pt}

    \begin{subfigure}[b]{0.48\linewidth}
        \includegraphics[width=\linewidth]{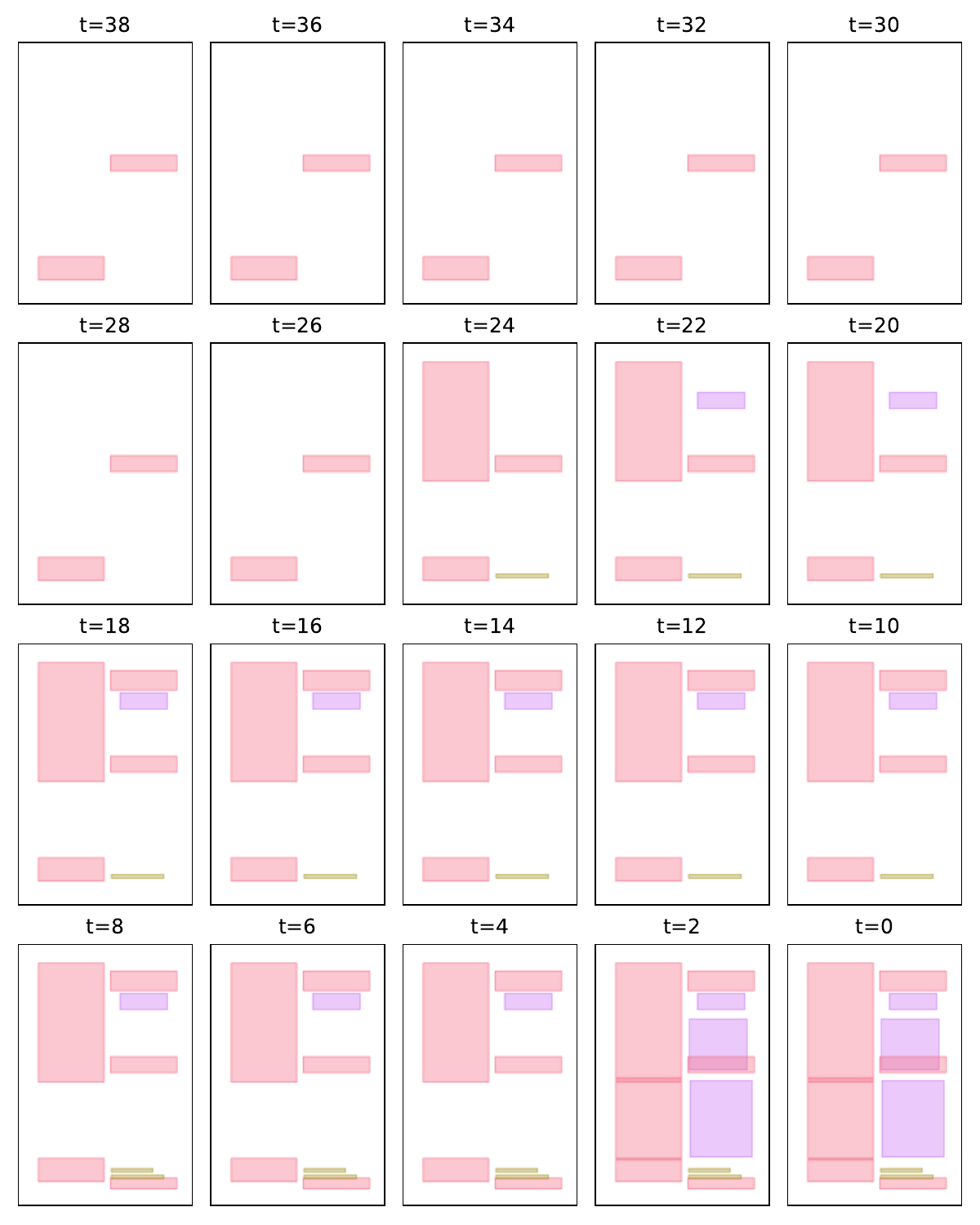}
        \caption{Example2: LayoutDM}
        \label{figure:supp_gen-process-publaynet_example2_layoutdm}
    \end{subfigure}
    \hfill
    \begin{subfigure}[b]{0.48\linewidth}
        \includegraphics[width=\linewidth]{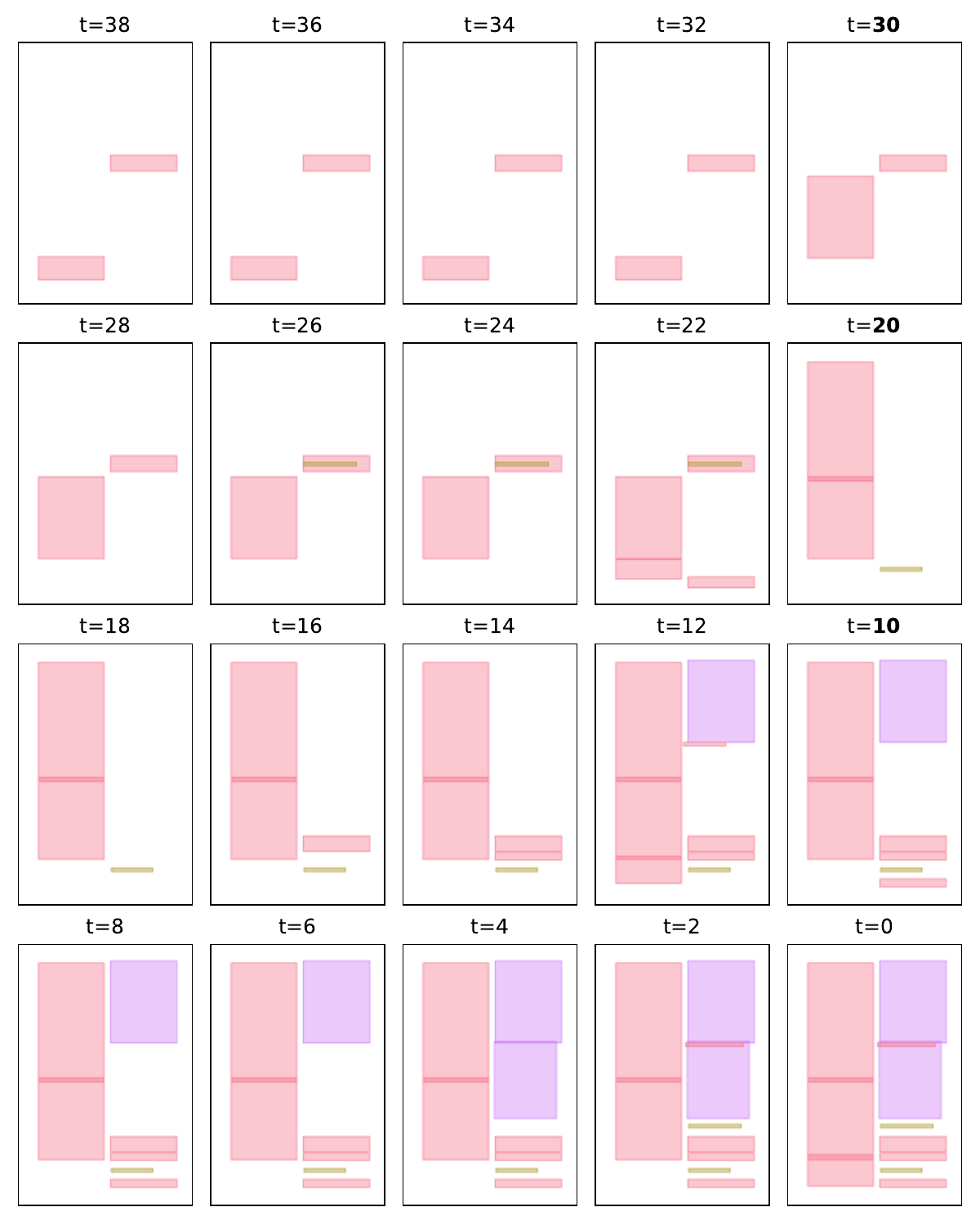}
        \caption{Example2: LayoutDM + Corrector}
        \label{figure:supp_gen-process-publaynet_example2_corrector}
    \end{subfigure}
    \caption{Comparison of unconditional generation process for PubLayNet. Left:~the results of LayoutDM. Right:~the results of LayoutDM in conjunction with \layoutcorrector{}. 
    The timestep is denoted at the top of each layout visualization, and the timesteps when the corrector is applied are highlighted by \textbf{bold} in \cref{figure:supp_gen-process-publaynet_example1_corrector} and \cref{figure:supp_gen-process-publaynet_example2_corrector}.}
    \label{figure:supp_gen-process-publaynet}
\end{figure*}

We present the layout visualization during the generation process for LayoutDM and its integration with \layoutcorrector{}. 
\cref{figure:supp_gen-process-rico}, \cref{figure:supp_gen-process-crello}, \cref{figure:supp_gen-process-publaynet} are the results of unconditional generation for the Rico, Crello, and PubLayNet datasets, respectively.
At earlier timesteps, such as $t \geq 40$, few elements have been generated, so we focus on visualizing the timesteps from $t=38$ to 0. 
The corrector is applied at timesteps $t=\{10,20,30\}$, which is the optimal schedule based on the FID score, as discussed in \cref{sec:corrector_analysis}. 
It is important to note that until $t > 30$, both models follow the identical generation process. 
The results demonstrate that \layoutcorrector{} effectively eliminates inharmonious elements at the timesteps when the corrector is applied, leading to more consistent results compared to the baseline.

\subsection{Fidelity-Diversity Trade-Off}

\cref{figure:supp_fidelity_diversity} displays the results from different scheduling scenarios of Layout-Corrector, illustrating layouts generated by LayoutDM~\cite{inoue2023layoutdm} with and without the Layout-Corrector under two distinct corrector schedules: $t=\{10,20,30\}$ and $t=\{10,20,\ldots,90\}$. Layouts generated with \layoutcorrector{} applied at $t=\{10,20,30\}$ demonstrate rich diversity. In contrast, more frequent application of \layoutcorrector{} at $t=\{10,20,\ldots,90\}$ results in a noticeable increase in layouts featuring centrally aligned elements along the horizontal axis, indicating reduced diversity. It is consistent with the observations in \cref{fig:fd_tradeoff}, where the more frequent application of \layoutcorrector{} to LayoutDM enhances fidelity but reduces diversity, highlighting a trade-off between these two aspects.

We showed the histogram of the width attribute across different corrector schedules in \cref{fig:attr_hist}.
Here, we also report the histogram of the other four attributes (\ie, category, x-center, y-center, and height) in \cref{figure:supp_attr_hist}. We observed the same trend as \cref{fig:attr_hist}, where the more frequent application of \layoutcorrector{} amplifies the frequency trends of the original data.

\subsection{Typical Failure Cases}

\begin{figure}
    \captionsetup[subfigure]{justification=centering}

    \begin{minipage}{0.49\linewidth}
        \centering
        \frame{\includegraphics[width=0.47\linewidth]{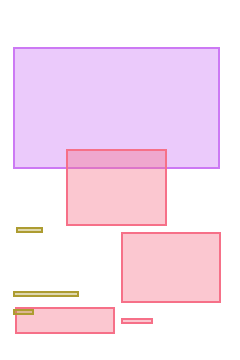}}
        \frame{\includegraphics[width=0.47\linewidth]{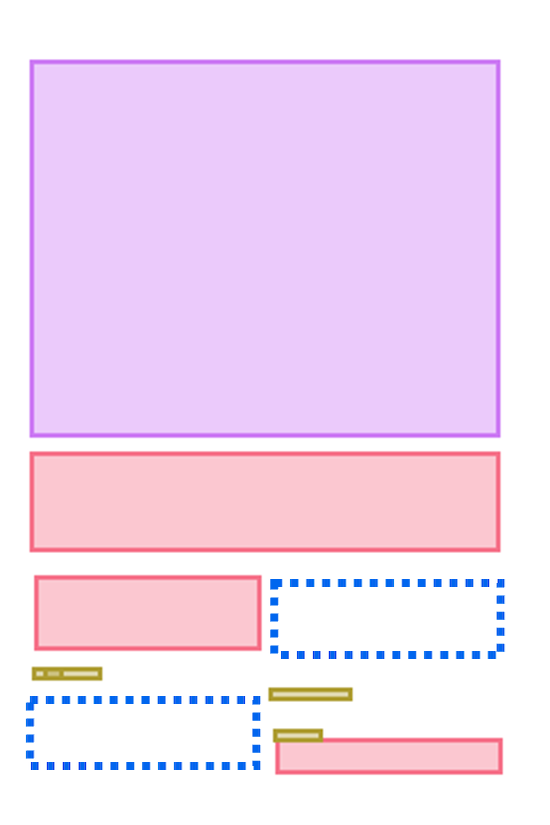}}
        \subcaption{Left: LayoutDM. \\ Right: LayoutDM + Corrector}
        \label{figure:supp_failure_example_1}
    \end{minipage}
    \begin{minipage}{0.49\linewidth}
        \centering
        \frame{\includegraphics[width=0.47\linewidth]{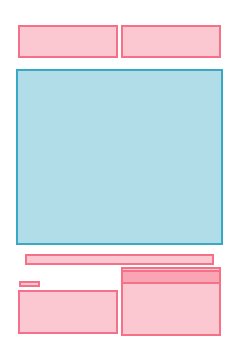}}
        \frame{\includegraphics[width=0.47\linewidth]{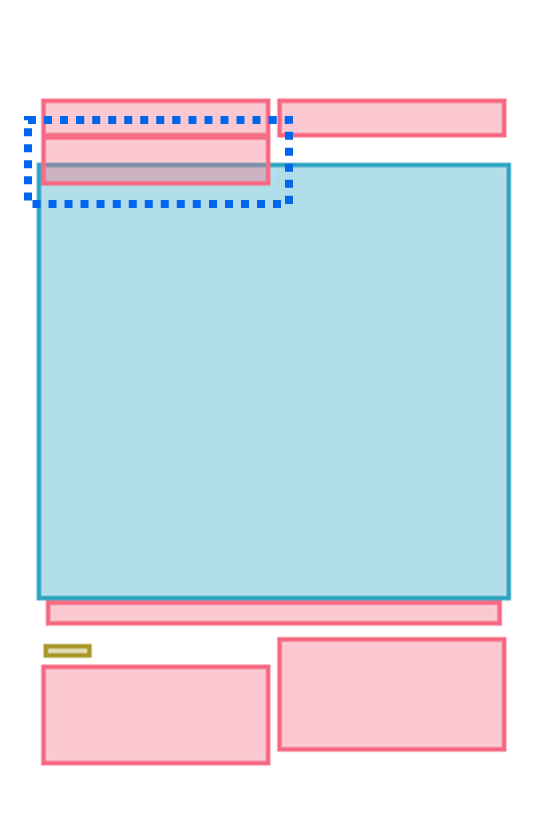}}
        \subcaption{Left: LayoutDM. \\ Right: LayoutDM + Corrector}
        \label{figure:supp_failure_example_2}
    \end{minipage}

    \caption{Typical failure cases on the unconditional task on PubLayNet dataset. We show the outputs from LayoutDM with and without \layoutcorrector{}. In \cref{figure:supp_failure_example_1}, although \layoutcorrector{} resolves overlapping elements found in the LayoutDM output, it leads to unnatural blank spaces in the LayoutDM + Corrector output. In \cref{figure:supp_failure_example_2}, while Layout-Corrector rectifies an overlap in the bottom-right of the LayoutDM output, a new overlap appears in the top-left in the LayoutDM + Corrector output.}
    \label{figure:supp_failure_cases}
\end{figure}

While \layoutcorrector{} can improve the generation quality of baseline models, it is not infallible. Typical failure cases are presented in \cref{figure:supp_failure_cases}, where we compare the layouts generated by LayoutDM~\cite{inoue2023layoutdm} with and without \layoutcorrector{}. 
In \cref{figure:supp_failure_example_1}, \layoutcorrector{} effectively resolves overlap and misalignment in LayoutDM's output, but this produces unnatural blank spaces in the output. 
This issue arises because, although Layout-Corrector enables DDMs to modify incorrectly generated layouts by resetting tokens with low correctness scores, it does not encourage DDMs to create additional elements, leading to these blank areas. In \cref{figure:supp_failure_example_2}, \layoutcorrector{} fixes an overlap in the bottom-right of LayoutDM's output, yet a new overlap emerges in the top-left of the LayoutDM + Corrector output. 
This is because Layout-Corrector can not correct tokens generated after its final application.

\section{Ablation Study}
\label{supp:ablation}

In this section, we present additional results of the ablation study, including Crello~\cite{yamaguchi2021canvasvae} and PubLayNet~\cite{zhong2019publaynet}, and architecture of \layoutcorrector{}, and threshold value $\theta_{th}$. In addition, we compare \layoutcorrector{} with rule-based post-processing~\cite{kikuchi2021constrained}.

\subsection{Additional Results}
\begin{table}[t]
    \caption{Ablation study on Crello~\cite{yamaguchi2021canvasvae} and PubLayNet~\cite{zhong2019publaynet} dataset with unconditional generation.}
    \label{tab:ablation_publaynet_crello}
    \begin{minipage}{0.49\linewidth}
        \centering
\resizebox{\textwidth}{!}{%
\begin{tabular}{lcccc}
\toprule
                                                  & \multicolumn{2}{c}{$T' = 100$}                                                & \multicolumn{2}{c}{$T' = 20$}                                                 \\ \cmidrule(r){2-3} \cmidrule(l){4-5}
                                                  & \multicolumn{1}{c}{FID$\downarrow$} & \multicolumn{1}{c}{Align.$\rightarrow$} & \multicolumn{1}{c}{FID$\downarrow$} & \multicolumn{1}{c}{Align.$\rightarrow$} \\ \cmidrule(r){1-1} \cmidrule(lr){2-2} \cmidrule(lr){3-3} \cmidrule(lr){4-4} \cmidrule(l){5-5}
\rowcolor[HTML]{EFEFEF}\textbf{Layout-Corrector} & \textbf{4.36} & 0.232 & \textbf{5.11} & 0.295\\
Mask estimation                                   & 4.71 & 0.285 & 6.22 & \textbf{0.336}\\
w/o Self-Atteniton                                & 4.42 & 0.260 & 6.11 & 0.317 \\
Top-K                                & 6.58 & \textbf{0.300} & 5.45 & 0.296                                   \\
Correcting at every $t$                               & 90.24 & 0.009 & 48.78 & 0.038                                  \\ \cmidrule(r){1-1} \cmidrule(lr){2-2} \cmidrule(lr){3-3} \cmidrule(lr){4-4} \cmidrule(l){5-5}
\textit{Real Data}                                & 2.32                                & 0.338                                   & 2.32                                & 0.338                                   \\ \bottomrule
\end{tabular}
}
\subcaption{Crello~\cite{yamaguchi2021canvasvae}}
\label{tab:ablation_crello}
    \end{minipage} \hfill
    \begin{minipage}{0.49\linewidth}
        \centering
\resizebox{\textwidth}{!}{%
\begin{tabular}{lcccc}
\toprule
                          & \multicolumn{2}{c}{$T' = 100$}        & \multicolumn{2}{c}{$T' = 20$}         \\ \cmidrule(r){2-3} \cmidrule(l){4-5}
                          & FID$\downarrow$ & Align.$\rightarrow$ & FID$\downarrow$ & Align.$\rightarrow$ \\ \cmidrule(r){1-1} \cmidrule(lr){2-2} \cmidrule(lr){3-3} \cmidrule(lr){4-4} \cmidrule(l){5-5}
\rowcolor[HTML]{EFEFEF} 
\textbf{Layout-Corrector} & 11.85 & 0.172 & \textbf{15.39} & 0.178              \\
Mask estimation           & \textbf{11.40} & 0.167 & 16.71 & 0.194              \\
w/o Self-Atteniton        & 13.49 & 0.172 & 20.71 & 0.286               \\
Top-K        &19.96 & 0.615 & 23.89 & 0.443               \\
Correcting at every $t$       & 69.21 & \textbf{0.125} & 53.26 & \textbf{0.092}     \\ \cmidrule(r){1-1} \cmidrule(lr){2-2} \cmidrule(lr){3-3} \cmidrule(lr){4-4} \cmidrule(l){5-5}
\textit{Real Data}        & 6.25            & 0.021               & 6.25            & 0.021               \\ \bottomrule
\end{tabular}
}
\subcaption{PubLayNet~\cite{zhong2019publaynet}}
\label{tab:ablation_publaynet}
    \end{minipage}
\end{table}

\cref{tab:ablation_publaynet_crello} shows the ablation results for Crello and PubLayNet in the unconditional generation task.
Please refer to \cref{subsec:ablation} regarding the configurations.
As with the results of Rico dataset~\cite{deka2017rico}, \layoutcorrector{} achieves solid performance on both Crello and PubLayNet datasets.
For Crello dataset shown in \cref{tab:ablation_crello}, we observed that removing the self-attention layer results in a less significant performance drop than in other datasets. We hypothesize that this phenomenon is due to the complex and diverse relationships between elements in Crello, as illustrated by real samples in \cref{fig:sup_comparison_unconditional_crello}. When the relationships among elements are complicated, it is challenging for the self-attention layers to capture these relationships, resulting in decreased effectiveness.

\subsection{Corrector Architecture}
We report the effects of varying the number of Transformer Encoder layers in \layoutcorrector{}. 
To investigate this, we trained \layoutcorrector{} with $\{1,2,4,6\}$ layers on the Rico dataset~\cite{deka2017rico} and evaluated FID scores, number of parameters, and inference speed. 
The application schedule for \layoutcorrector{} was set to $t=\{10,20,30\}$. 
The results, presented in \cref{table:supp_ablation_num-enc-layers}, indicate that the best FID is achieved with 4 encoder layers. 
Although the number of parameters increases with the number of layers, the impact on inference speed remains minimal since the corrector is applied just three times. 

\begin{table}[t]
    \centering
    \caption{The effect of the number of Transformer Encoder layers on Rico test set. The best FID result is highlighted in \textbf{bold}. }
    \label{table:supp_ablation_num-enc-layers}
    \small{
    \begin{tabular}{@{}cccc@{}}
    \toprule
    \begin{tabular}[c]{@{}c@{}}\# of layers\end{tabular} & FID$\downarrow$ & \# of params {[}M{]} & Time/sample {[}ms{]} \\ \cmidrule(r){1-1} \cmidrule(lr){2-2} \cmidrule(lr){3-3} \cmidrule(l){4-4}
    -~(LayoutDM) & 6.37               & 12.4   & 23.7  \\
    1 & 5.07                           & + 4.6  & 24.6 \\
    2 & 4.94                           & + 7.3  & 24.6 \\
    4 & \textbf{4.79}                  & + 12.6  & 24.6 \\
    6 & 4.93                           & + 17.9 & 24.6 \\ \bottomrule
    \end{tabular}
    }
\end{table}

\begin{table}[t]
\caption{The impact of threshold $\theta_{th}$ in unconditional generation on three dataset using LayoutDM~\cite{inoue2023layoutdm} as DDM. The best result is highlighted in \textbf{bold}. }
\label{table:supp_ablation_threshold}
\centering
\scriptsize{
\begin{tabular}{@{}cccccccccc@{}}
\toprule
\multicolumn{1}{l}{\multirow{2}{*}{Threshold $\theta_{th}$}}    & \multicolumn{3}{c}{Rico}  & \multicolumn{3}{c}{Crello} & \multicolumn{3}{c}{PubLayNet} \\ \cmidrule(lr){2-4} \cmidrule(lr){5-7} \cmidrule(lr){8-10}
\multicolumn{1}{l}{}     & FID$\downarrow$  & Precision$\uparrow$ & Recall$\uparrow$ & FID$\downarrow$   & Precision$\uparrow$ & Recall$\uparrow$ & FID$\downarrow$    & Precision$\uparrow$  & Recall$\uparrow$  \\ \cmidrule(lr){1-1} \cmidrule(lr){2-2} \cmidrule(lr){3-3} \cmidrule(lr){4-4} \cmidrule(lr){5-5} \cmidrule(lr){6-6} \cmidrule(lr){7-7} \cmidrule(lr){8-8} \cmidrule(lr){9-9} \cmidrule(lr){10-10}
0.3                     & 5.57 & 0.763     & 0.900  & 5.16  & 0.775     & \textbf{0.874}  & 12.88  & 0.605      & \textbf{0.920}   \\
0.4                     & 5.26 & 0.779     & 0.897  & 4.97  & 0.789     & 0.861  & 12.41  & 0.639      & 0.918   \\
0.5                     & 5.05 & 0.787     & \textbf{0.900}  & 4.75  & 0.799     & 0.861  & 12.06  & 0.668      & 0.914   \\
0.6                     & 4.90 & 0.794     & 0.892  & 4.45  & 0.806     & 0.859  & \textbf{11.78}  & 0.681      & 0.911   \\
\rowcolor[HTML]{EFEFEF}
\textbf{0.7}                     & \textbf{4.79} & 0.809     & 0.892  & \textbf{4.36}  & 0.822     & 0.851  & 11.85  & 0.711      & 0.890   \\
0.8                     & 5.01 & 0.822     & 0.876  & 4.51  & 0.824     & 0.834  & 11.88  & 0.727      & 0.887   \\
0.9                     & 5.89 & \textbf{0.844}     & 0.858  & 5.77  & \textbf{0.849}     & 0.811  & 12.60  & \textbf{0.729}      & 0.869   \\ \bottomrule
\end{tabular}%
}
\end{table}

\subsection{Threshold $\theta_{th}$}
We compared various threshold values $\theta_{th}$ in \cref{table:supp_ablation_threshold} on three datasets. The results show that $\theta_{th}=0.7$ yields the best FID on Rico and Crello, and the second-best on PubLayNet, demonstrating that it performs well across various datasets without tailored calibration.
Precision and Recall scores are also presented in the table to provide a more comprehensive analysis. A higher threshold keeps only high-scored tokens, leading to higher fidelity (Precision) at the expense of diversity (Recall). In contrast, a lower threshold allows the inclusion of low-scored tokens, potentially enhancing diversity at the cost of reduced fidelity.

\subsection{Effect of post-processing}
\label{subsec:supp_post_process}
\begin{table*}[t]
    \centering
    \caption{Performance comparison of baseline models with/without post-processing on the unconditional generation task.
    Metrics improved by post-processing are highlighted in \textbf{bold}. 
    }
    \label{tab:base_plus_post}
    \resizebox{\textwidth}{!}{
        \begin{tabular}{lccccccccc}
        \toprule
                & \multicolumn{3}{c}{Rico~\cite{deka2017rico}}  & \multicolumn{3}{c}{Crello~\cite{yamaguchi2021canvasvae}} & \multicolumn{3}{c}{PubLayNet~\cite{zhong2019publaynet}}  \\ 
          \cmidrule(r){2-4} \cmidrule(lr){5-7} \cmidrule(l){8-10}
        Model   & FID$\downarrow$ & Align.$\rightarrow$ & Overlap$\rightarrow$ & FID$\downarrow$ & Align.$\rightarrow$ & Overlap$\rightarrow$ & FID$\downarrow$ & Align.$\rightarrow$ & Overlap$\rightarrow$ \\ \cmidrule(r){1-1} \cmidrule(lr){2-2} \cmidrule(lr){3-3} \cmidrule(lr){4-4} \cmidrule(lr){5-5} \cmidrule(lr){6-6} \cmidrule(lr){7-7} \cmidrule(lr){8-8} \cmidrule(lr){9-9} \cmidrule(lr){10-10} 
        LayoutDM~\cite{inoue2023layoutdm}  & 6.37 & 0.223 & 0.841 & 5.28 & 0.279 & 1.733 & 13.72 & 0.185 & 0.142           \\
        \rowcolor[HTML]{EFEFEF} 
        ~~\textbf{+ post-processing}             &  \textbf{6.23} & \textbf{0.211} & 0.854 & \textbf{5.20} & 0.258 & 1.738 & 13.77 & \textbf{0.16} & \textbf{0.052}     \\
     \cmidrule(r){1-1} \cmidrule(lr){2-2} \cmidrule(lr){3-3} \cmidrule(lr){4-4} \cmidrule(lr){5-5} \cmidrule(lr){6-6} \cmidrule(lr){7-7} \cmidrule(lr){8-8} \cmidrule(lr){9-9} \cmidrule(l){10-10}
        LayoutDM + Corrector    & 4.79 & 0.167 & 0.884 & 4.36 & 0.232 & 1.829 & 11.85 & 0.172 & 0.082            \\
        \rowcolor[HTML]{EFEFEF} 
        ~~\textbf{+ post-processing}  & 4.87 & \textbf{0.158} & 0.897 & 4.36 & 0.215 & 1.834 & \textbf{10.81} & \textbf{0.120} & \textbf{0.023}     \\  
        \cmidrule(r){1-1} \cmidrule(lr){2-2} \cmidrule(lr){3-3} \cmidrule(lr){4-4} \cmidrule(lr){5-5} \cmidrule(lr){6-6} \cmidrule(lr){7-7} \cmidrule(lr){8-8} \cmidrule(lr){9-9} \cmidrule(l){10-10}
        \textit{Real data} & 1.85 & 0.109 & 0.665 & 2.32 & 0.338 & 1.625 & 6.25 & 0.021 & 0.0032     \\    \bottomrule
        \end{tabular}
}
\end{table*}

In this section, we investigate the applicability of post-processing to refine layouts. 
To achieve this, we use the layouts generated by DDMs and refine them using rule-based methods. 
Following the approach of CLG-LO~\cite{kikuchi2021constrained}, we apply constraint optimization to geometric metrics, including alignment and overlap scores, to minimize these costs while modifying geometric attributes. 
For datasets characterized by a large overlap, such as Rico and Crello, we adjust the optimization by omitting the overlap term from the objective function and focusing solely on minimizing the alignment.

\cref{tab:base_plus_post} shows the effect of post-processing on LayoutDM and its combination with Layout-Corrector.
We observe that post-processing does not significantly affect the FID score, except for LayoutDM + Corrector on PubLayNet, which has lower alignment and overlap scores.
We consider that optimization based on geometric constraints is ineffective for layouts with complex structures, such as Rico and Crello.
On the other hand, Layout-Corrector outperforms post-processing in terms of FID because it intervenes in the generation process to realize layout correction.
This suggests that our learning-based approach is far more effective than simple rule-based optimization.

\newcommand{\inplength}{0.17}
\newcommand{\inptblheader}{
    &
    \multirow{2}{*}{MaskGIT~\cite{chang2022maskgit}} &
    \multirow{2}{*}{\shortstack{MaskGIT\\+Corrector}} &
    \multirow{2}{*}{LayoutDM~\cite{inoue2023layoutdm}} &
    \multirow{2}{*}{\shortstack{LayoutDM\\+Corrector}}
    \\
}
\newcommand{\uctblheader}{
    \multirow{2}{*}{Real data} &
    \multirow{2}{*}{MaskGIT~\cite{chang2022maskgit}} &
    \multirow{2}{*}{\shortstack{MaskGIT\\+Corrector}} &
    \multirow{2}{*}{LayoutDM~\cite{inoue2023layoutdm}} &
    \multirow{2}{*}{\shortstack{LayoutDM\\+Corrector}}
    \\
}
\newcommand{\rftblheader}{
    Input & RUITE~\cite{rahman2021ruite} & \multicolumn{3}{c}{LayoutDM} & Ground Truth \\
}

\newcommand{\inprow}[4]{  %
    \frame{\includegraphics[height=\inplength\hsize]{figures/png/supp/comparison/#1_#2/#3_maskgit#4.png}} &
    \frame{\includegraphics[height=\inplength\hsize]{figures/png/supp/comparison/#1_#2/#3_maskgit_corrector#4.png}} &
    \frame{\includegraphics[height=\inplength\hsize]{figures/png/supp/comparison/#1_#2/#3_layoutdm#4.png}} &
    \frame{\includegraphics[height=\inplength\hsize]{figures/png/supp/comparison/#1_#2/#3_layoutdm_corrector_10_30#4.png}}
}

\newcommand{\inprows}[7]{
    Input & & & & & & \\
    \frame{\includegraphics[width=0.09\hsize]{figures/png/supp/comparison/#1_#2/#3_input_000.png}} & \inprow{#1}{#2}{#3}{#4} \\
    Real Data & \inprow{#1}{#2}{#3}{#5} \\
    \frame{\includegraphics[width=0.09\hsize]{figures/png/supp/comparison/#1_#2/#3_Real_000.png}} & \inprow{#1}{#2}{#3}{#6} \\
     & \inprow{#1}{#2}{#3}{#7}
}

\newcommand{\ucrow}[3]{
    \frame{\includegraphics[height=\inplength\hsize]{figures/png/supp/comparison/#1_#2/#3_Real.png}} &
    \inprow{#1}{#2}{#3}{}
}
\newcommand{\rflength}{0.16}

\newcommand{\rfrow}[3]{
    \frame{\includegraphics[height=\rflength\hsize]{figures/png/supp/comparison/#1_#2/#3_input.png}} &
    \frame{\includegraphics[height=\rflength\hsize]{figures/png/supp/comparison/#1_#2/#3_ruite_0.png}} &
    \frame{\includegraphics[height=\rflength\hsize]{figures/png/supp/comparison/#1_#2/#3_layoutdm_0.png}} &
    \frame{\includegraphics[height=\rflength\hsize]{figures/png/supp/comparison/#1_#2/#3_layoutdm_1.png}} &
    \frame{\includegraphics[height=\rflength\hsize]{figures/png/supp/comparison/#1_#2/#3_layoutdm_2.png}} &
    \frame{\includegraphics[height=\rflength\hsize]{figures/png/supp/comparison/#1_#2/#3_gt.png}}
}

{
    \begin{figure*}[t]
    \centering
    \setlength\tabcolsep{5pt}
    \tiny{
        \begin{tabular}{c@{\hspace{3em}}cccccc}
            \uctblheader \\
            \ucrow{rico}{unconditional}{seed0_00045} \\
            \ucrow{rico}{unconditional}{seed0_00053} \\
            \ucrow{rico}{unconditional}{seed0_00093} \\
            \ucrow{rico}{unconditional}{seed0_00121} \\
            \ucrow{rico}{unconditional}{seed0_00147} \\
            \ucrow{rico}{unconditional}{seed0_00159} \\
            \ucrow{rico}{unconditional}{seed0_00164} \\
            \ucrow{rico}{unconditional}{seed0_00189} \\
        \end{tabular}
    }
    \caption{
        Comparison of unconditional generation results on Rico, with eight samples from each model to show diversity.
    }
    \label{fig:sup_comparison_unconditional_rico}
    \end{figure*}
}

{
    \begin{figure*}[t]
    \centering
    \setlength\tabcolsep{5pt}
    \tiny{
        \begin{tabular}{c@{\hspace{3em}}cccccc}
            \uctblheader \\
            \ucrow{crello}{unconditional}{seed0_00002} \\
            \ucrow{crello}{unconditional}{seed0_00010} \\
            \ucrow{crello}{unconditional}{seed0_00030} \\
            \ucrow{crello}{unconditional}{seed0_00045} \\
            \ucrow{crello}{unconditional}{seed0_00073} \\
            \ucrow{crello}{unconditional}{seed0_00079} \\
            \ucrow{crello}{unconditional}{seed0_00169} \\
            \ucrow{crello}{unconditional}{seed0_00199} \\
        \end{tabular}
    }
    \caption{
        Comparison of unconditional generation results on Crello, with eight samples from each model to show diversity.
    }
    \label{fig:sup_comparison_unconditional_crello}
    \end{figure*}
}

{
    \begin{figure*}[t]
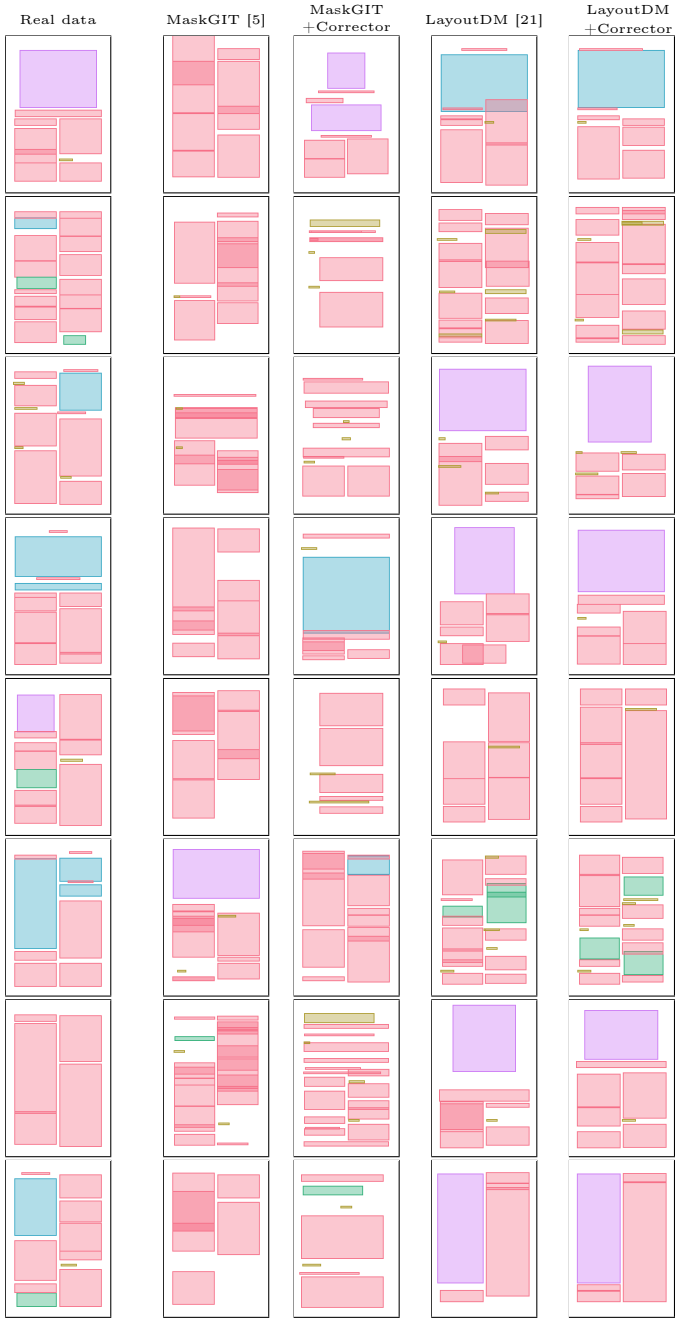

    \centering
    \setlength\tabcolsep{5pt}
    \tiny{
        \begin{tabular}{c@{\hspace{3em}}ccccc}
            \uctblheader \\
            \ucrow{publaynet}{unconditional}{seed0_00007} \\
            \ucrow{publaynet}{unconditional}{seed0_00025} \\
            \ucrow{publaynet}{unconditional}{seed0_00042} \\
            \ucrow{publaynet}{unconditional}{seed0_00081} \\
            \ucrow{publaynet}{unconditional}{seed0_00126} \\
            \ucrow{publaynet}{unconditional}{seed0_00155} \\
            \ucrow{publaynet}{unconditional}{seed0_00166} \\
            \ucrow{publaynet}{unconditional}{seed0_00192} \\
        \end{tabular}
    }
    \caption{
        Comparison of unconditional generation results on PubLayNet, with eight samples from each model to show diversity.
    }
    \label{fig:sup_comparison_unconditional_publaynet}
    \end{figure*}
}

{
    \begin{figure*}[t]
    \centering
    \setlength\tabcolsep{5pt}
    \tiny{
        \begin{tabular}{c@{\hspace{3em}}cccccc}
            \inptblheader \\
            \inprows{rico}{c}{seed0_00000}{_001}{_002}{_003}{_004} \\
             & & & & & & \\
            \inprows{rico}{c}{seed0_00083}{_001}{_002}{_003}{_004} \\
        \end{tabular}
    }
    \caption{
        Comparison of conditional generation results for C$\rightarrow$S+P on Rico, with four samples per condition input from each model to show diversity.
    }
    \label{fig:sup_comparison_c_rico}
    \end{figure*}
}

{
    \begin{figure*}[t]
    \centering
    \setlength\tabcolsep{5pt}
    \tiny{
        \begin{tabular}{c@{\hspace{3em}}cccccc}
            \inptblheader \\
            \inprows{crello}{c}{seed0_00000}{_001}{_002}{_003}{_004} \\
             & & & & & & \\
            \inprows{crello}{c}{seed0_00006}{_001}{_002}{_003}{_004} \\
        \end{tabular}
    }
    \caption{
        Comparison of conditional generation results for C$\rightarrow$S+P on Crello, with four samples per condition input from each model to show diversity.
    }
    \label{fig:sup_comparison_c_crello}
    \end{figure*}
}

{
    \begin{figure*}[t]
    \centering
    \setlength\tabcolsep{5pt}
    \tiny{
        \begin{tabular}{c@{\hspace{3em}}cccccc}
            \inptblheader \\
            \inprows{publaynet}{c}{seed0_00000}{_001}{_002}{_003}{_004} \\
             & & & & & & \\
            \inprows{publaynet}{c}{seed0_00008}{_001}{_002}{_003}{_004} \\
        \end{tabular}
    }
    \caption{
        Comparison of conditional generation results for C$\rightarrow$S+P on PubLayNet, with four samples per condition input from each model to show diversity.
    }
    \label{fig:sup_comparison_c_publaynet}
    \end{figure*}
}

{
    \begin{figure*}[t]
    \centering
    \setlength\tabcolsep{5pt}
    \tiny{
        \begin{tabular}{c@{\hspace{3em}}cccccc}
            \inptblheader \\
            \inprows{rico}{cwh}{seed0_00011}{_001}{_002}{_003}{_004} \\
             & & & & & & \\
            \inprows{rico}{cwh}{seed0_00026}{_001}{_002}{_003}{_004} \\
        \end{tabular}
    }
    \caption{
        Comparison of conditional generation results for C+S$\rightarrow$P on Rico, with four samples per condition input from each model to show diversity.
    }
    \label{fig:sup_comparison_cwh_rico}
    \end{figure*}
}

{
    \begin{figure*}[t]
    \centering
    \setlength\tabcolsep{5pt}
    \tiny{
        \begin{tabular}{c@{\hspace{3em}}cccccc}
            \inptblheader \\
            \inprows{crello}{cwh}{seed0_00018}{_001}{_002}{_003}{_004} \\
             & & & & & & \\
            \inprows{crello}{cwh}{seed0_00030}{_001}{_002}{_003}{_004} \\
        \end{tabular}
    }
    \caption{
        Comparison of conditional generation results for C+S$\rightarrow$P on Crello, with four samples per condition input from each model to show diversity.
    }
    \label{fig:sup_comparison_cwh_crello}
    \end{figure*}
}

{
    \begin{figure*}[t]
    \centering
    \setlength\tabcolsep{5pt}
    \tiny{
        \begin{tabular}{c@{\hspace{3em}}cccccc}
            \inptblheader \\
            \inprows{publaynet}{cwh}{seed0_00003}{_001}{_002}{_003}{_004} \\
             & & & & & & \\
            \inprows{publaynet}{cwh}{seed0_00010}{_001}{_002}{_003}{_004} \\
        \end{tabular}
    }
    \caption{
        Comparison of conditional generation results for C+S$\rightarrow$P on PubLayNet, with four samples per condition input from each model to show diversity.
    }
    \label{fig:sup_comparison_cwh_publaynet}
    \end{figure*}
}


\newcommand{\matsize}{14x2}
\newcommand{\fdwidth}{\linewidth}

\begin{figure*}[ht]

    \centering
    \begin{subfigure}[b]{\fdwidth}
        \includegraphics[width=\linewidth]{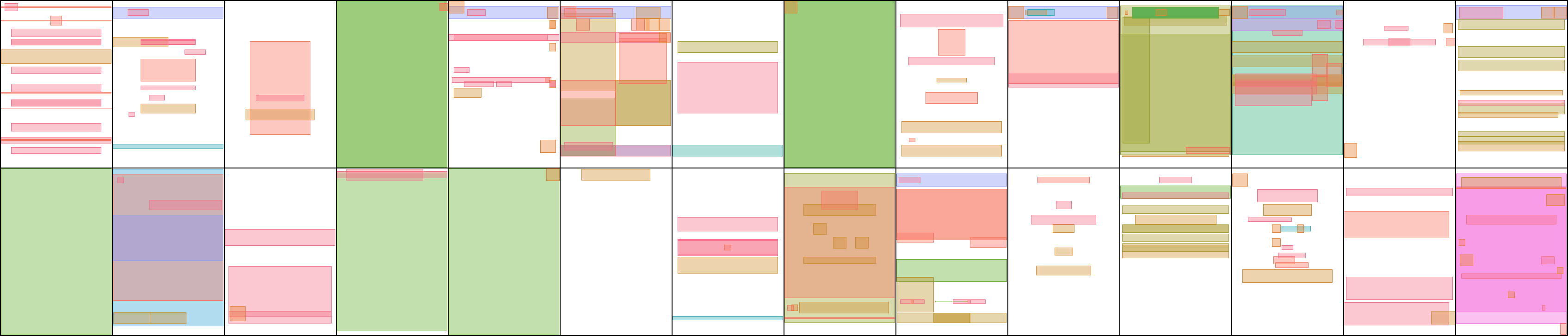}
        \caption{LayoutDM (FID $=6.38$, Precision $=0.750$)}
        \label{figure:rico_fidelity_diversity_layoutdm}
    \end{subfigure}
    \vspace{1pt}
    
    \begin{subfigure}[b]{\fdwidth}
        \includegraphics[width=\linewidth]{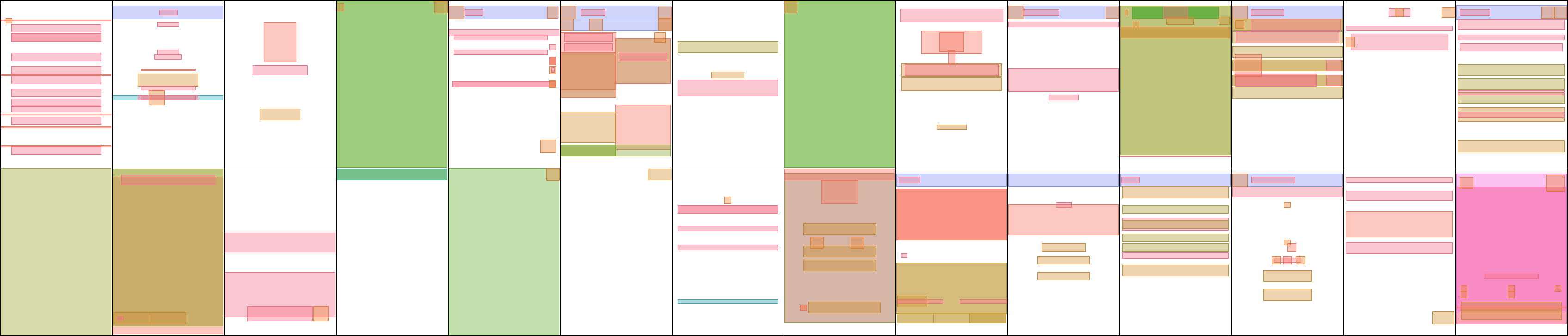}
        \caption{LayoutDM + Corrector $t=\{10, 20, 30\}$ (FID $=4.79$, Precision $=0.811$)}
        \label{figure:rico_fidelity_diversity_corr_t10_30}
    \end{subfigure}
    \vspace{1pt}
    
    \begin{subfigure}[b]{\fdwidth}
        \includegraphics[width=\linewidth]{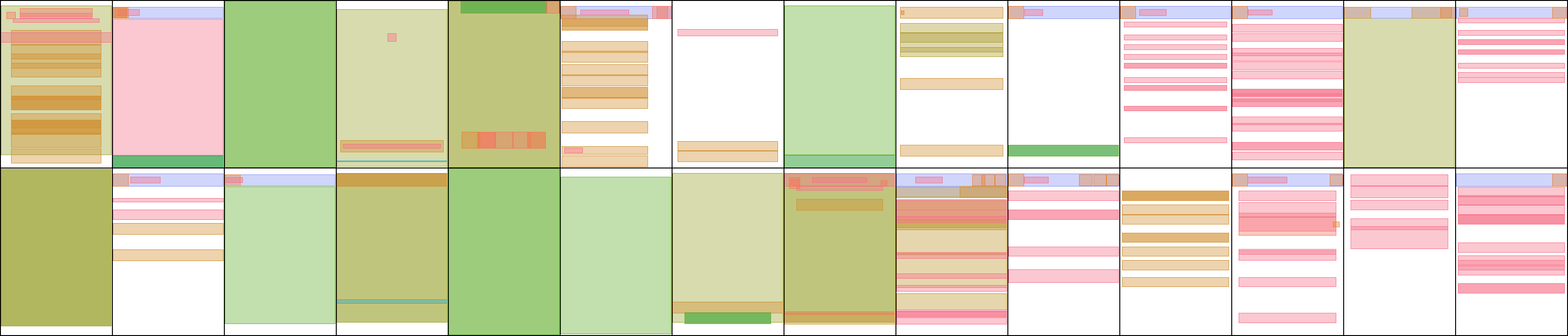}
        \caption{LayoutDM + Corrector $t=\{10, 20, 30, \ldots, 90\}$ (FID $=19.90$, Precision $=0.914$)}
        \label{figure:rico_fidelity_diversity_corr_t10_90}
    \end{subfigure}
    \vspace{1pt}
    
    \begin{subfigure}[b]{\fdwidth}
        \includegraphics[width=\linewidth]{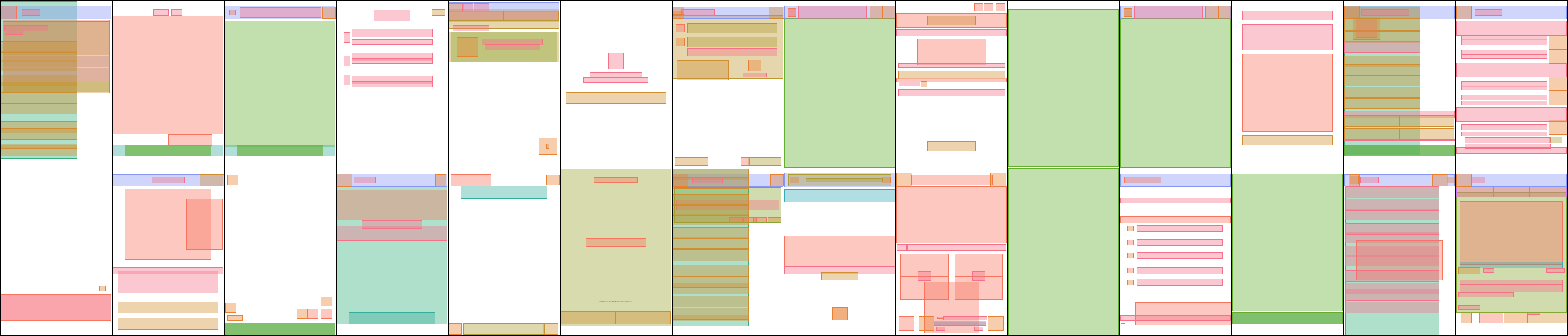}
        \caption{Real data}
        \label{figure:rico_fidelity_diversity_real}
    \end{subfigure}
    \caption{Visualization of unconditional generation on the Rico dataset. This figure displays outputs from LayoutDM and LayoutDM + Layout-Corrector under two distinct corrector scheduling scenarios. \cref{figure:rico_fidelity_diversity_corr_t10_30} illustrates the results of our default schedule ($t=\{10, 20, 30\}$), which produces high-quality and diverse layouts. In contrast, \cref{figure:rico_fidelity_diversity_corr_t10_90} shows that increasing the frequency of \layoutcorrector{} application leads to layouts with more elements centered along the horizontal axis, indicating reduced diversity.}
    \label{figure:supp_fidelity_diversity}
\end{figure*}
\begin{figure*}[ht]
    \centering
    \begin{subfigure}[b]{\fdwidth}
        \includegraphics[width=0.95\linewidth]{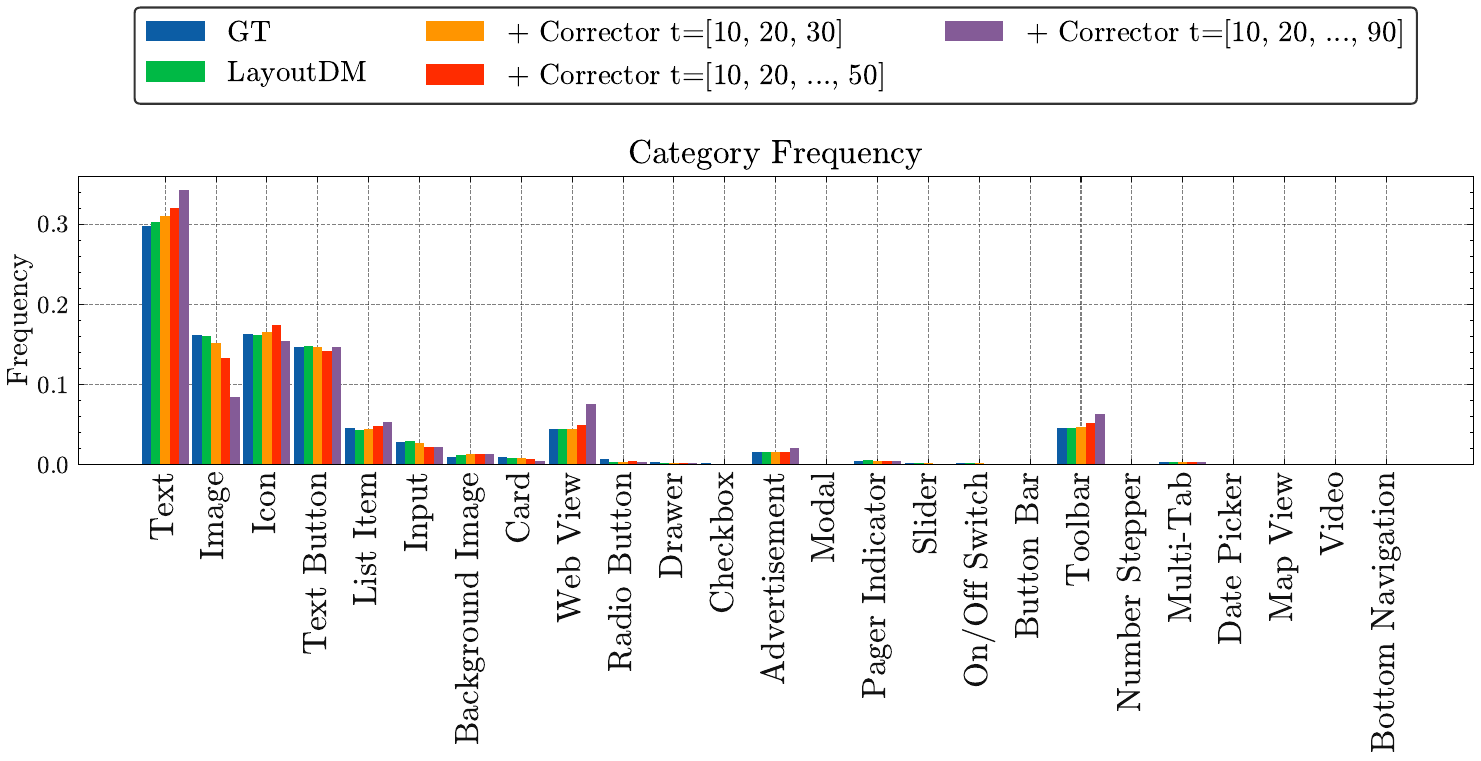}
        \caption{Category}
        \label{figure:rico_fidelity_diversity_hist_cat}
    \end{subfigure}
    
    \begin{subfigure}[b]{\fdwidth}
        \includegraphics[width=0.95\linewidth]{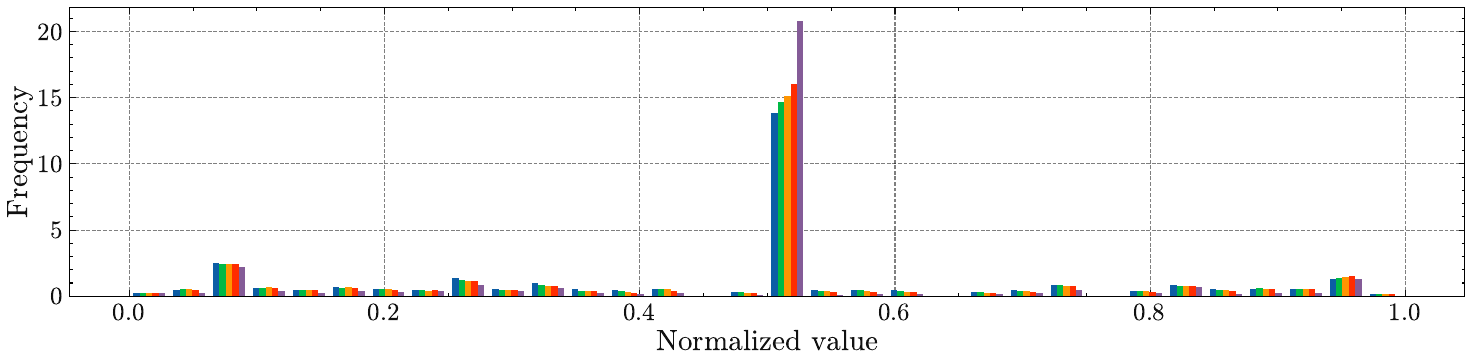}
        \caption{X-center}
        \label{figure:rico_fidelity_diversity_hist_x}
    \end{subfigure}
    
    \begin{subfigure}[b]{\fdwidth}
        \includegraphics[width=0.95\linewidth]{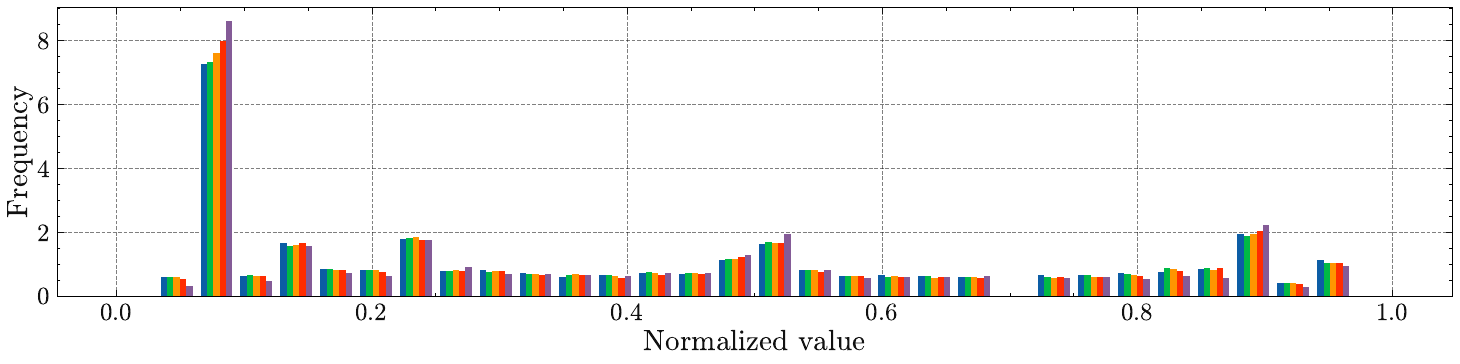}
        \caption{Y-center}
        \label{figure:rico_fidelity_diversity_hist_y}
    \end{subfigure}
    
    \begin{subfigure}[b]{\fdwidth}
        \includegraphics[width=0.95\linewidth]{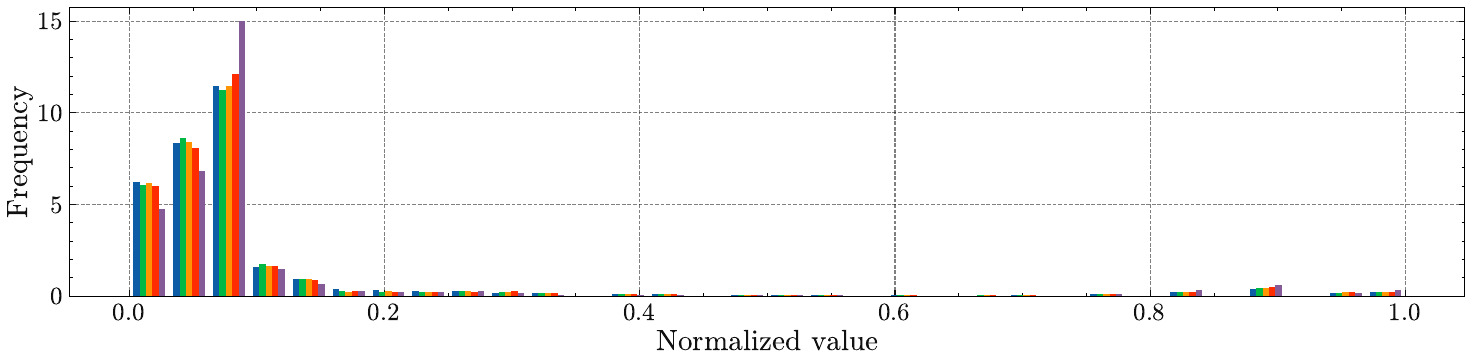}
        \caption{Height}
        \label{figure:rico_fidelity_diversity_hist_h}
    \end{subfigure}
    \caption{Histogram of the category (\cref{figure:rico_fidelity_diversity_hist_cat}), X-center (\cref{figure:rico_fidelity_diversity_hist_x}), Y-center (\cref{figure:rico_fidelity_diversity_hist_y}), and height (\cref{figure:rico_fidelity_diversity_hist_h}) of elements on the Rico dataset on different corrector schedules.}
    \label{figure:supp_attr_hist}
\end{figure*}

\end{document}